\documentclass[journal,twoside]{IEEEtran}
\usepackage{xr-hyper}

\usepackage{color, colortbl}\usepackage[hidelinks]{hyperref}
\usepackage{amsfonts,amsmath,relsize}
\usepackage{cite,latexsym}
\usepackage{graphicx,times} 

\usepackage{booktabs,multirow}
\usepackage{epsfig,epstopdf,float}
\usepackage{longtable,tabularx}
\usepackage{tikz,pgfplots, pifont}
\usepackage[mathcal]{euscript}

\usepackage{algorithmic,algorithm}

\usepackage[capitalise]{cleveref}

\hypersetup{
	colorlinks=false,
	linkcolor=blue,
	filecolor=magenta,      
	urlcolor=cyan,
}

\newcommand{\cmark}{\text{\ding{51}}}
\newcommand{\xmark}{\text{\ding{55}}}

\definecolor{LightCyan}{rgb}{0.88,1,1}

\begin{document}

\title{A Scalable Test Suite for Continuous Dynamic Multiobjective Optimisation}

\author{
Shouyong~Jiang, Marcus~Kaiser,~\IEEEmembership{Senior Member,~IEEE},  Shengxiang~Yang,~\IEEEmembership{Senior Member,~IEEE}, Stefanos~Kollias,~\IEEEmembership{Fellow,~IEEE}, and Natalio~Krasnogor \\[-5mm]
\vspace*{-3mm}
\thanks{SJ, MK, and NK acknowledge the EPSRC for funding project ``Synthetic Portabolomics: Leading the way at the crossroads of the Digital and the Bio Economies ({EP/N031962/1})''.}
\thanks{Shouyong Jiang, Marcus Kaiser, and Natalio Krasnogor are with Interdisciplinary Computing and Complex Biosystems (ICOS) research group, School of Computing, Newcastle University, Newcastle upon Tyne NE4 5TG, U.K. (email: math4neu@gmail.com, \{marcus.kaiser, natalio.krasnogor\}@ncl.ac.uk).}
\thanks{Shengxiang Yang is with the Centre for Computational Intelligence (CCI), School of Computer Science and Informatics, De Montfort University, Leicester LE1 9BH, U.K. (email: syang@dmu.ac.uk).}
\thanks{Stefanos Kollias is with the Machine Learning Group, School of Computer Science, University of Lincoln, Lincoln LN6 7TS, U.K. (email: SKollias@lincoln.ac.uk).}
}
\maketitle

\begin{abstract}
Dynamic multiobjective optimisation has gained increasing attention in recent years. Test problems are of great importance in order to facilitate the development of advanced algorithms that can handle dynamic environments well. However, many of existing dynamic multiobjective test problems have not been rigorously constructed and analysed, which may induce some unexpected bias when they are used for algorithmic analysis. In this paper, some of these biases are identified after a review of widely used test problems. These include poor scalability of objectives and, more importantly, problematic overemphasis of static properties rather than dynamics making it difficult to draw accurate conclusion about the strengths and weaknesses of the algorithms studied. A diverse set of dynamics and features is then highlighted that a good test suite should have. We further develop a scalable continuous test suite, which includes a number of dynamics or features that have been rarely considered in literature but frequently occur in real life. It is demonstrated with empirical studies that the proposed test suite is more challenging to the dynamic multiobjective optimisation algorithms found in the literature. The test suite can also test algorithms in ways that existing test suites can not.

\end{abstract}


\IEEEpeerreviewmaketitle


\section{Introduction}
\IEEEPARstart{M}{ultiobjective} optimisation problems (MOPs) involving dynamic features are reported in a variety of real applications \cite{DRK07, Fari_04:1,Zhang08, HSA11}. This kind of problems is known as dynamic multiobjective optimisation problems (DMOPs). Due to their dynamic nature, 
the optimisation of DMOPs is more challenging than that of static MOPs as it has 
to deal with not only conflicting objectives, but also changes in objective functions 
or constraints. In other words, evolutionary algorithms (EAs) for dynamic multiobjective optimisation must be capable of tracking the changing Pareto-optimal front (PF) and/or Pareto-optimal 
set (PS) to provide a set of diverse solutions that approximate each new PF or PS over time.

Dynamic multiobjective optimisation (DMO) has attracted increasing research interest 
in recent years. Significant contributions have been made mainly toward benchmarking \cite{Fari_04:1,Jiang2016_benchmark}, algorithm design \cite{ZJZ14,Jiang17_SGEA,Chen2017,Muruganantham2016}, and real-life application \cite{Zhang08,HSA11}. Despite these, there are still a number of open topics that need to be addressed in order to further advance the development of the DMO research.  One important topic is how to benchmark DMO test environments properly for algorithm analysis.

Roughly speaking, DMOPs can be regarded as a sequence of static MOPs (static and stationary are interchangable in this paper, both meaning unchanged states) for a while. In this sense, DMOPs can be constructed as a dynamic version of MOPs used in static multiobjective optimisation (SMO) \cite{Deb01}. For example, an MOP can be made dynamic by adding time-dependent elements, as observed in some early DMO studies \cite{Fari_04:1,GT09}. While this benchmarking strategy is of great use to easily develop test problems, there are also some potential limitations which need to be borne in mind. One limitation is that 
the DMOPs created in this way may be highly similar to each other and also to their static counterparts. As a result, the DMOPs have simple and undiversified problem properties. For example, some FDA \cite{Fari_04:1} and dMOP \cite{GT09} problems derived from the static ZDT \cite{ZDT00} test suite cannot capture (time-varying) mixed PFs which greatly challenge algorithms \cite{Jiang2016_benchmark}. A lack of diverse problem properties in benchmarks may not provide a comprehensive analysis of algorithms.

Another limitation is that the base MOPs can be the dominating factor for the difficulty of the resulting DMOPs and therefore decrease the importance of dynamics. A hard-to-solve MOP may not be a good candidate for DMOPs \cite{Gee2017}. This is because the critical factor that causes the failure of algorithms to solve the DMOP is not the underlying dynamics but rather the static properties (e.g. strong variable dependencies) of the problem. Some DMOPs like UDF \cite{BDSC14} and ZJZ \cite{ZJZ14} were created based on the LZ test suite \cite{LZ09}, but it remains unclear, of the dynamic and static features they involve, which is more challenging to algorithms, since LZ is already challenging in the SMO literature \cite{JY16_TPN,Zhou2016}.

Having realised the above limitations that exist in DMO test problems, Jiang and Yang \cite{Jiang2016_benchmark} have proposed a new benchmark generator for constructing DMOPs. While focusing more on the effect of dynamic features, the proposed benchmark also captures a number of dynamic features that have not or rarely been considered, such as time-varying mixed PF, mixed type of change mode. In another recent work \cite{Gee2017}, dynamics were intentionally outweighed over static properties when existing MOPs are used to create dynamic scenarios regarding modality, PF tradeoff connectivity, and PF degeneration.

Despite much effort, DMO benchmarking is still far from satisfying. It remains unknown what characteristics are desirable for a good test suite. For example, one important characteristic is the scalability of objectives in DMOPs. It is observed that many real-life applications have more objectives, and the number of objectives can be time-varying [6]. The scalability of objectives, the nature of its dynamics, and the balance between them are poorly studied in existing DMOPs [4], [19]. 
Another important characteristic that a good test suite should embrace is the detectability of environmental changes. To the best knowledge of the authors, there is no test problem whose changes cannot be detected with one re-evaluation of a random population member. As a result, most algorithms simply assume all changes are knowable, leading to the neglect of this important factor in the existing literature. 

In summary, although after more than a decade of research, DMO benchmarking still face a number of issues, which are highlighted as follows:
\begin{enumerate}
	\item \textit{Problems are not scalable or have limited scalability}. Most constructed DMOPs have a few objectives and the number of objectives is time independent \cite{Jiang2016_benchmark,BDSC14}.
	
	\item \textit{Static properties outweigh dynamics}. For example, deceptive or strong variable-linkage characteristics obscure the importance of dynamics \cite{BDSC14,ZJZ14}.
	
	\item \textit{Problem properties are similar and not diversified}. DMO benchmarks have similar PF/PS properties and environmental changes, such as FDA \cite{Fari_04:1} and dMOP\cite{GT09} cases. 
	
	\item \textit{Dynamic features are very limited.} Most existing benchmarks cover a small subset of dynamics from real-life applications, and leave 
	unconsidered many other dynamic features, e.g. predictability and visibility \cite{Nguyen2012}.
\end{enumerate}
Bearing these in mind, it is not trivial to rethink what characteristics contribute to a good DMO test suite. After enumerating and justifying a number of desirable characteristics that a test suite should have, we are motivated develop a new test suite for DMO. The test suite is aimed to present characteristics linking closely to real-world applications while addressing as many as possible the above-mentioned gaps, and to potentially serve as adversarial counter cases \cite{Ghosh2018} for testing DMO solvers. It provides a systematic tool for algorithm designers and practitioners to comprehensively analyse and develop algorithms.

The rest of this paper is organised as follows. Section II introduces related work and presents desirable features and dynamics for DMOPs. The test suite construction process and the resulting test suite are described in Section III. Section IV presents experimental studies on the proposed test suite. Section V concludes the paper.

\section{Related Work}
\subsection{Existing Dynamic Multiobjective Test Problems}
One of the earliest test suite is FDA \cite{Fari_04:1}. FDA suggested a benchmark design methodology which considers dynamics and multiobjectivity separately. Following this idea, five FDA problems were made by borrowing two important test suites, i.e. ZDT \cite{ZDT00} and DTLZ \cite{DTLZ01}, from the SMO literature, and adding dynamics of interest to them. Since establishment, the FDA design methodology has impacted greatly a number of DMO benchmarking studies \cite{GT09,HSA11,BDSC14,HE14,Gee2017}. Unlike FDA, Jin and Sendhoff \cite{JS04} developed an open scheme of aggregating objective functions of static problems by dynamically changing weights to form a low-dimensional DMOP. However, no well-defined DMOPs were derived from this scheme. Guan {\em et al.}~\cite{GCM05} studied DMOPs 
with objective replacement, where some objectives may be replaced with new objectives 
during the evolution. Mehnen {\em et al.}~\cite{MRW06} argued that the DTLZ and ZDT test 
suites are already challenging in their static version, and simpler test functions are 
needed to analyse the effect of dynamics in DMOPs. Hence, they suggested the DSW functions 
for DMOPs. Furthermore, they proposed a new generic scheme DTF, which is a generalised 
FDA function and allows a variable scaling of the complexity of the dynamic properties. 
They also added scalable and dynamic constraints to DMOPs by moving circular obstacles 
in the objective space. Recently, Helbig and Engelbrecht \cite{HE14} have made a sound 
investigation into the existing DMOPs used in the literature, and have highlighted the 
characteristics that an ideal DMO benchmark function suite should exhibit. In addition 
to highlighting shortcomings of existing DMOPs, they also provided several HE problems with complicated static characteristics borrowing from the SMO literature \cite{HHBW06,LZ09}. As a result, main challenges that the HE problems pose to algorithms are static problem properties instead of dynamics. Similarly, the recent UDF problems \cite{BDSC14} and Zhou \emph{et al.}'s work \cite{ZJZ14} introduce complicated static variable linkages, which can obscure the importance of dynamics.

In the meanwhile, there are several studies concentrating on dynamic aspects of DMOPs.  Huang \emph{et al.} \cite{HSA11} proposed some DMOPs where the number of objectives or variables can change over time. Following this direction, Chen \emph{et al.} \cite{Chen2017} highlighted challenges in DMOPs with a changing number of objectives, e.g. severe diversity loss, and suggested more DMOPs of this type. The study in \cite{Jiang2016_benchmark} proposed a new problem generator that can produce various time-dependent PF geometries for DMOPs. It also suggested some important dynamics, including time-varying variable linkages that cause unbalanced diversity, and mixed types of change modes. But, the study focused mainly on multiobjective cases and had scalability issues, which may hinder its wide application. Gee \emph{et al.} \cite{Gee2017} developed a number of GTA problems targeting dynamic modality, tradeoff connectivity, and PF degeneration. Despite great usefulness, the GTA problems limit performance assessment of EAs to the target dynamics.

There are also studies on dynamic time-linkage problems (DTPs) where solutions found for previous environments determine the behaviour of problems for future environments and this type of problems is very common in real applications \cite{Nguyen2012,Nguyen2012-timelinkage}. Almost all the studies about DTPs focus mainly on single-objective scenarios except \cite{HSA11}. DTPs can exhibit various characteristics, e.g. time/prediction deception \cite{Nguyen2012-timelinkage}, and many characteristics are not fully known. Benchmarking DTPs can be very difficult in dynamic multiobjective scenarios and therefore is not considered in this paper.

\vspace{-4mm}
\subsection{Desirable Features under Dynamic Environments} \label{desirables}
This subsection presents a number of desirable features and dynamics that have been ignored or rarely discussed in the literature, although we recognise that other dynamic features such as time-dependent variable dependencies, nonseparability are also important under certain circumstances.
Note that, the first two features are not dynamic but specially highlighted here due to importance in any scalable test problems.

\subsubsection{Scalability}  DMO test problems should be scalable to have any number of objectives and variables.

Importance: Scalable problems are beneficial for assessing the effect of variation in the number of objectives and variables. Similar to multiobjective benchmarks \cite{HHBW06}, this property should be considered when building DMO test problems.

\subsubsection{Shape of PF boundaries} The boundary of a $M$-objective PF after max-min normalisation fully or partially lies on the hyperplanes defined by all combination of $(M-1)$ axes. For example, normalised PF boundaries can be similar to those of a unit simplex or sphere in the case of $M=3$, a regular shape commonly used for 3-objective test problems.

Importance: The shape of PF boundaries have not been well studied, but it has recently attracted increasing attention due to its significant effect on some algorithms \cite{ISMN2017}. The majority of existing scalable test problems are designed such that the PF boundary (with or without normalisation) is triangular. However, real-world problems \cite{Giuliani2014,Pieri2018} often have more diverse PF shapes instead of triangles. Intuitively, irregular PF boundaries affect directly solution spread of algorithms in objective space, and therefore possibly affect objective normalisation which is very important for solution uniformity for certain problems. The shape of PF boundaries can also possibly influence the computation of corner solutions used in some objective reduction techniques when dealing with many objectives \cite{Singh2011}. 

\subsubsection{Dynamic concavity-convexity} The PFs of a DMOP are dynamically convex, linear, concave, or mixed in a series of changes.

Importance: Concavity-convexity of DMOPs has a direct impact on the performance of algorithms, as evidenced in their static counterparts \cite{HHBW06,JY16_TPN}. When a DMOP changes its concavity-convexity from one type to another due to an environmental change, it can cause the variation of uniformity of solutions along the PF and potentially lose diversity.  As a consequence, EAs face difficulties in uniformly redistributing solutions along the PF or rescuing diversity loss, the latter of which is even a big issue in many-objective cases \cite{Jiang17_SPEAR}.  

Time-varying concavity-convexity has been discussed in a number of biobjective DMOPs \cite{Fari_04:1,Jiang2016_benchmark}. However, this kind of characteristic have not been well understood in DMOPs with a larger or scalable number of objectives. 

\subsubsection{Dynamic PF connectivity} The PF has time-dependent connectivity. At any time the PF can be simply-connected, non-simply connected (e.g., a surface with holes), or have several disconnected PF segments. 

Importance: The connectivity of the PF manifold has a great influence on the performance of EAs, as demonstrated in \cite{JY16_TPN}. Time-varying PF connectivity has existed in some real-life problems \cite{Fari_04:1}. In the DMO literature, most DMOPs are constructed to have a simply-connected PF manifold. Although there are several PF-discontinuous DMOPs, none of them is non-simply connected.
Benchmarking DMOPs with time-dependent PF connectivity (i.e., time-varying number of PF segments or non-simply connected  PF) can help to further our understanding of the capability of EAs.

\subsubsection{Dynamic PF shapes} The overall PF shape is not fixed, but rather diverse over time.

Importance: Changes in PF shapes over time have been reported in many real applications. In control systems \cite{Butans2011,Zhang08}, the size of the PF can increase or decrease, resulting in different PF shapes over time. This type of change is very likely to affect solution distribution on the PF. In another study \cite{Xiong2017}, the PF varies partially subject to environmental changes. Partial PF variations may increase the difficulty of change detection for PF-based detection methods.  

\subsubsection{Dynamic modality} Objective functions can have different modal modes (either unimodal or multimodal) over time. 

Importance: Multimodal problems are generally more challenging than unimodal ones, as shown in a number of studies of static multiobjective optimisation \cite{DTLZ01,ZDT00}. They are an important type of optimisation problems that arise frequently from real-world applications. However, multimodal problems have not been well explored in DMO. Most existing DMO test problems are unimodal rather than multimodal, and a few multimodal problems have either a fixed number of local optima or a fixed location of global optima. Dynamic modality changes problem landscapes, making DMOPs more difficult to solve. It also helps to assess the change detection ability of EAs if the global optima is relocated in a new environment.

\subsubsection{Dynamic PF/solution favourability} From decision makers' point of view, dynamic PF favourability relates to time-dependent preference of some solutions, e.g. knee points/regions \cite{Zou2017}. From EAs' point of view, it relates to certain solutions that are easier to obtain than others \cite{Jiang17_SF,Liu2017} in different environments. 

Importance: Dynamic PF favourability regarding knee points is of high interest to the decision maker. This is because most often such points are more important and dynamically preferred. Likewise, the existence of PF favourability of solutions can create imbalanced difficulties of approximation to different PF regions. As a result, population members of EAs are very likely to be attracted toward easily obtainable PF regions, leaving harder PF regions under-explored or poorly approximated \cite{Liu2017}. This dynamic feature has been observed in \cite{Li2016}. 

\subsubsection{Time-dependent number of objectives/variables} In dynamic environments, DMOPs are allowed to add new or remove some existing objectives/variables. Note that, this is different from scalability for which the number of objectives remains unchanged over time.

Importance: DMOPs with a time-varying number of objectives/variables appear in many real-life scenarios \cite{Abello2014,Shen2015}, but have been rarely studied in the DMO literature. The effect of changes in the number of objectives in regular pattern has been recently studied in \cite{Chen2017}. Changes in the number of variables or objectives in irregular pattern can complicate DMOPs and create another scenario to study dynamics.

\subsubsection{Time-dependent degeneration} Most often, the PF of $M$-objective problems is a  $(M-1)$-dimensional manifold under Karush-Kuhn-Tucker conditions.   A degenerate PF can be of lower dimension than $(M-1)$. For example, a 3-D problem may degenerate to have a PF  that is a curve or line segment.

Importance: PF-degenerate problems can raise serious challenges to some algorithms. One challenge, for example, can be the need of effective diversity maintenance to guarantee a uniform spread of solutions. Degeneration has been reported in PID controller design problems \cite{Fari_04:1}. 

\subsubsection{Detectability} A DMOP can be easily or hardly detectable, depending on how much environmental changes are computationally visible to optimisers.

Importance: Detectability of DMOPs is of great importance, but it has received little attention in the field of EMO. Almost all existing DMO test problems are made easily detectable for every single environmental change, and a population-based algorithm therefore can exactly detect the change by simply re-evaluating any one solution of population \cite{Sahmoud2016}. The importance of detectability has been recognised in dynamic single-objective optimisation \cite{Richter2009}. In DMO, failing to detect changes that have actually occurred not only results in ineffective tracking of  the new PF, but also could mislead the search process  because nondominated solutions obtained so far may be no longer nondominated in new environments.

\subsubsection{Predictability} The location of PS can change in a predictable or unpredictable manner over time.

Importance: The importance of the predictability of optima has been already recognised in dynamic single-objective optimisation \cite{Yang2010}. Existing DMO test problems were made to vary PS locations in a regular and predictable pattern, e.g., the popular FDA \cite{Fari_04:1} and dMOP \cite{GT09} test problems have a PS of $x_{i=M:n}\!\!=\!\!\sin(0.5\pi t)$ ($n$ is the number of variables) at time $t$. It has been found that the dynamics of PS are more likely unpredictable and sometimes even random in many real-life problems \cite{Nimmergeers2016}. This gap should be addressed by investigating DMOPs with unpredictable or random changes of PS.

\section{SDP: Scalable Dynamic Multiobjective Test Suite}
\subsection{Basic Framework}
The proposed SDP test suite was constructed with component functions, a way that has been commonly used in building  various popular multiobjective problems, such as DTLZ \cite{DTLZ01}, WFG \cite{HHBW06}, and LZ  \cite{LZ09} test suites. Mathematically, SDP can be described as:
\begin{equation}
\text{Minimise} \qquad (f_1({\bf x},t), \dots, f_M({\bf x},t))
\label{eq:sdp}
\end{equation}
with
\begin{equation}
f_{i=1:M}({\bf x},t)=(1+g({\bf x},t))\mu_i({\bf x},t)+\nu(t)
\label{eq:sdpf}
\end{equation}
where
\begin{itemize}
\item $\mu({\bf x},t)=(\mu_1,\dots, \mu_M)$ is a time-dependent function that describes PF properties, such as geometry, degeneration.

\item $g({\bf x},t)$ is a time-dependent function that defines PS properties, such as optima location, modality, variable dependency. The PF of (\ref{eq:sdp}) is obtained when $g({\bf x},t)=0$.

\item $\nu(t)$ is a time-dependent function that allows the PF to move away from or back to its initial location (i.e., the PF at $t=0$). In the work, $\nu(t)$ is either zero or $|\sin(0.5 \pi t)|$.
\end{itemize}
${\bf x} = (x_1, \dots, x_n) \in R^{n}$ is the decision variable vector and can be divided into two subvectors, i.e., the distance-related ${\bf x_I}=(x_1, \dots, x_{M-1})$ for $\mu({\bf x},t)$ and position-related ${\bf x_{II}}=(x_M, \dots, x_{n})$  mainly for $g({\bf x},t)$, according to \cite{HHBW06}. This work constrains the search space ${\bf x}$ to lie in $[0,1]^{n}$, unless otherwise stated. The search space can be also scaled to a bigger range. $t\in R$ is the discrete time instant. $f_i({\bf x},t)$ is the $i$-th objective value of solution ${\bf x}$ at time $t$.

\vspace{-2mm}
\subsection{Building Component Functions}
In what follows, component functions $g({\bf x},t)$ and $\mu({\bf x},t)$ of (\ref{eq:sdpf}) are built to adhere to the desirable features in Section \ref{desirables}. $\nu(t)$ is ignored as it is fairly easy to build. For example $\nu(t)$ can be assigned zero if disregarding the shift of PF or $|\sin(0.5\pi t)|$ if considering a time-varying shift.

\vspace{1mm}
\noindent\emph{{1) Scalability and Continuous PF Geometry}}
\vspace{1mm}

These two are determined by $\mu({\bf x},t)$ and often coupled, and here PF convexity-concavity and boundaries shapes are the focus of discussion. In the SMO literature, $\mu({\bf x},t)$ was built to be a scalable linear hypersimplex \cite{HHBW06,Deb01} defined as 
\begin{equation}
\mu_{linear,i}\!\!=\!\!\begin{cases}
(1-x_{i})\prod_{j=1}^{i-1}x_j \!\!&\!\!\!\text{if~$i=1$}\\
x_1\dots x_{M-2} x_{M-1} \!\!&  \!\!\!\text{if~$2 \!\! \leq \!\! i \!\!<\!\! M$}\\
\end{cases}
\label{eq:sdp_linear}
\end{equation}
or spherical hypersurface (the first hyperorthant) defined as
\begin{equation}
\mu_{sphere,i}\!\!=\!\!\begin{cases}
\sin(\!{y_{i}}\!)\!\!\prod_{j=1}^{i-1}\!\!\cos(\!{y_j}\!) \!\!&  \!\!\!\text{if~$1 \!\! \leq \!\! i \!\!<\!\! M$}\\
\cos({y_1})\dots\cos(y_{M-1}) \!\!&\!\!\!\text{if~$i=M$,}\\
\end{cases}
\label{eq:sdp_sphere}
\end{equation}
where $y_i=0.5\pi x_i$. Both of them have non-concave geometries. Such plain PF shapes are well defined and easy to understand, but insufficient to help draw a comprehensive conclusion on the overall performance of any assessed algorithms. The limitation of hypersimplex or hypersphere based PFs has been identified recently \cite{ISMN2017}. It is also noticed that real-world applications have more complicated and diverse PF geometries \cite{Giuliani2014,Pieri2018}. In  what follows, we introduce several selected novel PF geometries with interesting properties.

\begin{figure*}[th]
	\centering
	\begin{tabular}{ccc}
		\includegraphics[width=0.32\linewidth]{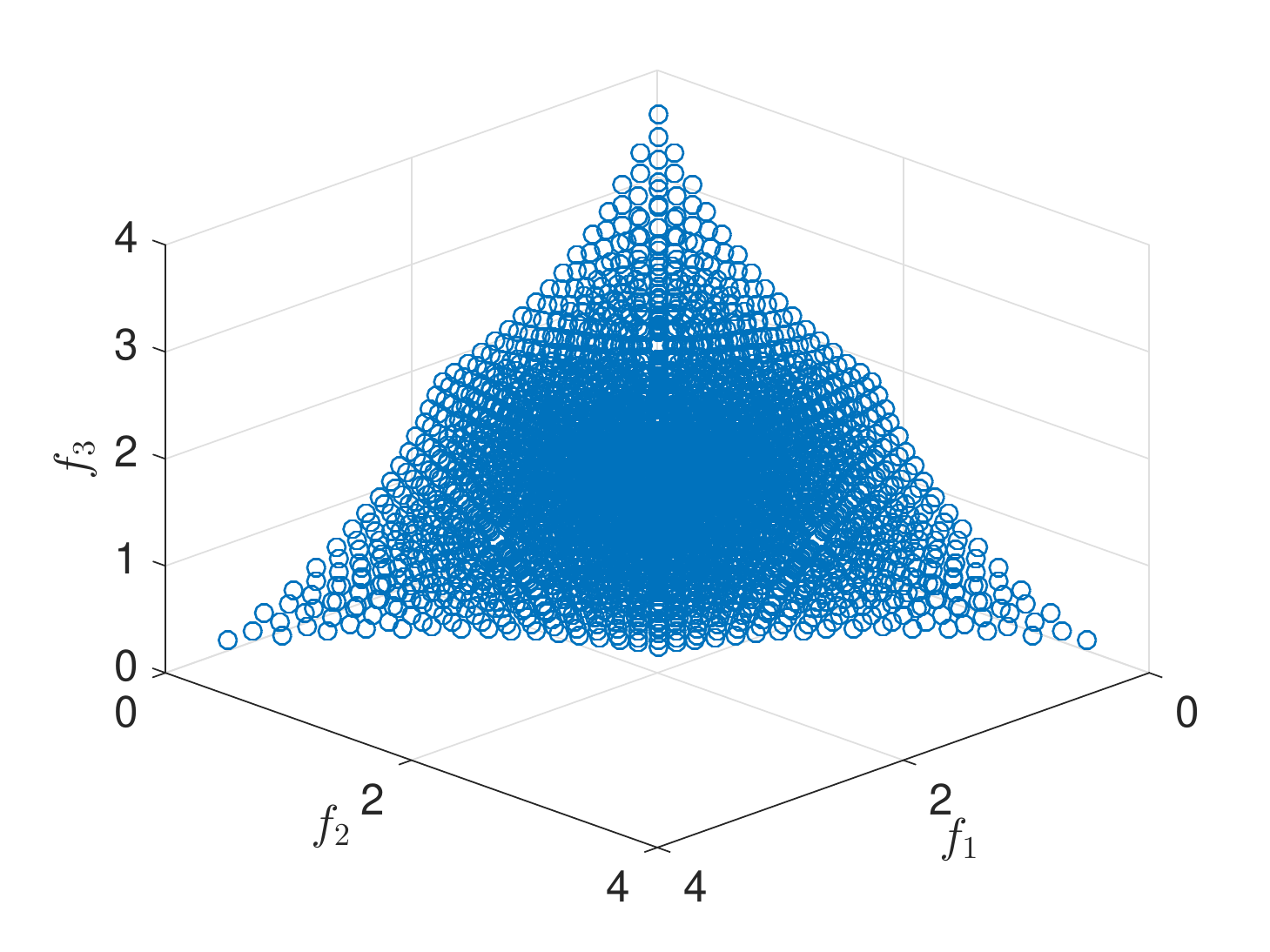} &
		\includegraphics[width=0.32\linewidth]{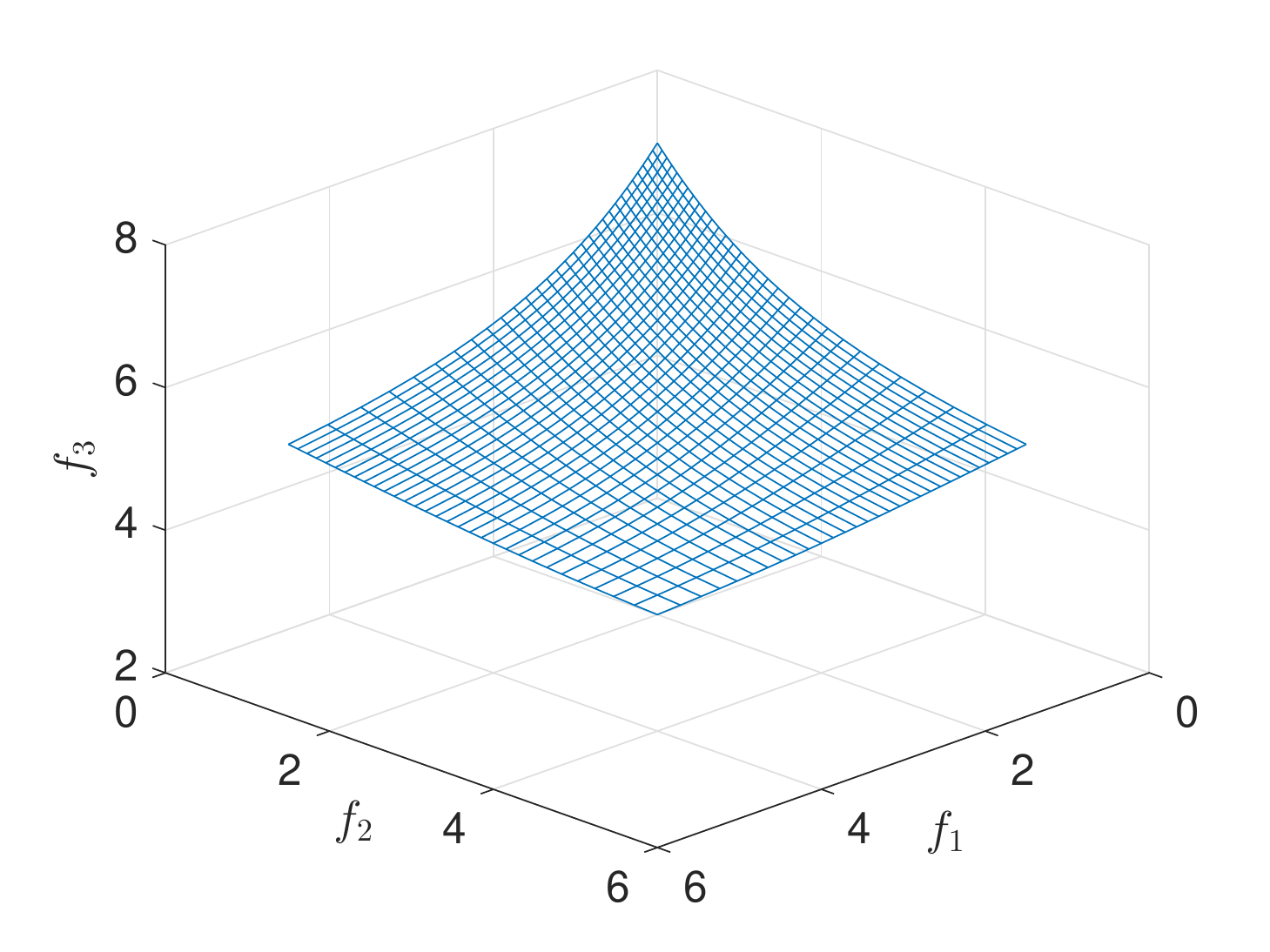} &
		\includegraphics[width=0.32\linewidth]{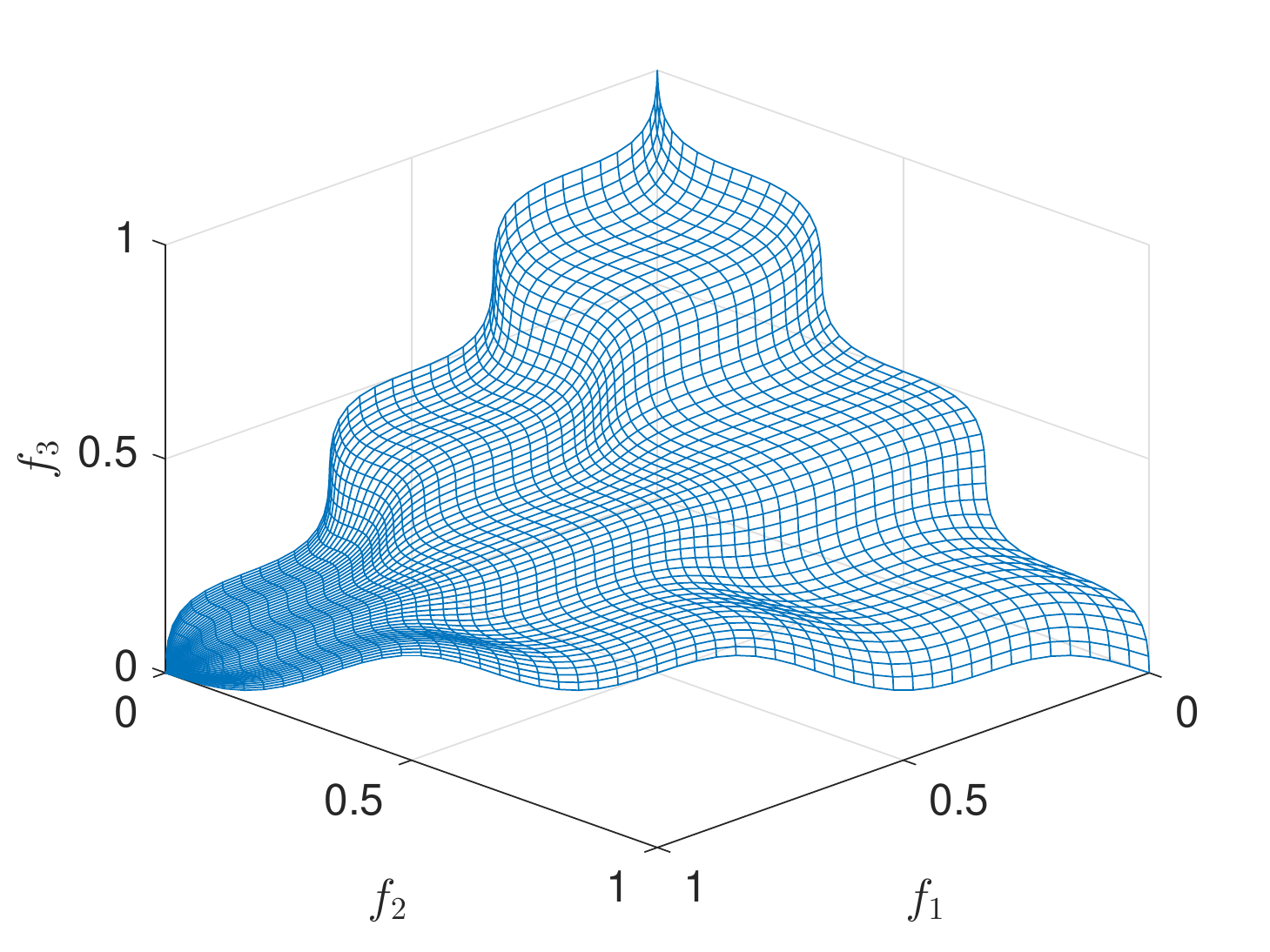}\\[-1mm]
		(a) & (b) & (c)\\[-2mm]
	\end{tabular}	
	\caption{Continuous triobjective PF geometries. (a) product-form PF with 1000 random variable vectors generated from [1,4]$^3$ at time $t=0$. (b) PF in the form of sum of reciprocals. (c) PF with knee regions with $w=6$.}
	\label{fig:sdp1_pf0}
\end{figure*}

\paragraph{Product-form PF} The product of all objective values is one. It means each objective is inversely proportional to other objectives. The PF is defined as 
\begin{equation}
\mu_{\mathsmaller{\mathsmaller{\prod}},i}=\frac{x_i}{\sqrt[M-1]{\prod\nolimits_{j=1,j \neq i}^{M} {x_j}}}
\label{eq:sdp1_pf}
\end{equation}
It is noticed that solutions are not uniformly distributed on the PF (Fig.~\ref{fig:sdp1_pf0}(a)), implying boundary solutions be much difficult to be found than intermediate solutions. Note that, the overall curvature of this  PF can be adjusted by a mapping: $\mu \rightarrow \mu^c$, where $c$ can be a constant (i.e, 2) or a time-varying function (e.g., $|\sin(t)|$). The mapping can further complicate the PF geometry and is applicable to any scalable PF mentioned in this paper.

\begin{figure}[th]
	\centering
	\includegraphics[width=0.65\linewidth]{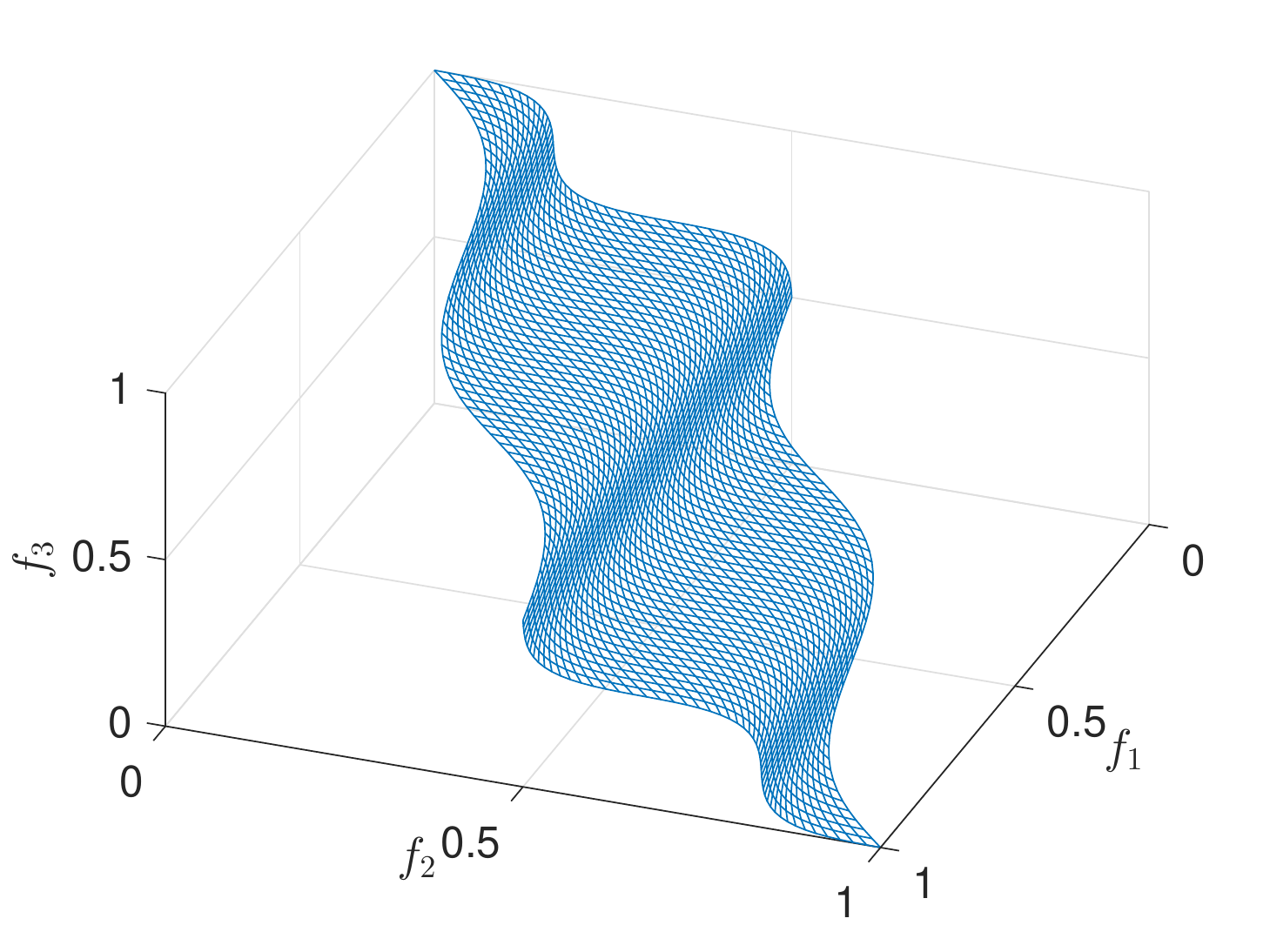}\\[-2mm]	
	\caption{Illustration of a triobjective mixed PF.}
	\label{fig:sdp4_pf0}
\end{figure}

\paragraph{Sum-of-reciprocals PF} The sum of reciprocals of $f_i+1$ equals one. $\mu({\bf x},t)$ is defined as
\begin{equation}
\mu_{\mathsmaller{\mathsmaller{\sum}},i} = \begin{cases}
\frac{1+t+\sum_{j=1, j\ne i}^{M-1}{x_j}}{x_i}  &\text{if $1 \leq i < M$}\\
\sum\nolimits_{i=1}^{M-1}{\frac{x_i}{1+t}} &\text{if $i=M$,}
\end{cases}
\label{eq:sdp2_pf}
\end{equation}
One important feature of this PF is that corner PF points are not located on the axes (Fig.~\ref{fig:sdp1_pf0}(b)), and algorithms relying on them for objective normalisation can be affected. Also, the irregular shape of the boundary poses challenges to decomposition-based algorithms \cite{ZL07} that employ weight vectors, as not all weight vectors sampled from a unit simplex pass through the PF.

\paragraph{Mixed/knee PF} The PF has both local convex and concave regions, and sometime produce some knee points that may be hard to approximate. A mathematical description can be
\begin{equation}
\mu_{knee,i} \!=\!\! \begin{cases}
\begin{aligned}
&(1\!-\!x_{i}\!+\!0.05\sin(w \pi x_{i}))\times\\[-1.5mm]
&~~ \prod\nolimits_{j=1}^{i-1}(x_{j}\!+\!0.05\sin(w\pi x_{j}))
\end{aligned}
&\text{if $1 \!\!\leq\!\! i \!<\!\! M$}\\[4mm]
\prod_{i=1}^{M-1}(x_{i}+0.05\sin(w\pi x_{i})) &\text{if $i=M$,}
\end{cases}
\label{eq:sdp3_pf}
\end{equation}
where $w$ is a parameter (either static or dynamic) controlling the number of knee regions on the PF (see the illustrative PF in Fig.~\ref{fig:sdp1_pf0}(c)). 
$p_{i}({\bf x},t)$ of (\ref{eq:sdp3_pf}) can have a slight adaptation
\begin{equation}
\mu_{mix,i} = \begin{cases}
x_i &\text{if $1 \!\leq\! i \!<\! M\!-\!1$}\\
(s+0.05\sin(w \pi s)) &\text{if $i=M-1$}\\
(1\!-\!s\!+\!0.05\sin(w \pi s)) &\text{if $i=M$,}
\end{cases}
\label{eq:sdp4_pf}
\end{equation}
where $s=\sum_{i=1}^{M-1}{x_i}/(M-1)$ and $w=\text{sgn}(rnd_t-0.5)\lfloor 6|\sin(0.5\pi t)|\rfloor$ ($sgn(\cdot)$ is the sign function and $rnd_t$ is a random value in [0,1]) is recommendation of use. This change results in a very different PF shape, 
and the PF has mixed regions and its boundary is irregularly shaped (Fig.~\ref{fig:sdp4_pf0}), which can be challenging for some algorithms \cite{ISMN2017}.

\vspace{1mm}
\noindent\emph{{2) PF Connectivity}}
\vspace{1mm} 
\setcounter{paragraph}{0}

The PF can be simply connected (sometimes called continuous), non-simply connected, or disconnected. Here, we focus on the construction of non-simply connected and disconnected $\mu({\bf x},t)$.

\begin{figure}[t]
	\centering
	\includegraphics[width=0.85\linewidth]{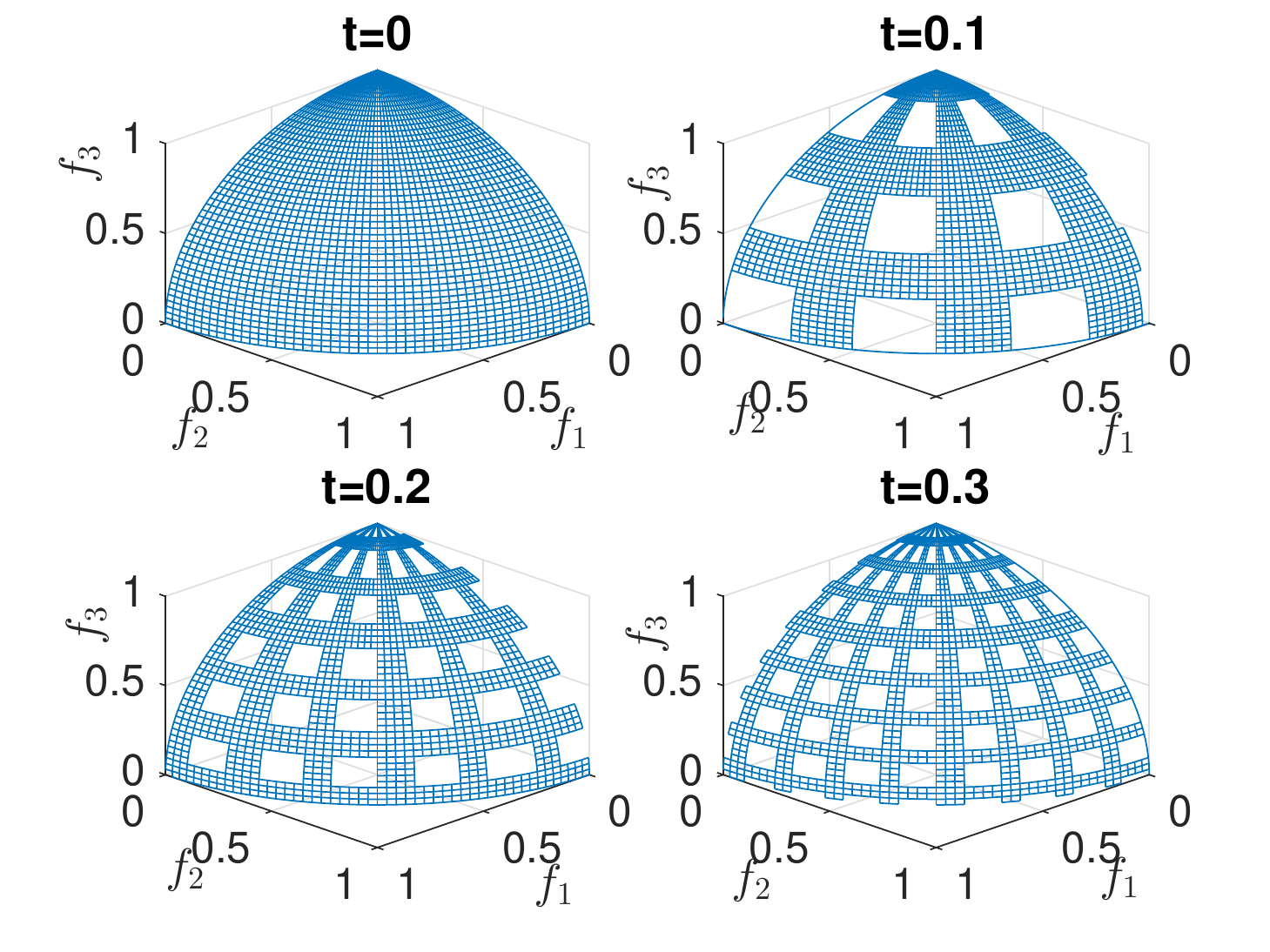}\\[-2mm]	
	\caption{Illustration of a triobjective  non-simply connected PF.}
	\label{fig:sdp8_pf0}
	\vspace{-1mm}
\end{figure}
\paragraph{Non-simply connected PF}
To the best of the authors' knowledge, there is no problem in the multiobjective literature that has a non-simply connected PF. To bridge this gap, we put forward a way to construct a non-simply connected PF. It works as follows. Let $\mu({\bf x},t)$ describes a simply connected PF, it is multiplied by $g({\bf x},t)$, a function that has a value larger than one for some ${\bf x}$ components and equals one for others. ${g}({\bf x},t)$ deforms some regions of the PF of $\mu({\bf x},t)$ such that points on these regions are dominated by the other regions. For example, the spherical PF $\mu_{sphere,i}$ can be multiplied by
\begin{equation}
	g_{nsc}=1\!+\!\left|\prod\nolimits_{j=1}^{M\!-\!1}{\sin({\lfloor{k_t(2x_j\!-\!r)}\rfloor}\pi/2)}\right|,
\end{equation}
where $r=1-\!\!\!\!\mod(k_t,2)$ and $k_t=\lfloor 10\sin(0.5\pi t)\rfloor$ is used to produce a changing number of holes on the PF generated by ${g}({\bf x},t)\mu({\bf x},t)$ (see Fig.~\ref{fig:sdp8_pf0}). 

\begin{figure}[t]
	\centering
	\includegraphics[width=0.85\linewidth]{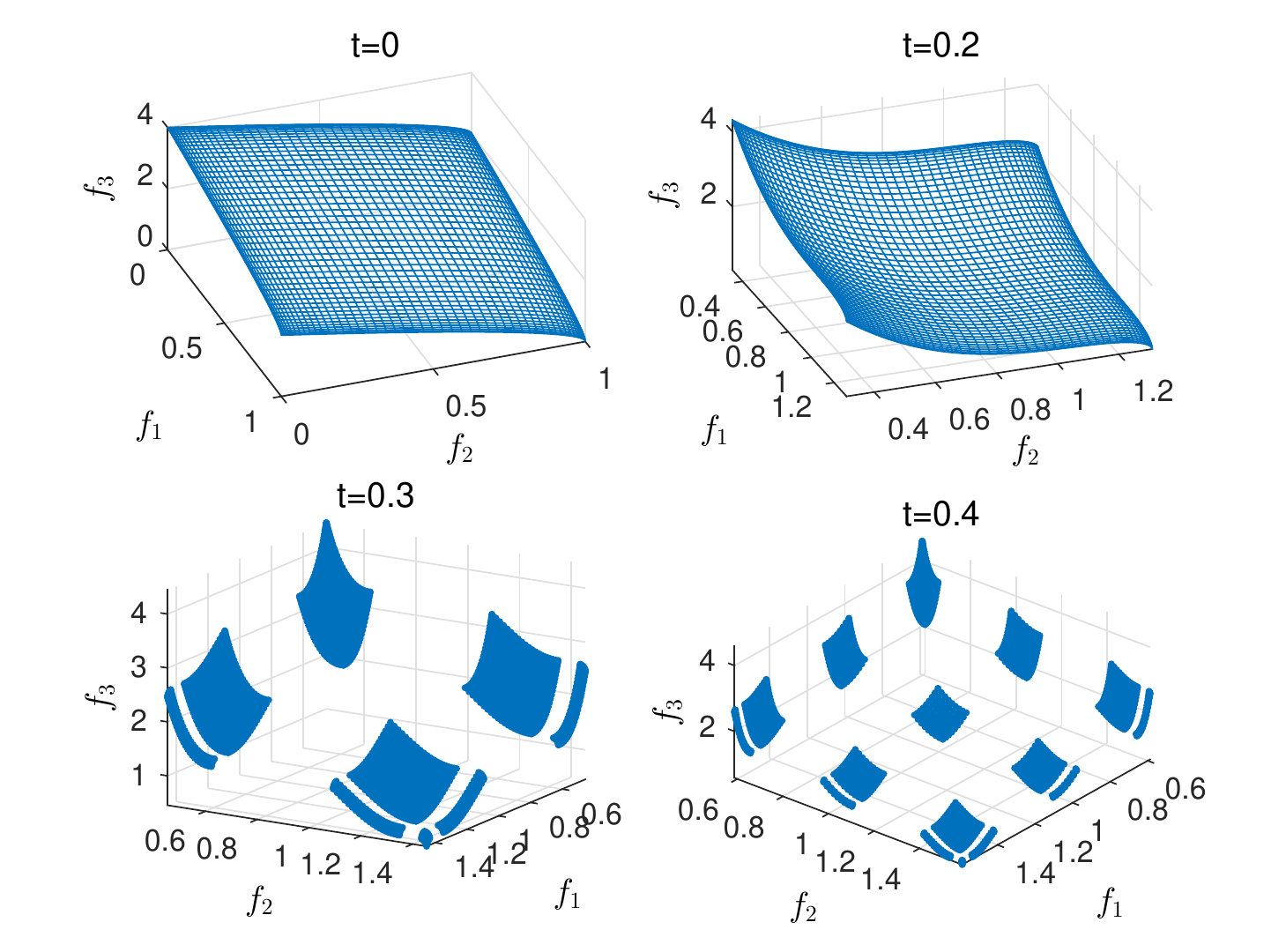}\\[-2mm]	
	\caption{Illustration of a triobjective disconnected PF obtained from $\mu_{disc1}$.}
	\label{fig:sdp9_pf0}
\end{figure}

\begin{figure}[t]
	\centering
	\includegraphics[width=0.85\linewidth]{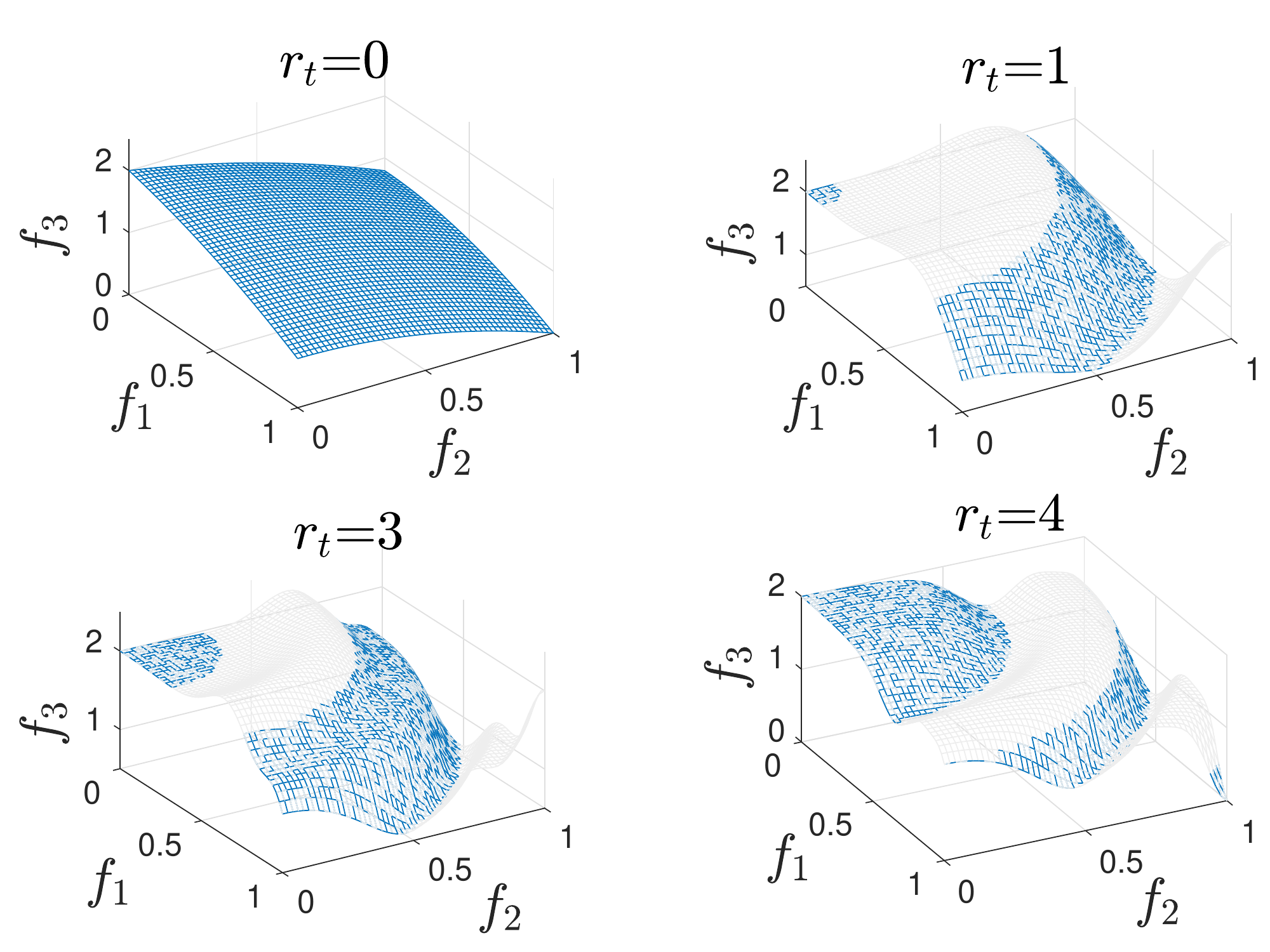}\\[-2mm]	
	\caption{Illustration of a triobjective disconnected PF (blue) obtained from $\mu_{disc2}$ with different $r_t$ values. Blue and grey segments comprise the whole geometry of $\mu({\bf x},t)$.}
	\label{fig:sdp10_pf0}
\end{figure}

\paragraph{Disconnected PF}
Unlike non-simply connected PFs, a disconnected PF consists of multiple disconnected PF components. This type of PF can be generated from 
\begin{equation}
\mu_{disc1,i}\!\!=\!\!\begin{cases}
\!\!\cos^2(0.5\pi x_{i}) \!\!& \!\!\!\!\!\!\text{if~$1 \!\! \leq \!\! i \!\!<\!\! M$}\\
\!\!\begin{aligned}
&\sum\nolimits_{j=1}^{M-1}{[\sin^2(0.5\pi x_j)}+\\[-1.5mm]
&~~ \sum\nolimits_{j=1}^{M\!-\!1}{\!\!\!\!\!\!\sin(0.5\pi x_j) {\cos^2(k_t\pi x_j)}]}
\end{aligned}
\!\!& \!\!\!\!\!\!\text{if~$i=M$,}\\
\end{cases}
\label{eq:sdp9_pf}
\end{equation}
where $k_t$ ($k_t\!\!=\!\!\lfloor 6\sin(0.5\pi t)\rfloor$ is suggested) renders a changing number of disconnected components (Fig.~\ref{fig:sdp9_pf0}).
Alternatively, we can also construct a disconnected PF as follows:
\begin{equation}
\mu_{disc2,i}\!\!=\!\!\begin{cases}
x_i & \!\!\!\text{if~$1 \!\! \leq \!\! i \!\!<\!\! M$}\\
(2-s-\sqrt{s}(-\sin(2.5\pi s))^{r_t}) &  \!\!\!\text{if~$i=M$,}\\
\end{cases}
\label{eq:sdp10_pf}
\end{equation}
where $s=\sum_{i=1}^{M-1}{x_i^2}/(M-1)$ and $r_t$ is a time-varying parameter impacting the PF shape, which is suggested to be $\lfloor10|\sin(0.5\pi t)|\rfloor$. Different $r_t$ settings result in various PF characteristics (Fig.~\ref{fig:sdp10_pf0}).

\vspace{1mm}
\noindent\emph{{3) Detectability}}
\vspace{1mm} 
\setcounter{paragraph}{0}


Environmental changes of existing DMOPs are made easily detectable within one re-evaluation of any solutions. Real-world dynamic changes should not be so simple.
\begin{figure}[th]
	\centering
	\includegraphics[width=0.65\linewidth]{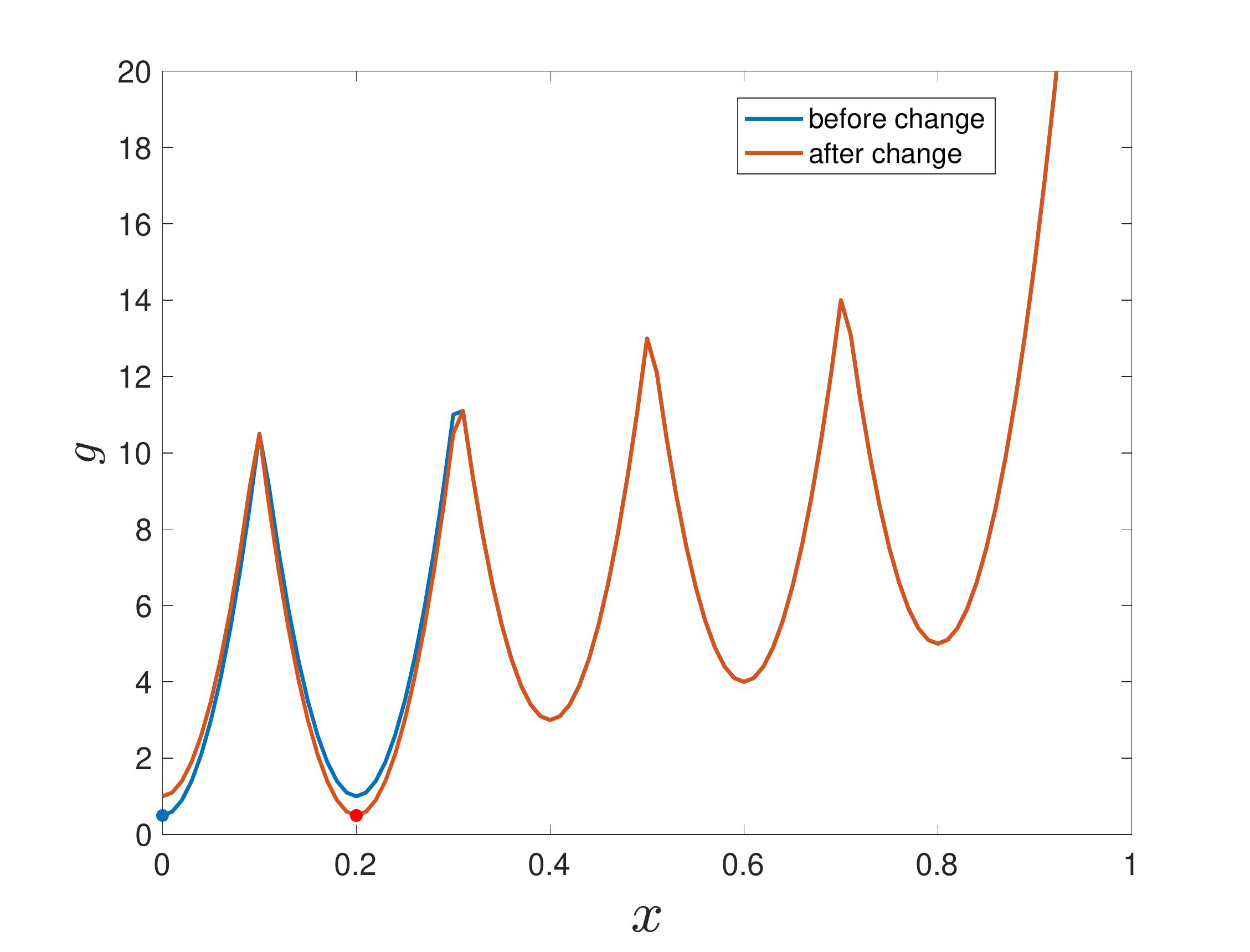}\\[-2mm]	
	\caption{Illustration of a deceptive scenario. Closed dots in blue and red are the global minima before and after an environmental change, respectively. A part of basins of attraction varies whereas the rest remains unchanged before and after a change.}
	\label{fig:sdp7_g}
\end{figure}

There are two ways to increase the difficulty of change detection. One way is to change the global basin of attraction of the previous environment into a local one of the new environment. This can be done by defining $g({\bf x},t)$ as
\begin{equation}
g_{multi}=\frac{1}{n\!\!-\!\!M\!\!+\!\!1}\!\!\sum\nolimits_{i=M}^{n}{\underset{k=1,\dots,5}{\min}\{h_k\!\!+\!\!10(10x_i\!\!-\!\!y_k)^2\},}
\label{eq:sdp7_g}
\end{equation}
where $h_k$ ($k=1,\dots,5$) defines 5 local minima, one of which is global. $y_k$ defines the location of the local minima. A recommendation is $h_k=k$ and $y_k=2(k-1)$. The global optima is set to $h_{k_t}=0.5$ with $k_t=\lceil 5rnd_t\rceil$, at time $t$. \cref{fig:sdp7_g} shows deceptive changes, which will leave undetected if detectors are in the unchanged basins of attraction (this happens if not all population members reach the global basin of attractions).

\begin{figure}[thb]
	\centering
	\begin{tabular}{cc}
		$t=0$ & $t=0.1$\\
		\includegraphics[width=0.5\linewidth]{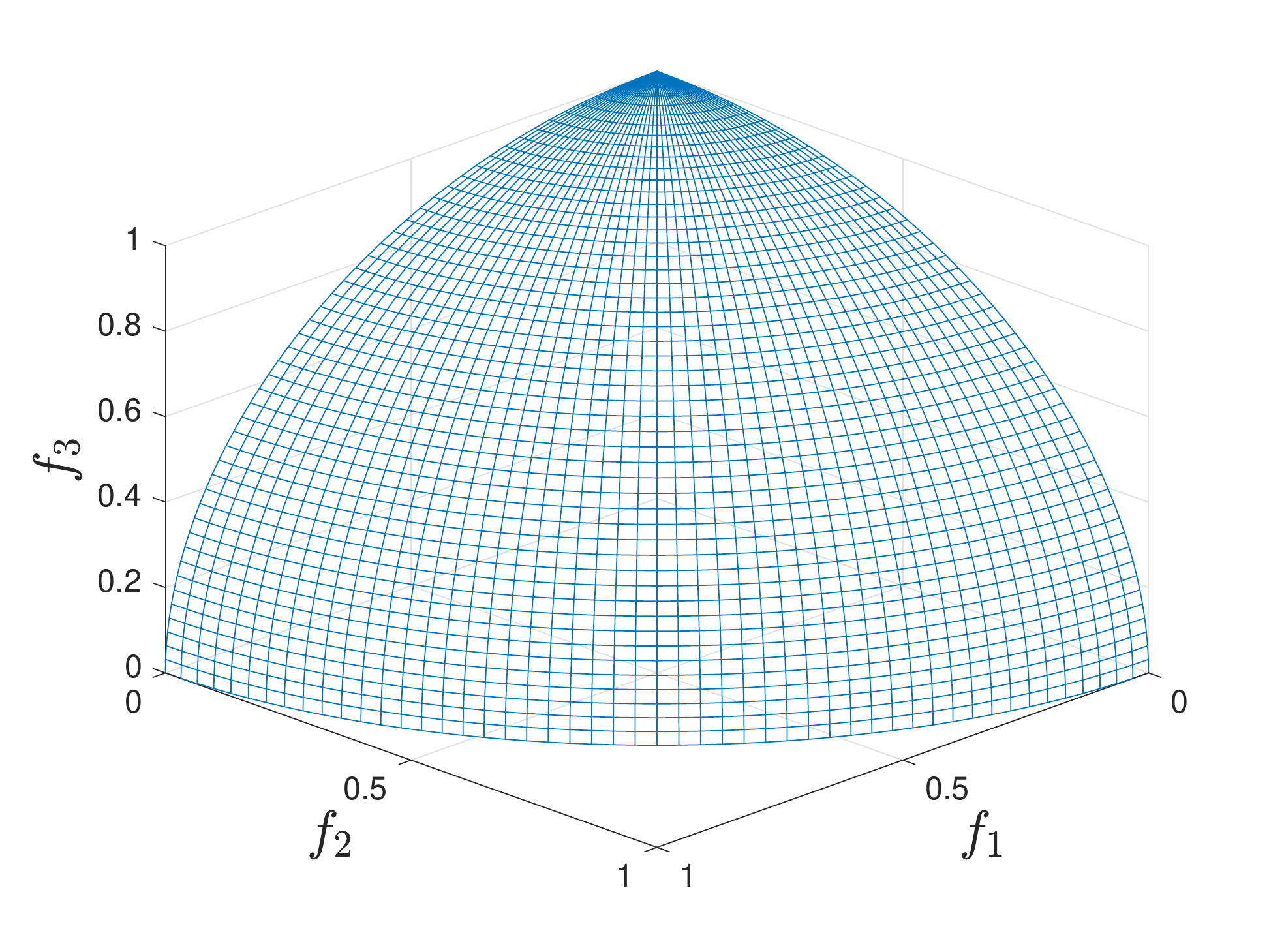}  &\includegraphics[width=0.5\linewidth]{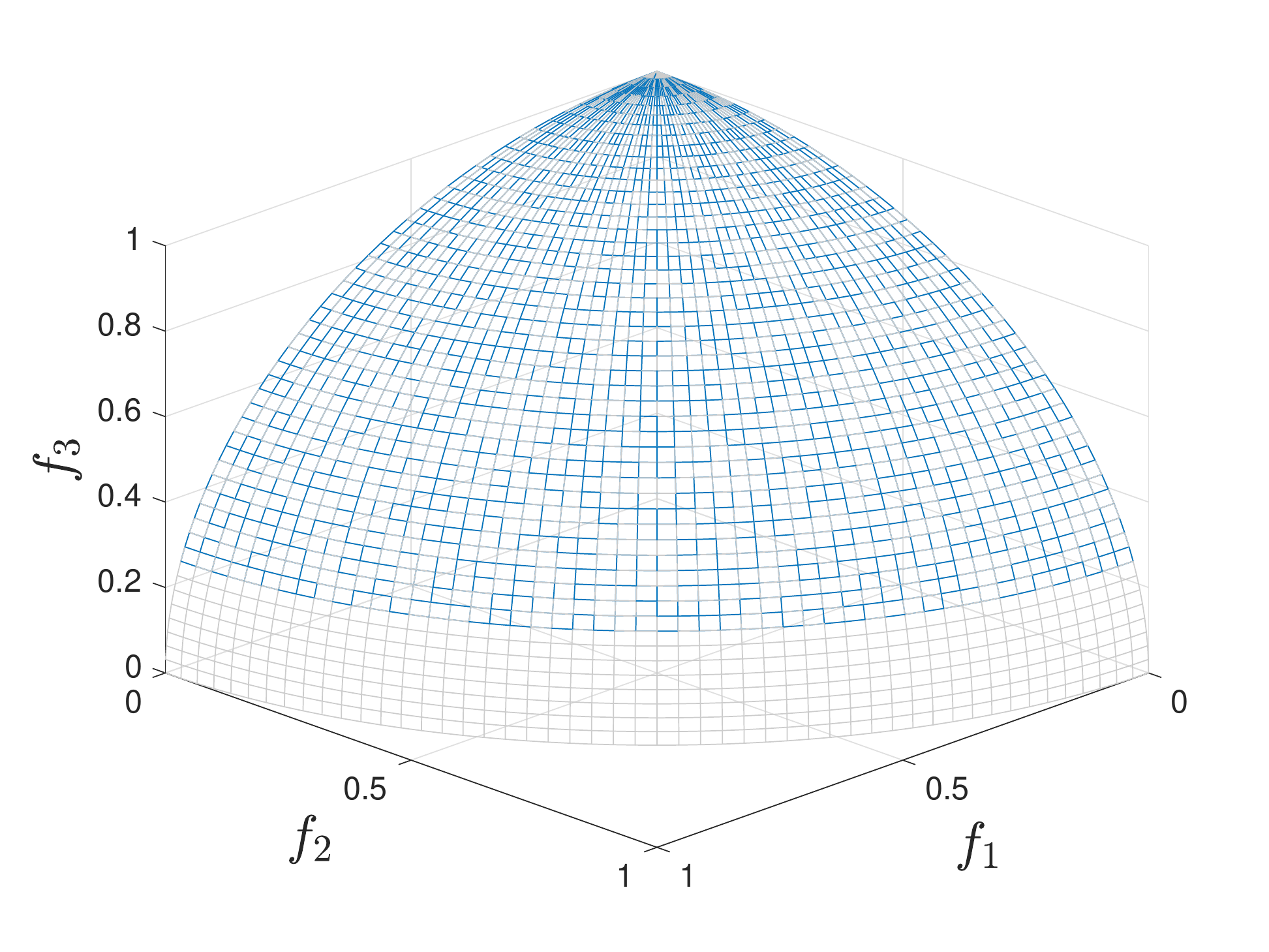}\\[-2mm]
		$t=0.2$ & $t=0.3$\\
		\includegraphics[width=0.5\linewidth]{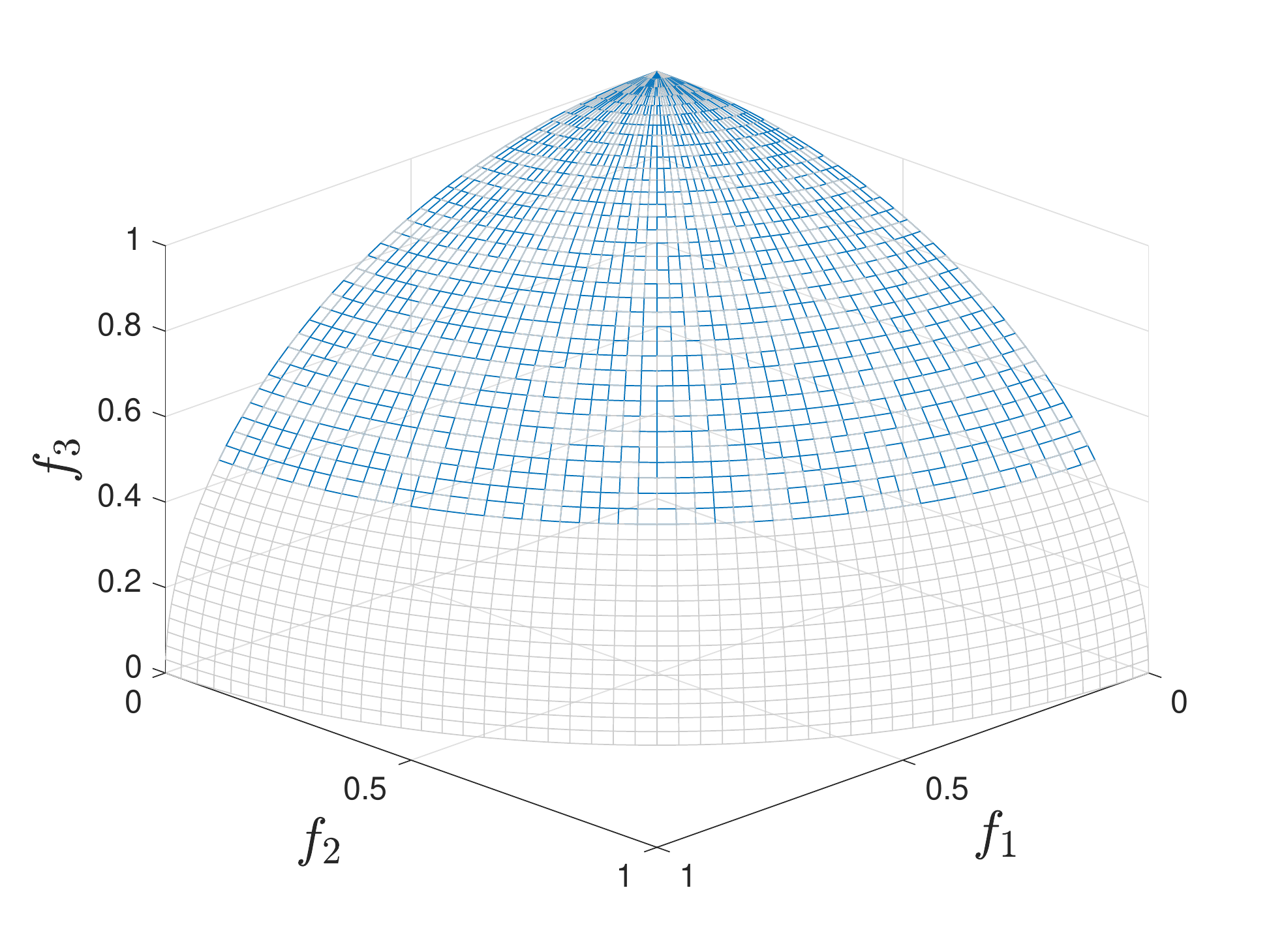}  & 
		\includegraphics[width=0.5\linewidth]{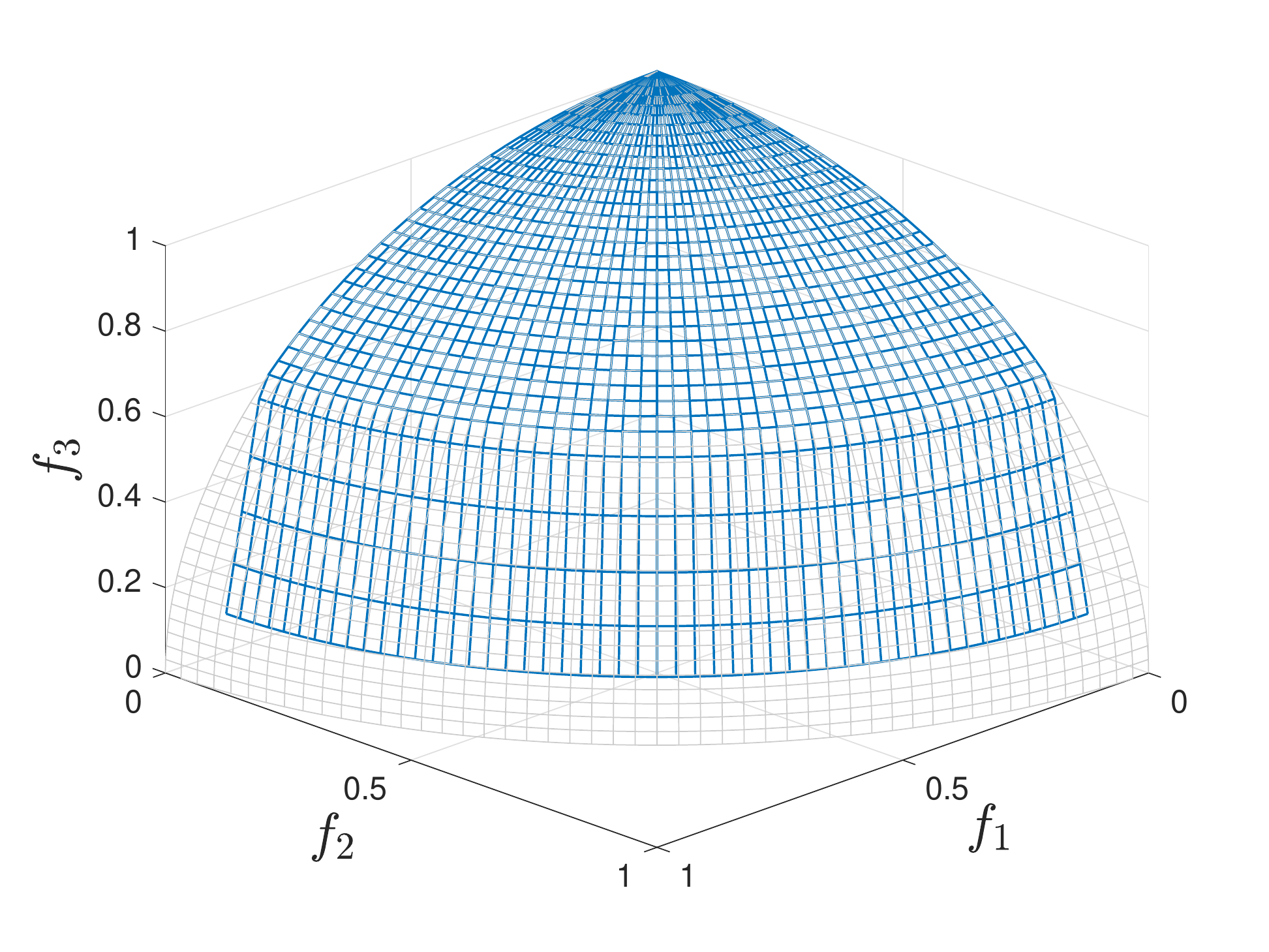}\\
	\end{tabular}
	\caption{Illustration of changes that occur on a part of the PF at different time steps and can be difficult to detect.}
	\label{fig:sdp6_pf0}
\end{figure}

Another way to create a less detectable scenario is to have a time-varying search subspace that causes objective variation. As a result, only detectors falling within this subspace can detect environmental changes. Let $\mu_{detect,i}=\mu_{sphere,i}$ for  $i=2,\dots,M$ and its first objective $\mu_{1}({\bf x},t)$ be
\begin{equation}
\mu_{detect,1}\!\!=\!\!\begin{cases}
\begin{aligned}
&\!\!|k_t(\cos(0.5 \pi x_1)\!\!-\!\!\cos(0.5\pi \alpha_t))\\[-1.5mm]
&~~ +\sin(0.5 \pi \alpha_t)|
\end{aligned}  &\!\!\!\!\text{if } x_1\leq \alpha_t\\
\!\!\sin(0.5\pi x_1),&\!\!\!\!\text{otherwise}
\end{cases}
\label{eq:sdp6_pf}
\end{equation}
\noindent where $\alpha_t \in (0,1)$ defines a subspace $[0, \alpha_t]$, where $x_1$ renders dynamic variation of the $M$-th objective, and $k_t$ is a parameter controlling the severity of change regarding the PF. Environmental changes on the PF can be detected only when detectors have its $x_1$ smaller than $\alpha_t$. We recommend the use of $\alpha_t=0.5|\sin(\pi t)|$ and $k_t=10\cos(2.5\pi t)$. 
\cref{fig:sdp6_pf0} illustrates the change of the PF derived from (\ref{eq:sdp6_pf}).

\vspace{1mm}
\noindent\emph{{4) PS randomness}}
\vspace{1mm} 

Existing DMOPs favour prediction-based EAs, because the PS changes in regular pattern in dynamic environments. Here, we present a PS (distance-related variables $x_i$) that changes in a random manner:
\begin{equation}
\bar{x}_i^t\!\!=\!\!\begin{cases}
\bar{x}_i^{t\!-\!1}\!+\!5(rnd_t\!-\!0.5)\sin(0.5\pi t) & \text{if } 0\!\! \leq \!\!\bar{x}_{i,t} \!\!\leq\!\! 1 \\
rnd_t & \text{otherwise}
\end{cases}
\label{eq:sdp1_ps}
\end{equation}
where $rnt_t$ is a random value from [0,1) and $\bar{x}_i^t$ is the optimal value of $x_i \in [0,1]$ at time $t$ and $\bar{x}_i^0=i/n$ is recommended. Note that (\ref{eq:sdp1_ps}) can be scaled to any range restriction of $x_i$.

\vspace{1mm}
\noindent\emph{{5) Dynamic number of objectives/variables}}
\vspace{1mm} 
\setcounter{paragraph}{0}

\paragraph{Dynamic number of variables}
Let $n_t$ be the number of variables at time $t$, the simplest way to make $n_t$ change over time is probably to pick $n_t$ randomly from a range $[n_l, n_u]$: 
\begin{equation}
n_t=n_l+\lfloor rnd_t(n_u-n_l) \rfloor
\label{eq:sdp12_nt}.
\end{equation}
Then the $n_t-M+1$ distance-related variables are used to build $g({\bf x},t)$, e.g., $g({\bf x},t)=\sum_{i=M}^{n_t}x_i^2$.

\paragraph{Dynamic number of objectives}
Let $M_t$ be the number of objectives at time $t$, it can can be defined similarly to $n_t$:
\begin{equation}
M_t=M_l+\lfloor rnd_t(M_u-M_l) \rfloor
\label{eq:sdp13_Mt},
\end{equation}
which means $M_t$ is randomly generated from a range $[M_l,M_u]$. Correspondingly, the $M_t$-dimensional PF can be constructed as
\begin{equation}
\mu_{dobj,i}=\sin(y_i)\prod\nolimits_{k=1}^{i-1}{\cos(y_k)}
\label{eq:sdp13_pf},
\end{equation}
where $y_k=\pi (x_k+1)/6$ for $k=1,\dots,M_t$. Thus, the PF always satisfies $\sum_{i=1}^{M_t\!-\!1}{\mu_i^2}+4\mu_{M_t}^2=1$ for any $M_t$.

\vspace{1mm}
\noindent\emph{{6) Degeneration}}
\vspace{1mm} 
\setcounter{paragraph}{0}

Time-varying PF degeneration can be achieved by specifying first a degeneration level $d_t$ at time $t$, and then constructing the $d_t$-D PF. The following presents two degenerate examples.

\begin{figure}[t]
	\begin{tabular}{cc}
		\includegraphics[width=0.5\linewidth]{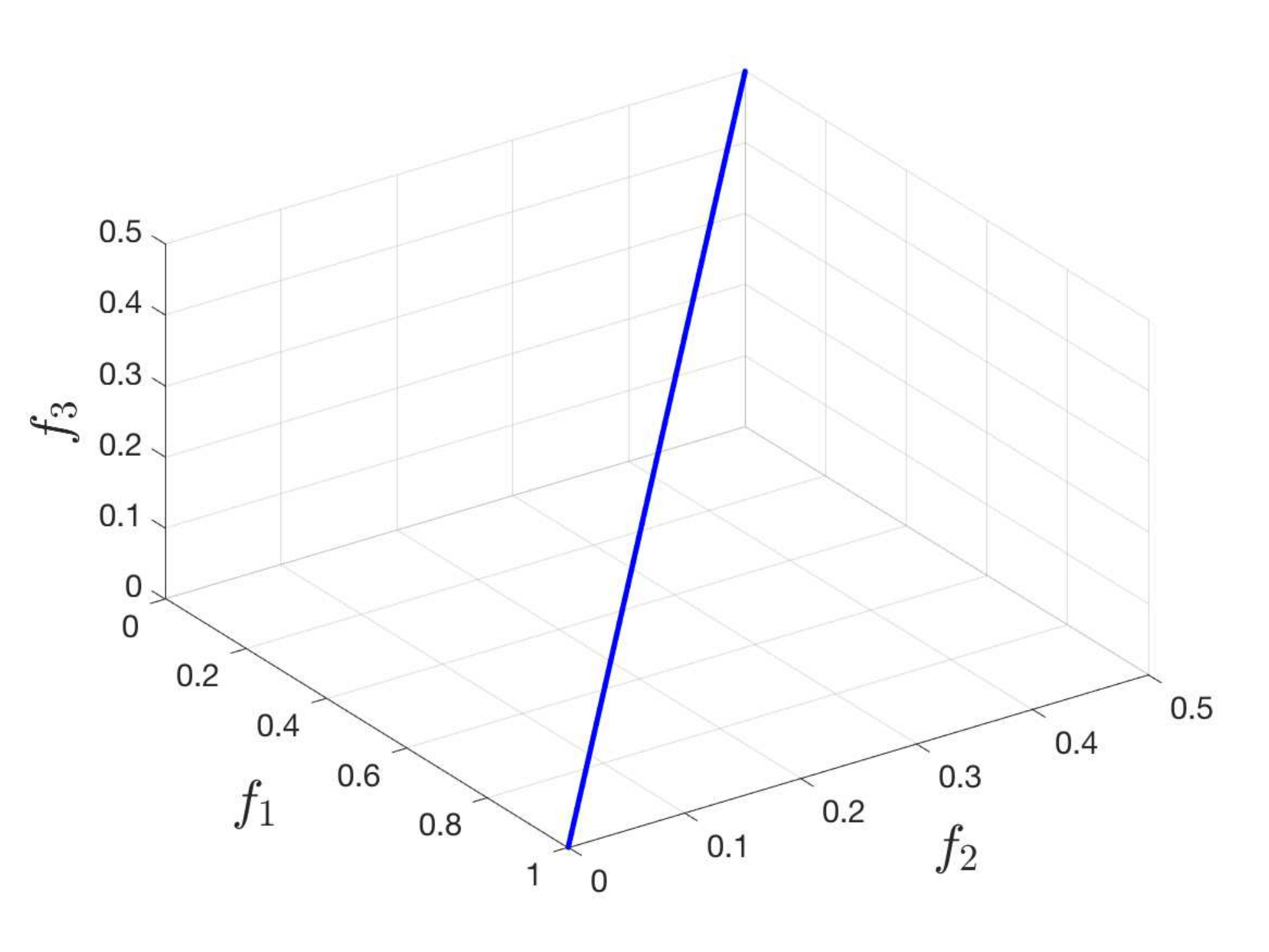}  & 
		\includegraphics[width=0.5\linewidth]{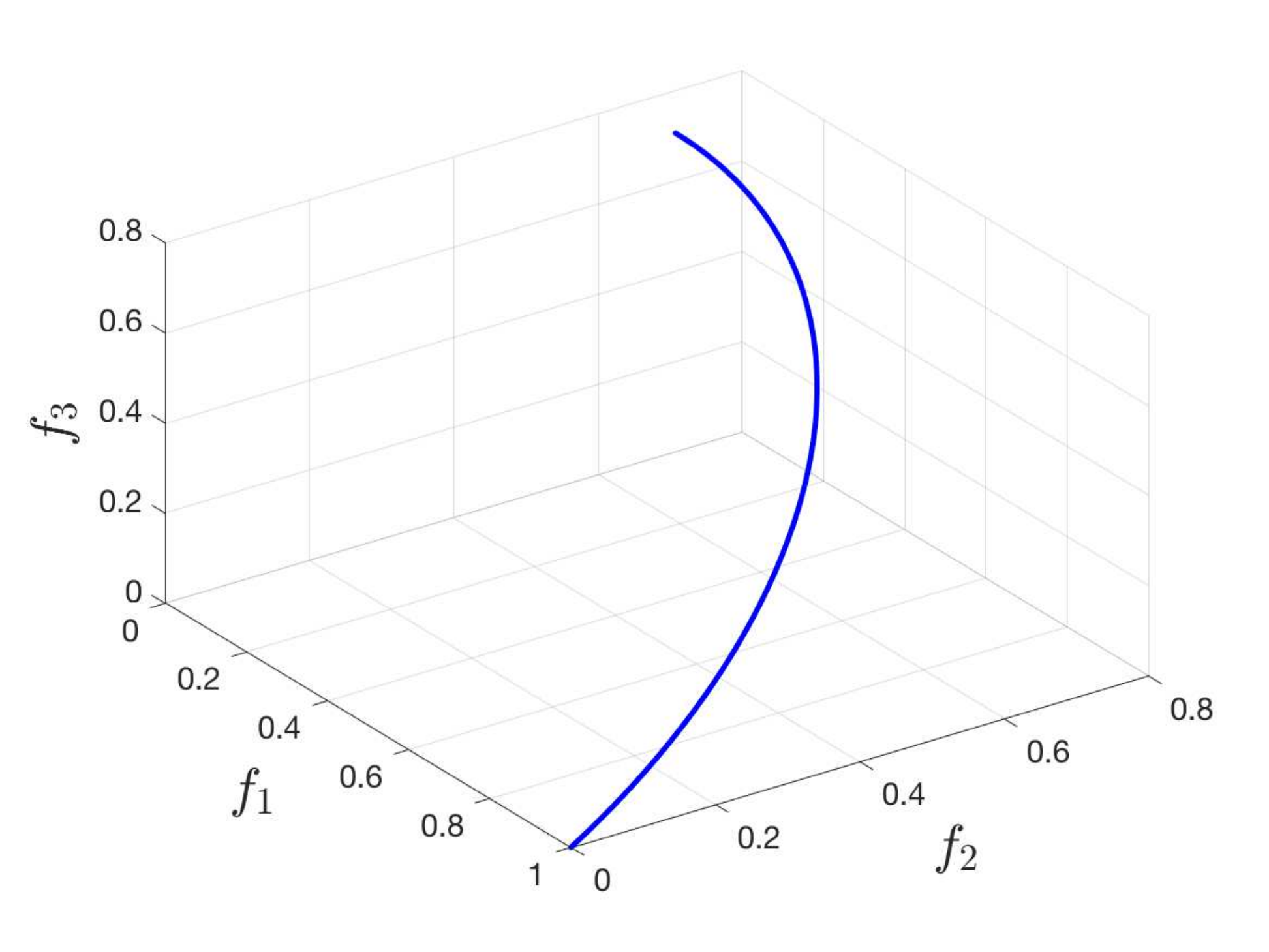}\\
		(a) linear & (b) spherical
	\end{tabular}
	\caption{Linear and spherical degenerate PFs of three objectives.}
	\label{fig:deg}
\end{figure}

\paragraph{Linear degenerate PF} This is based on the simplex-shape PF of (\ref{eq:sdp_linear}). It is defined as 
\begin{equation}
\mu_{ldeg,i}\!\!=\!\! \begin{cases}
(1+g-y_{i})\prod_{j=1}^{i-1}y_j &\text{if $1 \!\!\leq\!\! i \!<\!\! M$,}\\
y_1\dots y_{M-2}y_{M-1} &\text{if $i=M$.}
\end{cases}
\label{eq:sdp14_pf}
\end{equation}
with 
\begin{equation*}
y_i=
\begin{cases}
x_i, & \text{for } 1 \leq i \leq d_t\\
0.5+x_i|\sin(0.5\pi t)|g, & \text{for } d_t <i \leq M-1
\end{cases}
\end{equation*}
where the $g$ function is already defined in (\ref{eq:sdp}). It indicates that the last $M\!-\!d_t$ objectives are positively correlated. 
\cref{fig:deg}(a) shows a linear degenerate PF of $M=3$ and $d_t=1$. 

\paragraph{Spherical degenerate PF} This requires the sphere-shape PF. It is defined as
\begin{equation}
\mu_{sdeg,i}\!\!=\!\! \begin{cases}
\sin(y_{i})\prod_{j=1}^{i-1}\cos(y_j) &\text{if $1 \!\!\leq\!\! i \!<\!\! M$,}\\
\prod_{j=1}^{M-1}\cos(y_i) &\text{if $i=M$.}
\end{cases}
\label{eq:sdp15_pf}
\end{equation}
with 
\begin{equation*}
y_{k}=
\begin{cases}
0.5\pi x_k, & \text{for } 1 \leq i \leq d_t\\[-1.5mm]
\arccos(\frac{0.5\sqrt{2}}{1+x_k|\sin(0.5\pi t)|g}), & \text{for } d_t \!<\! i \leq M\!-\!1
\end{cases}
\end{equation*}
where $k=(p_t\!+\!i\!-\!1) \text{ mod } (M\!-\!1)\!+\!1$ with $p_t$ randomly chosen from $[1,M\!-\!1]$. This also results in $M\!-\!d_t$ correlated objectives whose indices are determined by $p_t$. Thus, the spherical degenerate PF is more difficult than the linear one. 
\cref{fig:deg}(b) shows a spherical degenerate PF of $M=3$ ($d_t=1$ and $p_t=2$).

\vspace{1mm}
\noindent\emph{{7) PF shrinkage/expansion}}
\vspace{1mm} 
\setcounter{paragraph}{0}

\begin{figure}[t]
	\centering
	\includegraphics[width=0.65\linewidth]{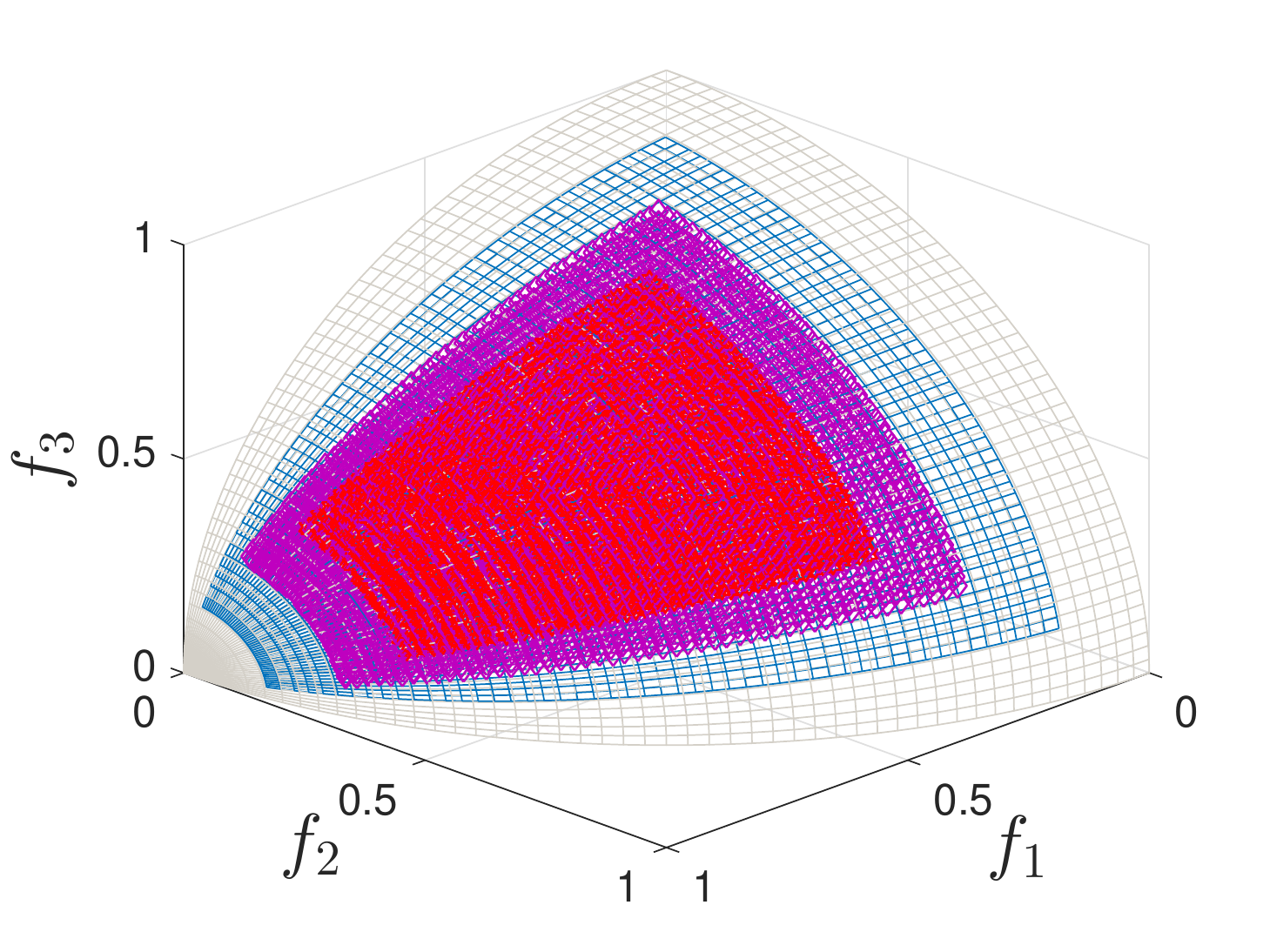}\\[-2mm]	
	\caption{Illustration of PF shrinkage over time: $t=0$ (grey), $t=0.1$ (blue), $t=0.2$ (purple), and $t=0.3$ (red).}
	\label{fig:sdp5_pf0}
	\vspace{-4mm}
\end{figure}

Another scenario we would like to consider is the shrinkage or expansion of the PF over time. Here, we create this scenario based on the spherical PF of (\ref{eq:sdp15_pf}), in which $y_i$ is defined as
\begin{equation}
y_{i}=\frac{\pi}{6}G_t+(\frac{\pi}{2}-\frac{\pi}{3}G_t)x_i,
\label{eq:sdp_exsh}
\end{equation}
where $G_t=|\sin(0.5\pi t)|$. 
\cref{fig:sdp5_pf0} shows the scenario of PF shrinkage over time.

\vspace{1mm}
\noindent\emph{{8) Search favourability}}
\vspace{1mm} 
\setcounter{paragraph}{0}

\begin{figure}[t]
	\begin{tabular}{cc}
		\includegraphics[width=0.5\linewidth]{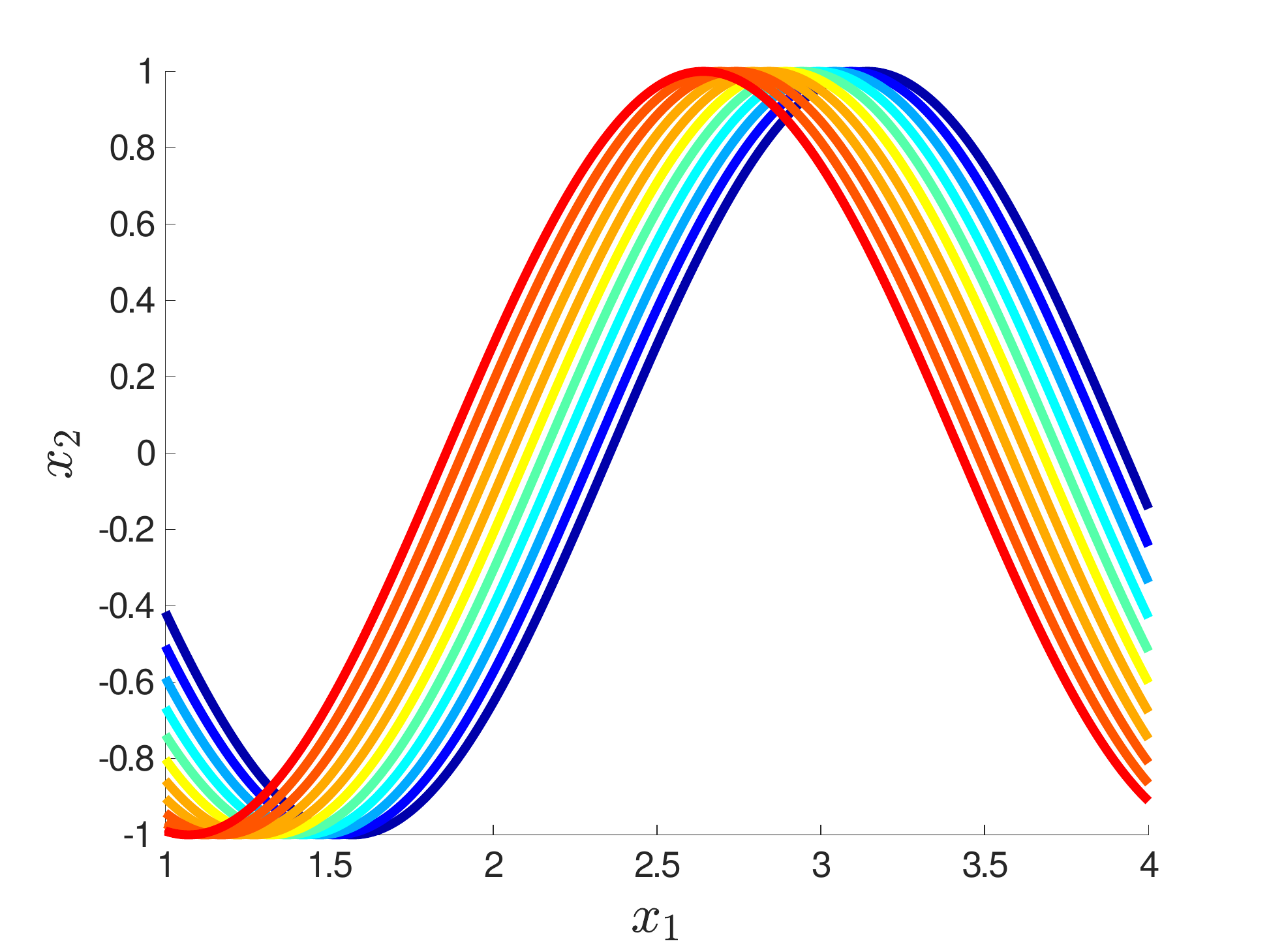}  & 
		\includegraphics[width=0.5\linewidth]{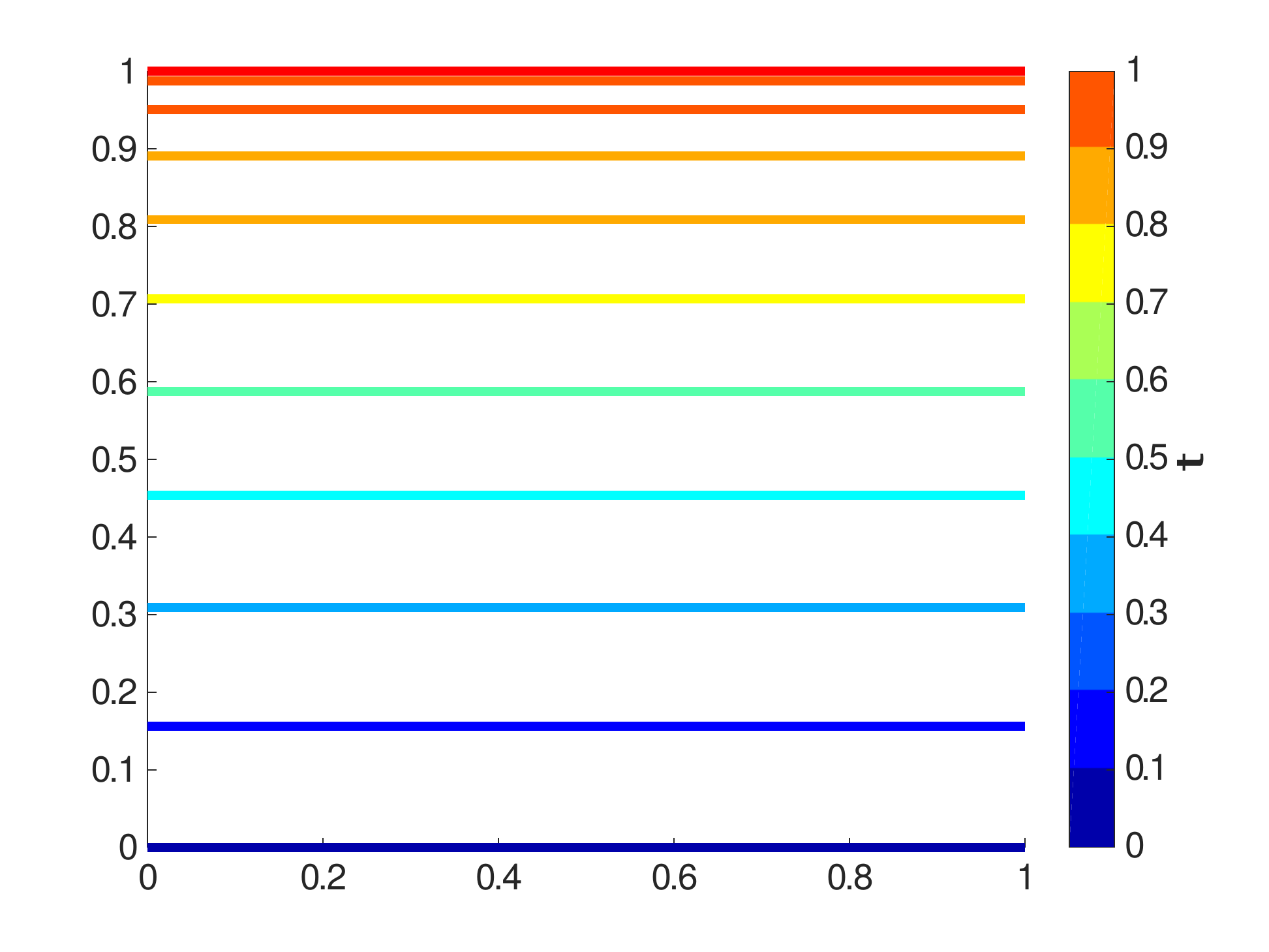}\\
		(a) PS of SDP2 & (b) PS of FDA1
	\end{tabular}
	\caption{PS comparison between SDP2 and FDA1 \cite{Fari_04:1}. The PS centroid and other PS regions for FDA1 move in the same direction (time steps shown in different colours), which is not the case with SDP2.}
	\label{fig:compPs}
	\vspace{-4mm}
\end{figure}
Often than not, EAs favour PF regions that are easy to approximate. It is interesting to create a dynamic scenario where favoured PF regions change over time and investigate its effect on EAs. Consider the following $g$ function
\begin{equation}
g_{favour}=
\begin{cases}
\sum_{i=M}^{n}{(-0.9p_i^2+|p_i|^{0.6})}, & x \in \Phi({\bf x},t),\\
p_i^2, & \text{otherwise.}
\end{cases}
\end{equation}
where $p_i\!\!=\!\!x_i\!-\!|\sin(0.5 \pi t)|$ and $\Phi({\bf x},t)$ is a time-varying subspace for which barriers exist around the global optima. This means the PF region corresponding to $\Phi({\bf x},t)$ is much unfavoured than other regions. Here, we recommend $\Phi({\bf x},t)\!\!=\!\!\{x\in [0,1]^{n}~|~3t\!-\!\lfloor 3t \rfloor \! \leq\! \sum_{i=1}^{M\!-\!1}x_{i} \leq 3t\!+\!0.2\!-\!\lfloor 3t\!+\!0.2 \rfloor \}$.

\subsection{Test Suite}
A total of 15 scalable dynamic problems (SDP) are instanced using the above-mentioned component functions to imitate a number of representative and well-known real-world features. These are briefly tabulated in Table~\ref{tab:SDP}, and a detailed description of SDP can be found in the supplementary material. 

\newcommand{\tabincell}[2]{\begin{tabular}{@{}#1@{}}#2\end{tabular}}

\begin{table*}[t]
	\centering
	\caption{Definition of SDP problems. PF and variable ranges can be adjusted based on \cite{HHBW06}. Also, $x_i \in {\bf x_I}$ is replaced by $x_i^{K(t)}$ where ${K(t)}=\exp(10\sin(0.5\pi t))$ if dynamics on diversity are concerned}
	\begin{tabular}{l|l|l|l}
		\hline
		Prob.  & Definition & Prob. & Definition\\
		\hline
		
		{SDP1}  & \tabincell{l}{
			$f_{i=1:M}({\bf x},t)=(1+g)\mu_{\mathsmaller{\mathsmaller{\prod}},i}$\\ 
			$g=\sum_{i=M\!+\!1}^{n}{\!\!(x_i-\bar{x}_{i,t})^2}$~ $\triangleright$ $\bar{x}_{i,t}$ updated by (\ref{eq:sdp1_ps})\\
			search space: $1 \leq x_{i=1:M} \leq 4, 0 \leq x_{i=M+1:n} \leq 1$}
		& {SDP9} & \tabincell{l}{$f_{i=1:M}({\bf x},t)=(1+g\cdot \max(0,i-M-1))\mu_{disc1,i}+G(t)$\\
			$g=\sum_{i=M}^{n}{(x_i-\frac{1}{\pi}{|\arctan(\cot(3\pi t^2))|})^2}$\\
			$G_t=|\sin(0.5\pi t)|$, ~~~
			search space: $0 \leq x_{i=1:n} \leq 1$} 
		\\\hline
		
		{SDP2}  & \tabincell{l}{
			$f_{i=1:M-1}({{\bf x}},t)=(1+g)\mu_{\mathsmaller{\mathsmaller{\sum}},i}$ \\
			$g=\sin(\frac{\pi}{8}x_1)\sum_{i=M}^{n}{(x_i-\cos(t+2x_1))^2}$\\
			search space: $1 \leq x_{i=1:M-1} \leq 4, -1 \leq x_{i=M:n} \leq 1$}
		& {SDP10}  & \tabincell{l}{
			$f_{i=1:M}({\bf x},i)=(1+g\cdot \max(0,i-M-1))\mu_{disc2,i}$ \\
			$g=\sum_{i=M}^{n}{(x_i-\sin(x_1+0.5 \pi t))^2}$\\	
			search space: $0 \leq x_{i=1:M-1} \leq 1, -1 \leq x_{i=M:n} \leq 1$}
		\\\hline
		
		{SDP3}  & \tabincell{l}{	
			$f_{i=1:M}({\bf x},t)=(1+g)\mu_{knee,i}$ ~~$y_i=x_i\!-\!\cos(t)$\\
			$k_t=\lfloor5|\sin(\pi t)|\rfloor$, ~~$g=\sum_{i=M}^{n}{\!\!(4y_i^2\!-\!\cos(2 k_t\pi y_i)+1)}$\\	
			search space: $0 \leq x_{i=1:M-1} \leq 1, -1 \leq x_{i=M:n} \leq 1$}
		& {SDP11} & \tabincell{l}{
			$f_{i=1:M}({\bf x},t)=(1+g_{favour})\mu_{sphere,i}$\\
			search space: $0 \leq x_{i=1:n} \leq 1$
		}
		\\\hline
		
		{SDP4}  & \tabincell{l}{
			$f_{i=1:M}({\bf x},t)=(1+g \cdot \max(0,i-M-1))\mu_{mix,i}$\\
			$g=\sum_{i=M}^{n}{(x_i-\cos(t+x_1+x_{i-1}))^2}$\\
			search space: $0 \leq x_{i=1:M-1} \leq 1, -1 \leq x_{i=M:n} \leq 1$\\} 
		& {SDP12}  & \tabincell{l}{
			$f_{i=1:M-1}({\bf x},t)=(1+g)(1-x_{i})\mu_{linear,i}$\\
			$g=\sum_{i=M}^{n_t}{(x_i-\sin(n_t t)\sin(2\pi x_1))^2}$~ $\triangleright$ $n_t$ from (\ref{eq:sdp12_nt})\\	
			search space: $0 \leq x_{i=1:M-1} \leq 1, -1 \leq x_{i=M:n_t} \leq 1$
		}
		\\\hline
		
		{SDP5}  & \tabincell{l}{
			$f_{i=1:M}({\bf x},t)=(1+g)\mu_{sphere,M\!-\!i\!+\!1}$~ $\triangleright$ $y_i$ from (\ref{eq:sdp_exsh})\\
			$g=G_t+\sum_{i=M}^{n}{(x_i-0.5G_t x_1)^2}, G_t=|\sin(0.5\pi t)|$\\
			search space: $0 \leq x_{i=1:n} \leq 1$}
		& {SDP13}  & \tabincell{l}{
			$f_{i=1:M_t}({\bf x},t)=(1+g)\mu_{dobj,i}$ ~$\triangleright$$M_t$ from (\ref{eq:sdp13_Mt})\\
			$g=\sum_{i=M\!+\!1}^{i=n}{(x_i-\frac{it}{M+it})^2}$\\	
			search space: $0 \leq x_{i=1:n} \leq 1$ 
		} 
		\\\hline
		
		{SDP6}  & \tabincell{l}{
			$f_{i=1:M}({\bf x},t)=(1+g)\mu_{detect,M-i+1}$\\		
			$g=\sum_{i=M}^{n}{(x_i-0.5)^2(1+|\cos(8\pi x_i)|)}$\\
			search space: $0 \leq x_{i=1:n} \leq 1$}
		& {SDP14}  & \tabincell{l}{
			$f_{i=1:M-1}({\bf x},t)=(1+g)\mu_{ldeg,i}$\\
			$d_t=1+\lfloor|(M-2)\cos(0.5\pi t)|\rfloor$, $g=\sum_{i=M}^{n}{(x_i-0.5)^2}$\\
			search space: $0 \leq x_{i=1:n} \leq 1$
		} 
		\\\hline
		
		{SDP7}  & \tabincell{l}{
			$f_{i=1:M}({\bf x},t)=(0.5+g_{multi})\mu_{linear,i}$\\	
			search space: $0 \leq x_{i=1:n} \leq 1$	}
		& {SDP15} & \tabincell{l}{
			$f_{i=1:M-1}({\bf x},t)=(1+g)^i\mu_{sdeg,i}$\\
			$d_t=rndi^1_t(1,M-1)$, $p_t=rndi^2_t(1,M-1)$ ~$\triangleright$ $p_t$ for (\ref{eq:sdp15_pf})
			\\$g=\sum_{i=M}^{n}{(x_i-\frac{d_t}{M-1})^2}$ ~~
			search space: $0 \leq x_{i=1:n} \leq 1$
		}
		\\\hline
		
		{SDP8} & \multicolumn{3}{l}{$f_{i=1:M}({\bf x},t)=(1+g)\mu_{sphere,M-i+1}$, ~~ $g=\sum_{i=M}^{n}{(x_i-\sin(t x_1))^2} +g_{nsc}$, ~~search space: $0 \leq x_{i=1:M} \leq 1, -1 \leq  x_{i=M:n} \leq 1$}
		\\\hline
		
		\multicolumn{4}{l}{The time $t$ is $\frac{1}{n_t} \lfloor\frac{\tau}{\tau_t} \rfloor$, where $n_t$, $\tau_t$, $\tau$ represent the severity of change, frequency of change, and iteration counter, respectively.}\\
		\multicolumn{4}{l}{$rndi_t(a,b)$ generates a random integer from $[a,b]$ at time $t$. sgn($\cdot$) denotes the sign function. $\triangleright$ starts a comment.}
		\label{tab:sdf}
	\end{tabular}%
	\label{tab:SDP}
	\vspace{-2mm}
\end{table*}

Note that, SDP uses different $g$ functions, capturing a variety of PS shapes.
For example, there are several problems (e.g. SDP2) for which the PS centroid and some PS regions move in distinct directions in the event of changes (most existing problems do not have this feature, see Fig.~\ref{fig:compPs} for details). This may be a new challenge for recently popular centroid-based prediction algorithms \cite{ZJZ14,Zhou2015}. Prediction algorithms may encounter difficulties when solving problems like SDP1, since the PS varies in a random manner. The PS shapes for all the problems could be made as complicated as LZ problems \cite{LZ09}, but such complexity is not, and also should not be the main focus of DMOPs,  as its existence shadows the importance of landscape dynamics \cite{Gee2017}.

\begin{table*}[th]
	\addtolength{\tabcolsep}{-3.5pt}
	\centering
	\caption{Comparison between SDP and existing DMO problems}
	\begin{tabular}{l|c|c|c|c|c|c|c|c|c|c|c|l}
		\hline
		{Test Suite} & {Scalability} & \tabincell{l}{Multi-\\modality} 		
		& \tabincell{l}{Variable\\linkage} & \tabincell{l}{Irregular PF\\boundary} & \tabincell{l}{PS unpre-\\dictability}&  \tabincell{l}{Multi-\\knee} & \tabincell{l}{Search fa-\\vourability} & \tabincell{l}{Degen-\\eration} & \tabincell{l}{PF con-\\nectivity} & \tabincell{l}{dynamic\\$M$/$n$}& \tabincell{l}{Detectability\\of change} &{Origination} \\\hline
		
		\rowcolor{LightCyan}
		\tabincell{l}{SDP}  & \cmark & \cmark & \cmark & \cmark & \cmark & \cmark & \cmark & \cmark & {s,ns,d} & {M,n}  & \cmark &  {new} \\\hline
		
		\tabincell{l}{FDA \cite{Fari_04:1}}  & \xmark & \xmark & \xmark & \xmark & \xmark & \xmark & \xmark & \xmark & {s} & \xmark  & \xmark   & \tabincell{l}{ZDT/DTLZ} \\\hline	
		
		\tabincell{l}{DIMP \cite{KGT10}}  & \xmark & \xmark & \xmark & \xmark & \xmark & \xmark & \xmark & \xmark & {s,d} & \xmark  & \xmark   & \tabincell{l}{ZDT} \\\hline
		
		\tabincell{l}{DSW \cite{MRW06}}  & \xmark & \xmark & \xmark & \xmark & \xmark & \xmark & \xmark & \xmark & {s} & \xmark  & \xmark   & \tabincell{l}{} \\\hline
		
		\tabincell{l}{dMOP \cite{GT09}}  & \xmark & \xmark & \xmark & \xmark & \xmark & \xmark & \xmark & \xmark & {s} & \xmark  & \xmark   & \tabincell{l}{ZDT} \\\hline	
		
		\tabincell{l}{T \cite{HSA11}}  & \cmark & \cmark & \xmark & \xmark & \xmark & \xmark & \xmark & \xmark & {s} & {M,n}  & \xmark   & \tabincell{l}{DTLZ} \\\hline	
		
		\tabincell{l}{ZJZ \cite{ZJZ14}}  & \xmark & \xmark & \cmark & \xmark & \xmark & \xmark & \xmark & \xmark & {s} & \xmark  & \xmark   & \tabincell{l}{FDA/LZ} \\\hline
		
		\tabincell{l}{HE \cite{HE14}}  & \xmark & \cmark & \cmark & \xmark & \xmark & \xmark & \xmark & \xmark & {s,d} & \xmark  & \xmark   & \tabincell{l}{FDA/LZ/WFG} \\\hline
		
		\tabincell{l}{UDF \cite{BDSC14}}  & \xmark & \cmark & \xmark & \xmark & \cmark & \xmark & \xmark & \xmark & {s,d} & \xmark  & \xmark   & \tabincell{l}{ZDT} \\\hline	
		
		\tabincell{l}{JY \cite{Jiang2016_benchmark}}  & \xmark & \cmark & \cmark & \xmark & \xmark & \cmark & \xmark & \xmark & {s,d} & \xmark  & \xmark  & {} \\\hline
		
		\tabincell{l}{GTA \cite{Gee2017}}  & \xmark & \cmark & \cmark & \xmark & \xmark & \xmark & \xmark & \cmark & {s,d} & \xmark  & \xmark   & \tabincell{l}{ZDT/DTLZ} \\\hline	
		
		\tabincell{l}{CLY\cite{Chen2017}}  & \cmark & \cmark & \xmark & \xmark & \xmark & \xmark & \xmark & \xmark & {s} & {M}  & \xmark &{T/DTLZ}\\\hline
		
		\multicolumn{13}{l}{`s', `ns', or `s' in the column of PF connectivity denote DMOPs exist having a simply-connected, non-simply connected, or disconnected PF, respectively.}\\
		\multicolumn{13}{l}{`M' or `n' in column of `dynamic M/n' means DMOPs exist having a time-varying number of objectives or variables, respectively.}
		\label{tab:comptestsuite}
	\end{tabular}%
	\label{tab:compDMOPs}
	\vspace{-5mm}
\end{table*}

Table \ref{tab:compDMOPs} provides a comparison between SDP and existing DMOPs. For each property, `$\cmark$' (`$\xmark$') represents the presence (absence) of this property for the corresponding test suite. In other words, environmental changes in those problems can be detected with one re-evaluation of a random population member. As can be seen, multimodality and variable linkages have gained most attention, and SDP embraces a variety of rarely studied but important characteristics. The source code of SDP is available at {\color{blue}\url{http://homepages.cs.ncl.ac.uk/shouyong.jiang/}}.

\vspace{-3mm}
\subsection{Links between SDP and Applications}
SDP captures a number of real-world features. Examples include but not limited to the following applications:
Electromagnetic micromirrors \cite{Pieri2018}, crop growth \cite{Ramirez2012}, 
railway rescheduling \cite{Eaton2017}, dynamic subset sum \cite{Rohlfshagen2010}, speed reducer model \cite{Zhang2012-cmm}, pressure vessel model \cite{Zhang2012-cmm}, greenhouse control \cite{Zhang08}, predictive control \cite{Butans2011}, weapon selection \& planning \cite{Xiong2017}, PID control \cite{Fari_04:1}, unstable plant control \cite{HSA11}, self-paced learning \cite{Li2016}, spacecraft equipment layout \cite{Lau2014}, distance minimisation \cite{Zille2015}, job shop scheduling \cite{Shen2015}, engine calibration \cite{Lygoe2013}, and water distribution systems \cite{Rahmani2015}. 

\begin{figure}[t]
	\centering
	\includegraphics[width=1.1\linewidth]{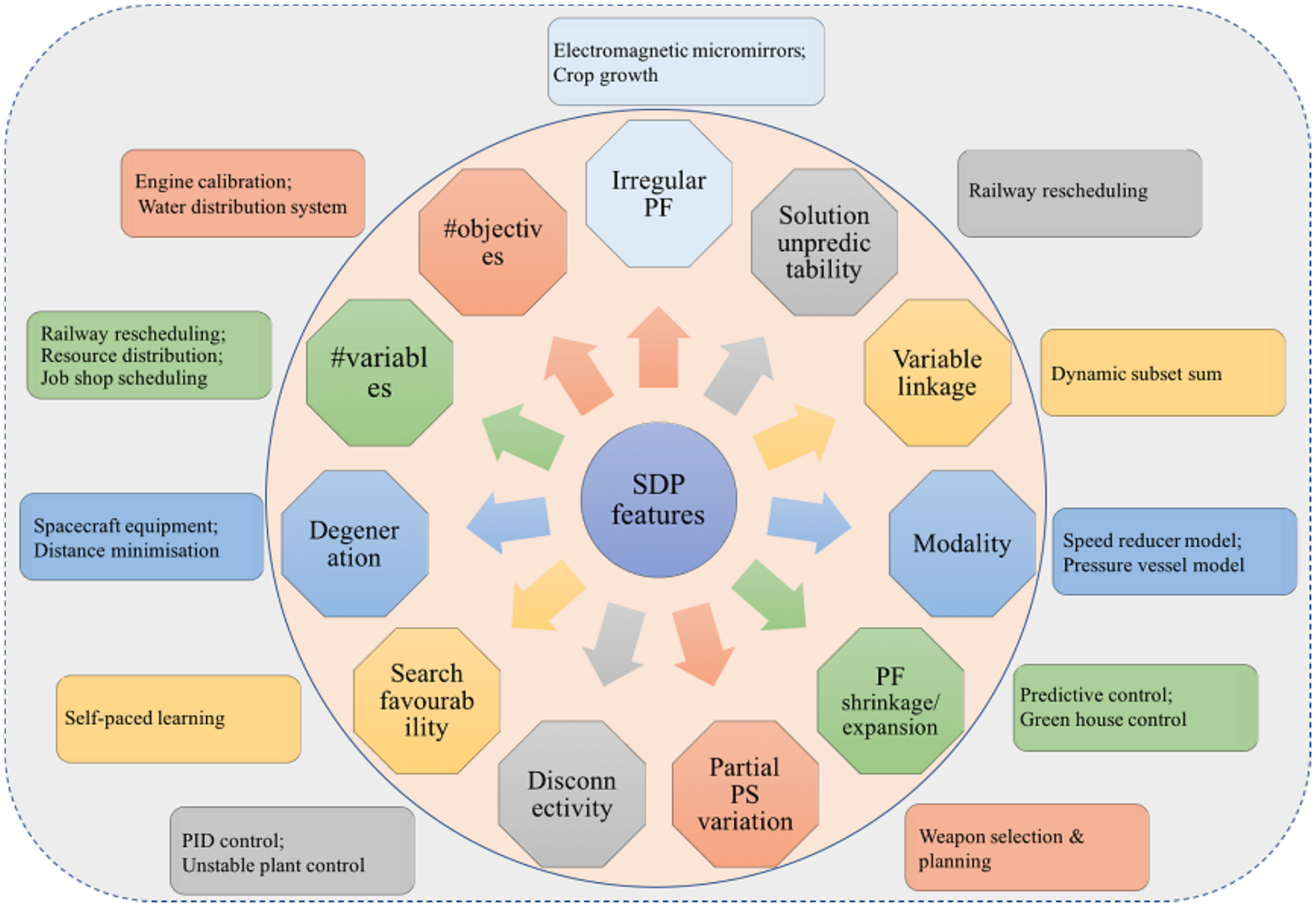}\\[-2mm]	
	\caption{Links between SDP features and real-world applications.}
	\label{fig:link}
	\vspace{-4mm}
\end{figure}

Fig.~\ref{fig:link} briefly shows the links between the above applications and SDP features. Please refer to the supplementary material for more details on how they are linked to each other.

The benchmarks have a few limitations due to the fact that real applications are highly complex and diverse, and there are still a lot features that are not well characterised. Thus, it is difficult to develop a perfect test suite that can model and reflect well all possible real-world features. Despite that, constructing some test scenarios based on well characterised features is of great importance for the analysis of EAs. The proposed SDP test suite simulates a number of real-world features in a simplified manner from the literature. The simplification somehow reduces realisticity but is still helpful for understanding of strengths and weaknesses of algorithms. Also, some of the test cases may serve as adversarial examples for DMOP solvers.

\section{Experimental Studies}
\subsection{DMO Algorithms}
Five different DMO algorithms are employed for studying the proposed benchmarks, each representing a type of EAs. They are 1) the Pareto dominace based DNSGA-II\cite{DRK07}; the decomposition-based MOEA/D \cite{ZL07}; the multipopulation based dCOEA \cite{GT09}; the population prediction based PPS \cite{ZJZ14}; (5) and the steady-generational SGEA \cite{Jiang17_SGEA}. In this paper, PPS with regularity model \cite{Zhang2008rm} (PPS+RM2) and nondominated sorting (PPS+NS) are both analysed. All the algorithms are allowed to use 10\% of the population as change detectors in every generation. Each algorithm has a way to respond to the detected environmental changes. To be more specific, DNSGA-II hypermutates a portion of population members. MOEA/D re-evaluates the population. PPS records history search information and predicts the new location of the PF/PS. dCOEA increases population diversity by retrieving its memory population whereas SGEA relocates a portion of population that have high density. More detailed information can be referred to the original papers. 

In the experiment, 2, 3 and 5 objectives were tested for SDP1-12. The number of objectives varied from 2 to 5 for SDP13, and was fixed to 5 for SDP14-15 to simulate PF degeneration. Since all the algorithms except MOEA/D were originally designed for multiobjective cases, their performance may degrade vastly for the 5-objective case, a well-recognised issue in many-objective optimisation \cite{Jiang17_SPEAR}. Here, such algorithms were incorporated with the shift-based density estimator (SDE) \cite{LYL14-SDE} to alleviate this scalability issue. The number of variables ($n$) for all SDP instances except SDP12 was 10.  For SDP12, the lower bound $n_l$and upper bound $n_u$ were 10 and 20, respectively. The severity and frequency of change was set to $n_t\!\!=\!\!10$  and $\tau_t\!\!=\!\!10$, respectively. All the algorithms with a population size of 100 (note that, the same number of weight vectors was generated for MOEA/D) were executed 30 runs for each SDP instance. Each execution was terminated after $30\tau_t\!+\!50$ generations, where the first environmental change occurs after 50 generations.
\vspace{-3mm}
\subsection{DMO Performance Measures}
The paper uses the following three popular performance measures for algorithm analysis.
\subsubsection{Mean inverted Generational Distance (MIGD)}
MIGD \cite{Jiang2016_benchmark} is adapted from IGD \cite{Zhang2008rm}, a static performance indicator that measures both the convergence and diversity of solutions found by an algorithm. The reference set consists of around 1000 uniformly sampled points from the PF.
\subsubsection{Mean Hypervolume Difference (MHVD)}
The MHVD \cite{Jiang17_SGEA} is a modification of the static measure HVD\cite{ZJZST07} that computes the gap between the hypervolume of the obtained PF and that of the true PF. The reference point for the computation of $M$-dimensional hypervolume is $(z_1+0.5, z_2+0.5, \cdots, z_M+0.5)$, where $z_j$ is the maximum value of the $j$-th objective of the true PF.
\subsubsection{Mean Detection Timeliness (MDT)}
MDT is proposed to measure how timely an environmental change is detected, which is defined as: 
\begin{equation}
	MDT=\frac{1}{T}\sum\nolimits_{i=1}^{T}DT_t
\end{equation}
with
\begin{equation}
DT_t=
\begin{cases}
\frac{Eval_t-1}{n_d \tau_t-1} & \text{if change detected,}\\
1 & \text{otherwise,}
\end{cases}
\end{equation}
where $Eval_t$ is the number fitness re-evaluations to detect a change. $n_d$ is the number of candidate detectors in every generation. The smaller $DT_t$, the earlier a change detected. The earlier the change detected, the more time left for algorithms to react to the change. This new measure is expected to facilitate detectability analysis of some SDP problems, i.e. SDP6-7.

\begin{table*}[tbp]
	\footnotesize
	\addtolength{\tabcolsep}{-4pt}
	\renewcommand{\arraystretch}{-0.2}
	\centering
	\caption{Mean and standard deviation values of MIGD obtained for the 15 SDP instances (Best values are highlighted in boldface)}
	\begin{tabular}{llllllll}
		\toprule
		Prob. & M     & \multicolumn{1}{c}{SGEA} & \multicolumn{1}{c}{MOEA/D} & \multicolumn{1}{c}{DNSGA-II} & \multicolumn{1}{c}{dCOEA} & \multicolumn{1}{c}{PPS+RM2} & \multicolumn{1}{c}{PPS+NS} \\
		\midrule
		\multirow{3}[0]{*}{SDP1} & 2     & 9.3261E-2(1.5004E-2) & 6.2156E-1(5.5887E-2) & 4.1969E-1(4.8711E-2) & \textbf{8.4615E-2(1.5374E-2)} & 2.5867E-1(1.8990E-2) & 2.0789E-1(3.0608E-2) \\
		& 3     & 4.3905E-1(4.7661E-2) & 7.6216E-1(5.2187E-2) & 7.2701E-1(5.1762E-2) & \textbf{1.4974E-1(4.9059E-3)} & 3.1697E-1(9.3400E-3) & 3.3883E-1(1.9306E-2) \\
		& 5     & 7.8830E-1(4.8308E-2) & 1.0694E+0(4.8052E-2) & 1.1462E+0(6.0372E-2) & \textbf{5.4438E-1(6.6552E-3)} & 6.0859E-1(1.1731E-2) & 7.9588E-1(3.6405E-2) \\ [0.6mm]
		\hline \\[0.6mm]
		\multirow{3}[0]{*}{SDP2} & 2     & 8.6043E-1(1.1621E-1) & 8.1100E-1(5.1193E-2) & 6.8745E-1(1.0307E-1) & 5.1928E-1(6.9548E-2) & \textbf{2.5903E-2(2.2241E-3)} & 7.2711E-1(6.6190E-2) \\
		& 3     & 1.3053E+0(6.1254E-2) & 1.3564E+0(4.8000E-2) & 1.2079E+0(7.2509E-2) & 1.7822E+0(2.1026E-1) & \textbf{1.0311E+0(1.0054E-2)} & 1.2531E+0(5.5772E-2) \\
		& 5     & 3.2826E+0(2.1455E-1) & 3.3515E+0(1.2682E-1) & 3.1491E+0(1.3555E-1) & 3.0590E+0(1.2614E-1) & \textbf{2.9072E+0(1.1269E-1)} & 3.1111E+0(1.1649E-1)  \\ [0.6mm]
		\hline \\[0.6mm]
		\multirow{3}[0]{*}{SDP3} & 2     & \textbf{2.5206E-1(3.9477E-2)} & 1.2454E+0(1.5892E-1) & 7.0913E+0(1.8304E+0) & 3.8346E-1(4.8950E-2) & 2.7499E+0(3.7374E-1) & 1.2471E+0(3.4526E-1) \\
		& 3     & 3.4950E-1(3.1703E-2) & 1.0354E+0(2.5927E-1) & 6.8759E+0(1.4236E+0) & \textbf{3.0616E-1(4.0351E-2)} & 2.0703E+0(2.4872E-1) & 1.1363E+0(2.1795E-1) \\
		& 5     & 7.9270E-1(8.3751E-2) & 7.0601E-1(7.9973E-2) & 6.6870E+0(9.9340E-1) & \textbf{3.1347E-1(2.5673E-2)} & 4.4833E+0(4.2683E-1) & 2.7619E+0(3.1562E-1)  \\ [0.6mm]
		\hline \\[0.6mm]
		\multirow{3}[0]{*}{SDP4} & 2     & 4.2917E-2(6.8911E-3) & 6.2079E-2(2.9912E-3) & 5.7338E-2(1.8803E-2) & 2.7497E-1(2.1211E-2) & \textbf{3.6684E-2(9.8533E-3)} & 7.3563E-2(1.1994E-2) \\
		& 3     & \textbf{1.7443E-1(9.7604E-3)} & 3.0083E-1(2.3235E-2) & 1.9266E-1(8.3212E-3) & 3.0893E-1(1.8741E-2) & \textbf{1.7506E-1(1.1775E-2)} & 1.9601E-1(5.1456E-2) \\
		& 5     & 1.3461E+0(2.2638E-2) & 1.2264E+0(6.9346E-2) & \textbf{1.0224E+0(1.9923E-2)} & 1.3248E+0(1.9636E-2) & 1.1373E+0(2.6283E-2) & \textbf{1.0706E+0(5.3471E-2)} \\ [0.6mm]
		\hline \\[0.6mm]
		\multirow{3}[0]{*}{SDP5} & 2     & 7.7841E-2(8.3963E-3) & 4.4920E-2(4.5506E-3) & 7.6415E-2(5.6092E-3) & 9.5632E-2(9.3391E-3) & \textbf{1.3263E-2(7.6399E-3)} & 1.4875E-1(6.7328E-2) \\
		& 3     & 9.7171E-2(2.5169E-3) & 1.0186E-1(3.1057E-3) & 9.7959E-2(1.5543E-3) & 1.1189E-1(6.0643E-3) & \textbf{7.6826E-2(2.5474E-3)} & 1.8331E-1(3.2985E-2) \\
		& 5     & \textbf{2.2225E-1(3.5064E-3)} & 3.0259E-1(5.3612E-3) & 2.5540E-1(4.6899E-3) & 2.3456E-1(2.2762E-3) & 3.0122E-1(9.3411E-3) & 2.5659E-1(4.9830E-3)  \\ [0.6mm]
		\hline \\[0.6mm]
		\multirow{3}[0]{*}{SDP6} & 2     & \textbf{2.1138E-2(2.2306E-3)} & 3.6280E-2(3.3572E-3) & \textbf{2.0604E-2(1.7274E-3)} & 3.8971E-2(4.1611E-3) & 3.5673E-2(1.9470E-2) & 2.5695E-2(3.0682E-3) \\
		& 3     & 7.5464E-2(1.6515E-3) & 9.5727E-2(6.2962E-3) & \textbf{7.2154E-2(9.5353E-4)} & 7.7151E-2(2.9435E-3) & 1.0151E-1(4.2321E-3) & 7.8461E-2(4.4459E-3) \\
		& 5     & 1.6138E-1(2.5653E-3) & 2.2895E-1(1.2371E-2) & 2.0986E-1(5.5562E-3) & \textbf{1.5694E-1(4.8012E-3)} & 2.6286E-1(1.1650E-2) & 2.4659E-1(1.5094E-2)  \\ [0.6mm]
		\hline \\[0.6mm]
		\multirow{3}[0]{*}{SDP7} & 2     & 2.9198E-1(3.8628E-2) & 3.5146E-1(4.9663E-2) & 2.8187E-1(2.6422E-2) & \textbf{2.1163E-1(2.3197E-2)} & 1.7686E+0(6.6517E-2) & 3.2446E-1(6.0296E-2) \\
		& 3     & 2.4117E-1(2.0602E-2) & 3.2199E-1(6.5497E-2) & 2.6487E-1(2.5710E-2) & \textbf{2.0020E-1(1.7453E-2)} & 1.4098E+0(4.7409E-2) & 3.4077E-1(5.2185E-2) \\
		& 5     & 2.6467E-1(1.3462E-2) & 3.6265E-1(5.0484E-2) & 1.7475E+0(1.8353E-1) & \textbf{2.5931E-1(2.3388E-2)} & 1.8269E+0(4.6239E-2) & 1.5925E+0(9.8340E-2) \\ [0.6mm]
		\hline \\[0.6mm]
		\multirow{3}[0]{*}{SDP8} & 2     & 2.1374E-1(9.1497E-3) & 2.4222E-1(9.0841E-3) & 2.2207E-1(8.6575E-3) & 2.0508E-1(5.4325E-3) & \textbf{1.9401E-1(1.5567E-3)} & 2.0814E-1(1.1178E-2) \\
		& 3     & \textbf{3.5877E-1(5.3062E-3)} & 4.8510E-1(1.6154E-2) & 4.0433E-1(6.8888E-3) & 3.6180E-1(5.4997E-3) & 4.3083E-1(4.9985E-3) & 3.8212E-1(4.5574E-3) \\
		& 5     & 9.4780E-1(1.2979E-2) & 1.2480E+0(2.0159E-2) & 1.1082E+0(3.5435E-2) & \textbf{7.2180E-1(1.3229E-2)} & 9.4551E-1(1.2909E-2) & 9.2950E-1(1.5506E-2) \\ [0.6mm]
		\hline \\[0.6mm]
		\multirow{3}[0]{*}{SDP9} & 2     & \textbf{1.0632E-1(1.7644E-2)} & 1.6641E-1(1.9159E-2) & 1.4244E-1(1.6680E-2) & 3.3963E-1(3.6917E-2) & 1.4570E-1(1.4506E-2) & 2.0100E-1(9.3230E-2) \\
		& 3     & 3.1771E-1(8.2711E-2) & 2.4645E-1(8.2593E-2) & 2.1103E-1(2.7109E-2) & 3.0319E-1(3.0942E-2) & \textbf{1.2553E-1(1.0021E-2)} & 2.3610E-1(2.7317E-2) \\
		& 5     & \textbf{3.0103E-1(3.7656E-2)} & \textbf{3.0861E-1(2.0241E-2)} & 5.4918E-1(3.8125E-2) & 3.2919E-1(3.0595E-2) & 3.7375E-1(1.2328E-2) & 5.4869E-1(2.7953E-2) \\ [0.6mm]
		\hline \\[0.6mm]
		\multirow{3}[0]{*}{SDP10} & 2     & 1.6685E-1(4.0694E-3) & 2.9397E-1(6.9083E-2) &1.6911E-1(8.4562E-3) & 3.4644E-1(3.6425E-2) &  \textbf{9.1859E-2(3.1605E-2)} & 1.7205E-1(8.6120E-2) \\
		& 3     & 3.5954E-1(3.4616E-2) & 4.2109E-1(7.0327E-2) & 3.8218E-1(1.9520E-2) & 4.2340E-1(2.8644E-2) & \textbf{1.2764E-1(1.0371E-2)} & 3.1060E-1(6.9885E-2) \\
		& 5     & \textbf{5.7639E-1(5.7568E-2)} & 6.4993E-1(3.0252E-2) & 7.7668E-1(4.3188E-2) & 6.5982E-1(2.9976E-2) & 6.1715E-1(4.4344E-2) & 6.9704E-1(1.6013E-1)  \\ [0.6mm]
		\hline \\[0.6mm]
		\multirow{3}[0]{*}{SDP11} & 2     & 6.4371E-2(1.6809E-3) & 2.6178E-1(2.7305E-2) & 1.5095E-1(9.6704E-3) & 6.0314E-2(2.5296E-3) & \textbf{4.4515E-2(3.1305E-5)} & \textbf{4.5203E-2(2.3995E-4)} \\
		& 3     & 1.9079E-1(1.0090E-2) & 3.6762E-1(3.2448E-2) & 3.6157E-1(2.6587E-2) & \textbf{1.1329E-1(2.5548E-3)} & 1.7449E-1(7.2012E-3) & 1.5676E-1(7.3411E-3) \\
		& 5     & 4.0221E-1(2.5269E-2) & 3.4661E-1(1.5415E-2) & 6.3570E-1(7.3994E-2) & \textbf{1.9136E-1(8.4729E-3)} & 4.3770E-1(1.8185E-2) & 3.5915E-1(2.8904E-2)  \\ [0.6mm]
		\hline \\[0.6mm]
		\multirow{3}[0]{*}{SDP12} & 2     & 1.9170E-1(6.2410E-2) & 5.2264E-1(1.5950E-1) & 4.4837E-1(1.0061E-1) & \textbf{8.5506E-2(3.8072E-2)} & 5.5428E-1(4.6491E-2) & 1.2531E-1(3.8519E-2) \\
		& 3     & 4.4132E-1(6.5372E-2) & 4.5028E-1(5.6600E-2) & 7.2564E-1(1.1774E-1) & \textbf{1.5165E-1(6.8493E-3)} & 5.2718E-1(2.6685E-2) & 2.7889E-1(2.4520E-2) \\
		& 5     & 3.9855E-1(4.3436E-2) & 4.2115E-1(5.2160E-2) & 1.4175E+0(1.7749E-1) & \textbf{1.8283E-1(7.5622E-3)} & 6.7036E-1(5.1655E-2) & 8.1231E-1(5.4737E-2)  \\[0.6mm]
		\hline \\[0.6mm]
		SDP13 & 5     & 8.4061E-2(7.9721E-3) & 1.0180E-1(6.1239E-3) & 8.4798E-2(7.0145E-3) & \textbf{7.0975E-2(1.0061E-2)} & 9.5572E-2(2.9776E-3) & 9.6748E-2(4.2568E-3)  \\ [0.6mm]
		\hline \\[0.6mm]
		SDP14 & 5     & \textbf{6.5487E-2(4.3806E-3)} & 1.0630E-1(7.6648E-3) & \textbf{6.4276E-2(1.0656E-3)} & 7.1703E-2(1.9505E-3) & 8.4592E-2(1.5736E-3) & 7.6039E-2(3.2563E-3)  \\ [0.6mm]
		\hline \\[0.6mm]
		SDP15 & 5     & 1.0257E+0(2.9432E-1) & 1.0275E+0(1.7245E-1) & 1.2973E+0(2.8247E-1) & \textbf{3.5952E-1(6.4377E-3)} & 1.0731E+0(2.5774E-1) & 2.0513E+0(1.1109E+0) \\
		\hline
	\end{tabular}%
	\label{tab:migd1}%
\end{table*}%
\begin{table*}[t]
	\footnotesize
	\addtolength{\tabcolsep}{-4pt}
	\renewcommand{\arraystretch}{-0.2}
	\centering
	\caption{Mean and standard deviation values of MHVD obtained for the 15 SDP instances (Best values are highlighted in boldface)}
	\begin{tabular}{llllllll}
		\toprule
		Prob. & M     & \multicolumn{1}{c}{SGEA} & \multicolumn{1}{c}{MOEA/D} & \multicolumn{1}{c}{DNSGA-II} & \multicolumn{1}{c}{dCOEA} & \multicolumn{1}{c}{PPS+RM2} & \multicolumn{1}{c}{PPS+NS} \\
		\midrule
		\multirow{3}[0]{*}{SDP1} & 2     & 5.2670E-1(8.4645E-2) & 3.7076E+0(2.8571E-1) & 2.3853E+0(2.6906E-1) & \textbf{4.5425E-1(8.1736E-2)} & 1.6651E+0(1.0725E-1) & 1.2369E+0(1.6458E-1) \\
		& 3     & 1.0081E+1(1.1146E+0) & 1.8618E+1(1.1888E+0) & 1.5581E+1(1.3468E+0) & \textbf{2.1401E+0(2.9449E-1)} & 9.3530E+0(2.8352E-1) & 7.8706E+0(4.3013E-1) \\
		& 5     & 3.4308E+2(1.9660E+1) & 3.7717E+2(2.5780E+1) & 4.2600E+2(2.9409E+1) & \textbf{8.6496E+1(4.3172E+0)} & 2.4362E+2(7.1802E+0) & 2.6235E+2(1.4253E+1) \\[0.6mm]
		\hline \\[0.6mm]
		\multirow{3}[0]{*}{SDP2} & 2     & 1.4214E+0(1.4505E-1) & 1.3381E+0(8.4062E-2) & 1.1815E+0(1.5437E-1) & 1.0472E+0(9.9469E-2) & \textbf{7.8860E-2(6.5308E-3)} & 1.2609E+0(1.0338E-1) \\
		& 3     & \textbf{1.7937E+1(2.8299E+0)} & 1.8971E+1(2.2038E+0) & 1.8674E+1(3.1659E+0) & 2.1636E+1(2.6174E+0) & 2.9704E+1(1.1565E+0) & 1.9138E+1(1.7765E+0) \\
		& 5     & 4.8964E+4(4.1439E+3) & 5.1299E+4(1.7184E+3) & 4.8501E+4(2.3572E+3) & \textbf{4.3268E+4(2.7646E+3)} & 5.1660E+4(2.5962E+3) & 5.0428E+4(2.8583E+3)  \\[0.6mm]
		\hline \\[0.6mm]
		\multirow{3}[0]{*}{SDP3} & 2     & \textbf{4.6648E-1(5.3196E-2)} & 1.2608E+0(4.8253E-2) & 1.4479E+0(4.0328E-2) & 6.8189E-1(6.2141E-2) & 1.4802E+0(9.5330E-2) & 1.1990E+0(9.6029E-2) \\
		& 3     & 8.7498E-1(1.0329E-1) & 2.0218E+0(1.4425E-1) & 2.6696E+0(3.3988E-2) & \textbf{7.4566E-1(9.7200E-2)} & 2.5846E+0(1.3062E-1) & 2.1027E+0(1.7571E-1) \\
		& 5     & \textbf{3.4385E+0(2.7773E-1)} & 3.7021E+0(2.1595E-1) & 7.4506E+0(9.6280E-2) & 8.2845E-1(1.3288E-1) & 7.3242E+0(1.0500E-1) & 6.8406E+0(3.7360E-1)  \\[0.6mm]
		\hline \\[0.6mm]
		\multirow{3}[0]{*}{SDP4} & 2     & 8.1477E-2(1.3599E-2) & 1.4387E-1(8.6390E-3) & 1.1231E-1(3.8207E-2) & 4.8753E-1(3.3632E-2) & \textbf{6.9047E-2(2.0264E-2)} & 1.5118E-1(2.7351E-2) \\
		& 3     & \textbf{1.4874E-1(1.9171E-2)} & 2.3408E-1(2.2798E-2) & 1.9692E-1(1.7906E-2) & 3.8331E-1(2.3196E-2) & 2.9492E-1(3.1545E-2) & 2.8471E-1(5.5207E-2) \\
		& 5     & \textbf{1.2488E+0(8.0639E-2)} & 1.3055E+0(2.5871E-2) & 2.2199E+0(2.0760E-1) & 1.5170E+0(6.2289E-2) & 2.6974E+0(2.8236E-1) & 1.7558E+0(8.0618E-2)  \\[0.6mm]
		\hline \\[0.6mm]
		\multirow{3}[0]{*}{SDP5} & 2     & 2.3832E-1(2.4182E-2) & 1.4220E-1(1.7169E-2) & 2.3023E-1(1.9446E-2) & 2.6834E-1(1.7748E-2) & \textbf{3.4563E-2(2.1645E-2)} & 3.3865E-1(9.8831E-2) \\
		& 3     & 4.7111E-1(2.0059E-2) & \textbf{3.4290E-1(2.1790E-2)} & 3.6919E-1(1.0846E-2) & 6.0938E-1(3.4418E-2) & 3.5609E-1(2.3444E-2) & 9.8601E-1(1.3767E-1) \\
		& 5     & 2.8730E+0(1.1463E-1) & \textbf{2.3685E+0(1.3710E-1)} & 3.0259E+0(1.0498E-1) & 3.9472E+0(1.1764E-1) & 5.9986E+0(1.3350E-1) & 4.5930E+0(1.7151E-1)  \\[0.6mm]
		\hline \\[0.6mm]
		\multirow{3}[0]{*}{SDP6} & 2     & \textbf{1.2993E-1(8.8157E-3)} & 1.4749E-1(1.4536E-2) & \textbf{1.2553E-1(9.9012E-3)} & 1.3763E-1(9.0469E-3) & \textbf{1.3005E-1(2.2813E-2)} & 1.4622E-1(1.0947E-2) \\
		& 3     & 3.0757E-1(2.0117E-2) & 2.9608E-1(1.7057E-2) & \textbf{2.8315E-1(1.1314E-2)} & 3.1867E-1(1.9034E-2) & 3.4421E-1(9.3072E-3) & 3.2164E-1(2.6158E-2) \\
		& 5     & \textbf{8.9204E-1(5.8709E-2)} & 9.1682E-1(4.2633E-2) & 9.2485E-1(3.4038E-2) & \textbf{8.7249E-1(4.6442E-2)} & 1.3241E+0(4.5983E-2) & 1.0209E+0(5.7778E-2)  \\[0.6mm]
		\hline \\[0.6mm]
		\multirow{3}[0]{*}{SDP7} & 2     & 5.3493E-1(7.5356E-2) & 6.5722E-1(9.4988E-2) & 5.1620E-1(5.0548E-2) & \textbf{4.1199E-1(4.6310E-2)} & 1.7483E+0(2.0452E-3) & 6.1091E-1(1.1539E-1) \\
		& 3     & 4.5983E-1(4.2078E-2) & 7.3671E-1(1.8894E-1) & 5.3748E-1(6.7511E-2) & \textbf{4.0980E-1(4.5337E-2)} & 3.1279E+0(2.3341E-2) & 7.6677E-1(1.6843E-1) \\
		& 5     & \textbf{5.0916E-1(3.9290E-2)} & 1.1914E+0(3.2619E-1) & 7.3788E+0(1.3191E-1) & 5.1857E-1(7.8802E-2) & 7.4986E+0(2.2756E-2) & 7.1892E+0(2.5740E-1)  \\[0.6mm]
		\hline \\[0.6mm]
		\multirow{3}[0]{*}{SDP8} & 2     & 4.9242E-1(1.5994E-2) & 5.2028E-1(1.6206E-2) & 5.0538E-1(1.2588E-2) & 4.6989E-1(1.2951E-2) & \textbf{4.5508E-1(3.2778E-3)} & 4.8187E-1(1.7134E-2) \\
		& 3     & \textbf{1.0964E+0(8.8786E-3)} & 1.1930E+0(1.3547E-2) & 1.1764E+0(4.3077E-3) & \textbf{1.1010E+0(1.1515E-2)} & 1.2464E+0(5.8024E-3) & 1.1639E+0(7.5370E-3) \\
		& 5     & \textbf{4.5715E-1(1.0970E-3)} & \textbf{4.5284E-1(8.2506E-4)} & 4.6233E-1(9.1946E-4) & 4.5968E-1(1.7243E-3) & 4.6776E-1(4.2948E-4) & 4.6536E-1(5.6262E-4) \\[0.6mm]
		\hline \\[0.6mm]
		\multirow{3}[0]{*}{SDP9} & 2     & \textbf{2.2857E-1(4.9302E-2)} & 2.6550E-1(2.7889E-2) & 2.9947E-1(5.0364E-2) & 9.8261E-1(9.2122E-2) & 3.5121E-1(3.9151E-2) & 6.1371E-1(2.2572E-1) \\
		& 3     & 5.8076E-1(1.6534E-1) & 3.7587E-1(1.8286E-1) & \textbf{3.5519E-1(5.3603E-2)} & 1.1706E+0(1.5741E-1) & \textbf{3.5877E-1(5.6254E-2)} & 8.8392E-1(2.3261E-1) \\
		& 5     & \textbf{4.3026E-1(7.0510E-2)} & 8.1945E-1(2.2719E-1) & 5.4562E+0(1.3253E+0) & 4.1443E+0(8.3755E-1) & 6.6607E+0(3.9037E-1) & 8.0503E+0(7.3767E-1)  \\[0.6mm]
		\hline \\[0.6mm]
		\multirow{3}[0]{*}{SDP10} & 2     & 4.8535E-1(1.2233E-2) & 8.6991E-1(1.8656E-1) &5.2545E-1(2.8350E-2) & 9.0088E-1(1.1700E-1) & \textbf{2.0786E-1(7.7987E-2)} & 4.5091E-1(2.2974E-1) \\
		& 3     & 9.7466E-1(7.5181E-2) & 1.1273E+0(1.4791E-1) & 1.1313E+0(4.8748E-2) & 1.2097E+0(4.8728E-2) & \textbf{5.7826E-1(4.9466E-2)} & 1.0329E+0(1.5705E-1) \\
		& 5     & \textbf{2.4277E+0(2.3532E-1)} & 2.9952E+0(1.7315E-1) & 4.6277E+0(2.1824E-1) & 2.5887E+0(7.0161E-2) & 4.7821E+0(1.8992E-1) & 3.2804E+0(3.6742E-1)  \\[0.6mm]
		\hline \\[0.6mm]
		\multirow{3}[0]{*}{SDP11} & 2     & 1.1458E-1(1.3897E-3) & 5.3194E-1(5.3010E-2) & 2.8412E-1(1.8581E-2) & 1.2340E-1(1.0901E-2) & \textbf{1.0049E-1(2.5854E-4)} & \textbf{1.0577E-1(1.4009E-3)} \\
		& 3     & 3.9996E-1(2.8039E-2) & 1.0635E+0(1.0904E-1) & 9.6794E-1(8.7234E-2) & \textbf{2.7464E-1(1.6005E-2)} & 4.4127E-1(2.4336E-2) & 3.7317E-1(2.3627E-2) \\
		& 5     & 1.6579E+0(2.0796E-1) & 1.4669E+0(1.7189E-1) & 3.6437E+0(3.3855E-1) & \textbf{4.2182E-1(4.5408E-2)} & 3.0480E+0(1.7683E-1) & 1.7025E+0(2.2780E-1)  \\[0.6mm]
		\hline \\[0.6mm]
		\multirow{3}[0]{*}{SDP12} & 2     & 3.5179E-1(1.0285E-1) & 7.4422E-1(2.0397E-1) & 6.6953E-1(1.4368E-1) & \textbf{1.8136E-1(7.6552E-2)} & 1.0534E+0(6.3282E-2) & 2.5144E-1(8.3329E-2) \\
		& 3     & 1.0168E+0(1.4743E-1) & 1.0614E+0(8.7546E-2) & 1.5128E+0(1.9773E-1) & \textbf{2.5619E-1(2.0821E-2)} & 1.5673E+0(6.6313E-2) & 6.3320E-1(7.1795E-2) \\
		& 5     & 1.4547E+0(2.6833E-1) & 1.9472E+0(2.3630E-1) & 5.8462E+0(4.5737E-1) & \textbf{3.1837E-1(6.1445E-2)} & 3.7956E+0(3.3165E-1) & 4.4119E+0(3.7432E-1)  \\[0.6mm]
		\hline \\[0.6mm]
		SDP13 & 5     & 1.3846E-1(2.2282E-2) & 1.3974E-1(1.8114E-2) & 1.2871E-1(2.5265E-2) & \textbf{4.4366E-2(6.0979E-3)} & 2.0457E-1(9.2113E-3) & 1.7922E-1(1.4749E-2)  \\[0.6mm]
		\hline \\[0.6mm]
		SDP14 & 5     & 1.7422E-1(1.7950E-2) & 3.9490E-1(5.3944E-2) & \textbf{1.4128E-1(1.2443E-2)} & \textbf{1.4094E-1(2.1197E-2)} & 4.4362E-1(1.2034E-2) & 3.3273E-1(4.3247E-2)  \\[0.6mm]
		\hline \\[0.6mm]
		SDP15 & 5     & 8.4046E+0(2.2752E+0) & 1.1001E+1(1.0079E+0) & 1.1219E+1(1.6377E+0) & \textbf{7.8571E+0(1.6837E+0)} & 1.3903E+1(1.8491E+0) & 9.1073E+0(1.05288E+0) \\
		\hline
	\end{tabular}%
	\label{tab:hvd1}%
\end{table*}
\vspace{-3mm}
\subsection{Algorithm Comparison on SDP Test Suite}
Six algorithms are compared to study their strengths and weaknesses on the proposed SDP test suite. Both MIGD and MHVD values of these algorithms are detailed in \cref{tab:migd1} and \cref{tab:hvd1}, which are roughly consistent with each other. For simplicity, we only focus on in Table \ref{tab:rank_migd} the ranking of six algorithms for each SDP problem based on the MIGD measure. The ranking is calculated as follows. For every SDP problem with $M$ ($M \in \{2, 3, 5\}$) objectives, each algorithm is compared with the rest, according to MIGD, using the Wilcoxon rank-sum test 
at the 0.05 significance level with the standard Bonferroni correction \cite{Abdi07}. The algorithms are then ranked according to the total number of wins over different $M$ values. Algorithms are allowed to have the same rank if there is a tie. This procedure applies to all the SDP problems. Finally, the six algorithms are also assigned a unique rank (the last column in Table \ref{tab:rank_migd}) for their performance on average on the SDP test suite, which is done by sorting the sum of individual ranks for each SDP problem.   
\begin{table}[t]
	\centering
	\addtolength{\tabcolsep}{-2.8pt}
	\caption{Performance ranking based on the MIGD measure for SDP1-15}\vspace{-2mm}
	\begin{tabular}{l|ccccccccccccccc|c}
		\hline
		\multirow{2}[1]{*}{Alg.} &\multicolumn{15}{c|}{SDP} & Avg. \\
		\cline{2-16}
		& 1  & 2  & 3  & 4  & 5  & 6  & 7  & 8  & 9  & 10 & 11 & 12 & 13 & 14 & 15 & {rank} \\
		\hline
		SGEA  & 3     & 5     & 2     & 2     & 2     & 2     & 2     & 2     & 1     & 2     & 4     & 3     & {2} & 1     & 3     & 1 \\
		MOEA/D & 6     & 6     & 3     & 5     & 4     & 5     & 4     & 6     & 3     & 5     & 6     & 4     & {6} & 6     & 2     & 6 \\
		DNSGA-II & 5     & 3     & 6     & 3     & 3     & 1     & 3     & 5     & 3     & 4     & 5     & 6     & {3} & 1     & 5     & 4 \\
		dCOEA & 1     & 4     & 1     & 6     & 5     & 3     & 1     & 1     & 6     & 6     & 1     & 1     & {1} & 3     & 1     & 2 \\
		PPS+RM2 & 2     & 1     & 5     & 1     & 1     & 6     & 6     & 2     & 2     & 1     & 3     & 5     & {4} & 5     & 3     & 3 \\
		PPS+NS & 4     & 2     & 4     & 4     & 6     & 4     & 5     & 2     & 5     & 3     & 2     & 2     & 5     & 4     & 6     & 5 \\
		\hline
	\end{tabular}%
	\label{tab:rank_migd}%
	\vspace{-4mm}
\end{table}%

\begin{figure*}[thb]
	\begin{tabular}{ccc}
		\includegraphics[width=5.5cm, height=3cm]{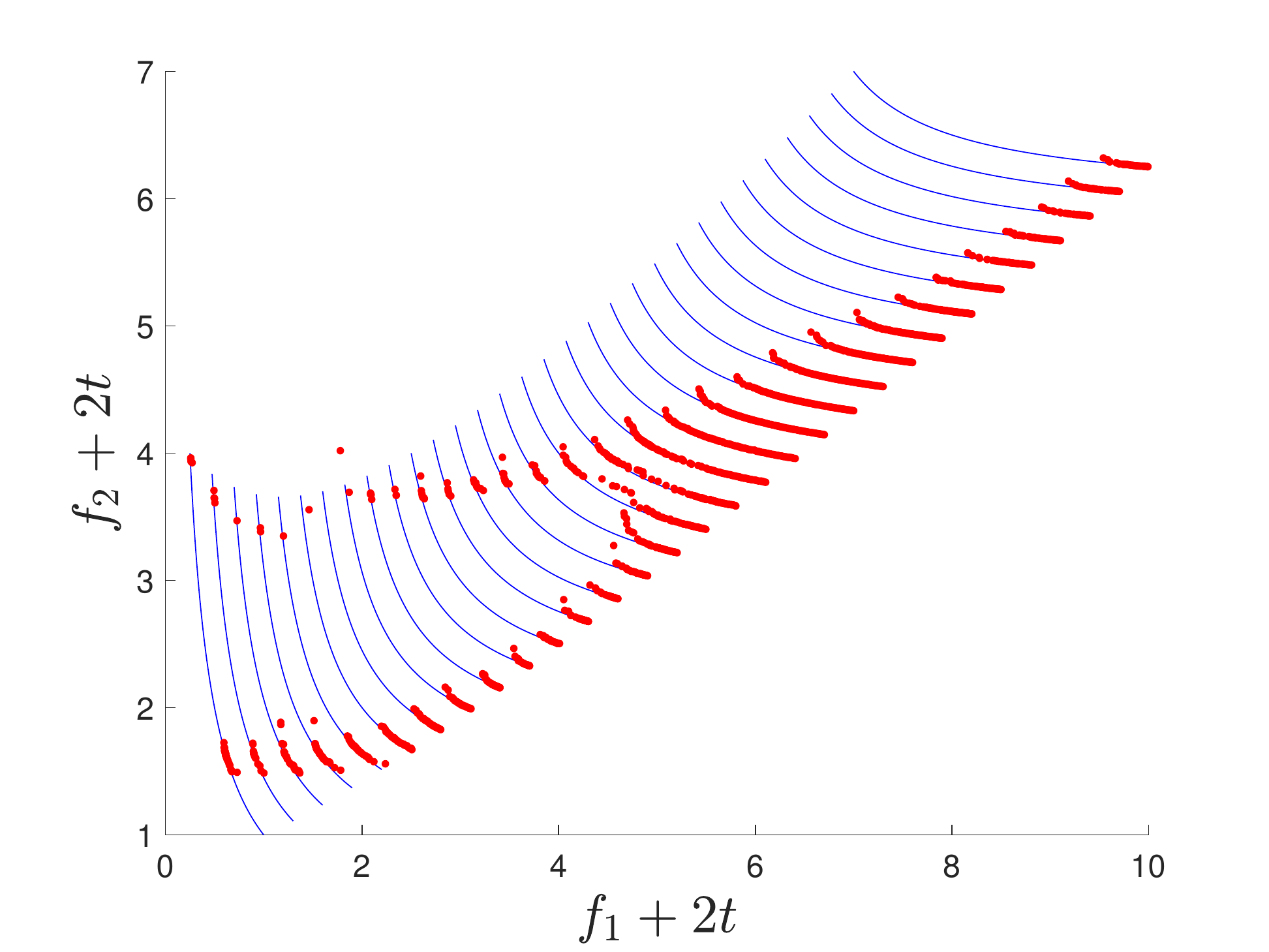}  & 
		\includegraphics[width=5.5cm, height=3cm]{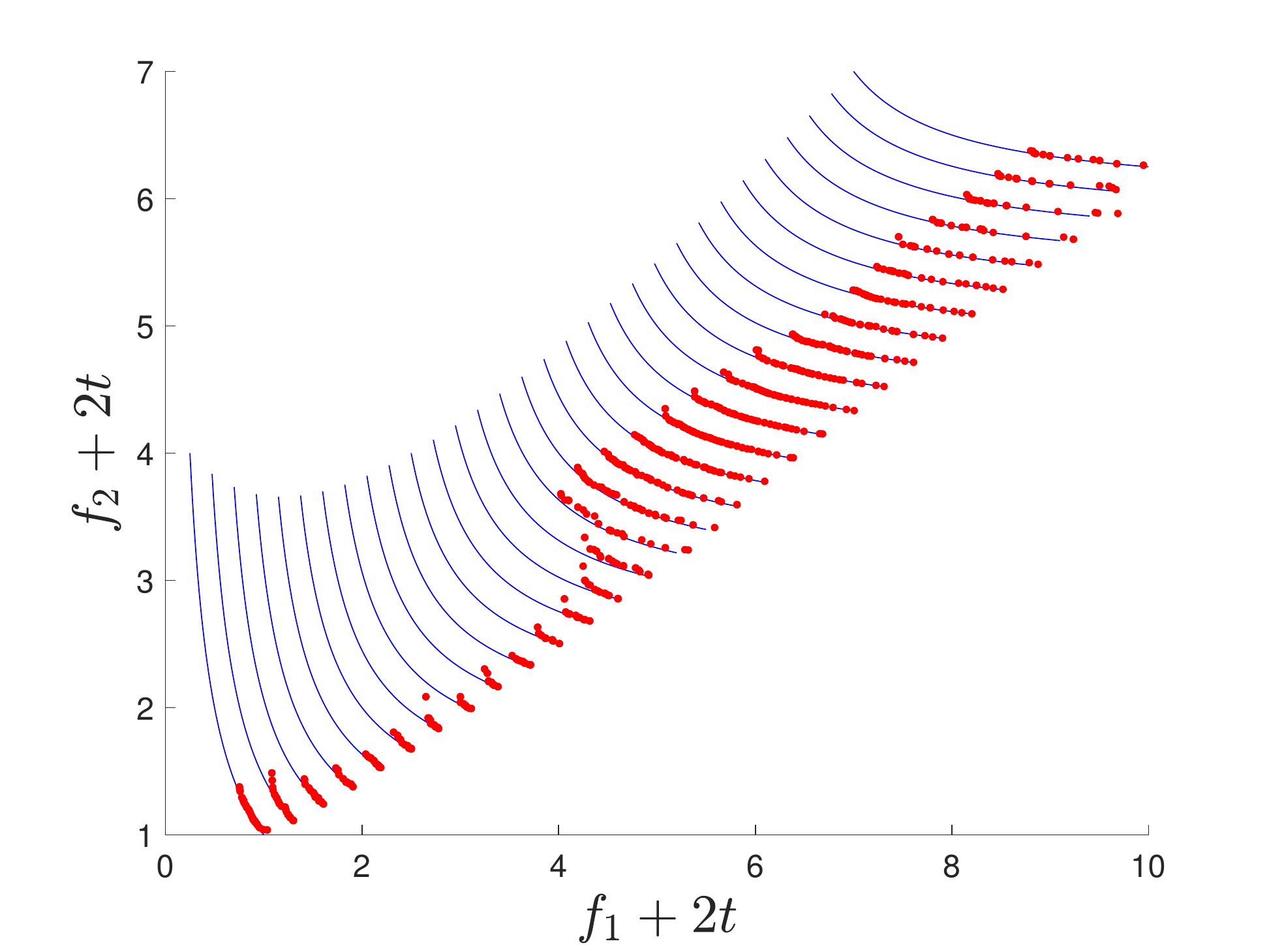} &
		\includegraphics[width=5.5cm, height=3cm]{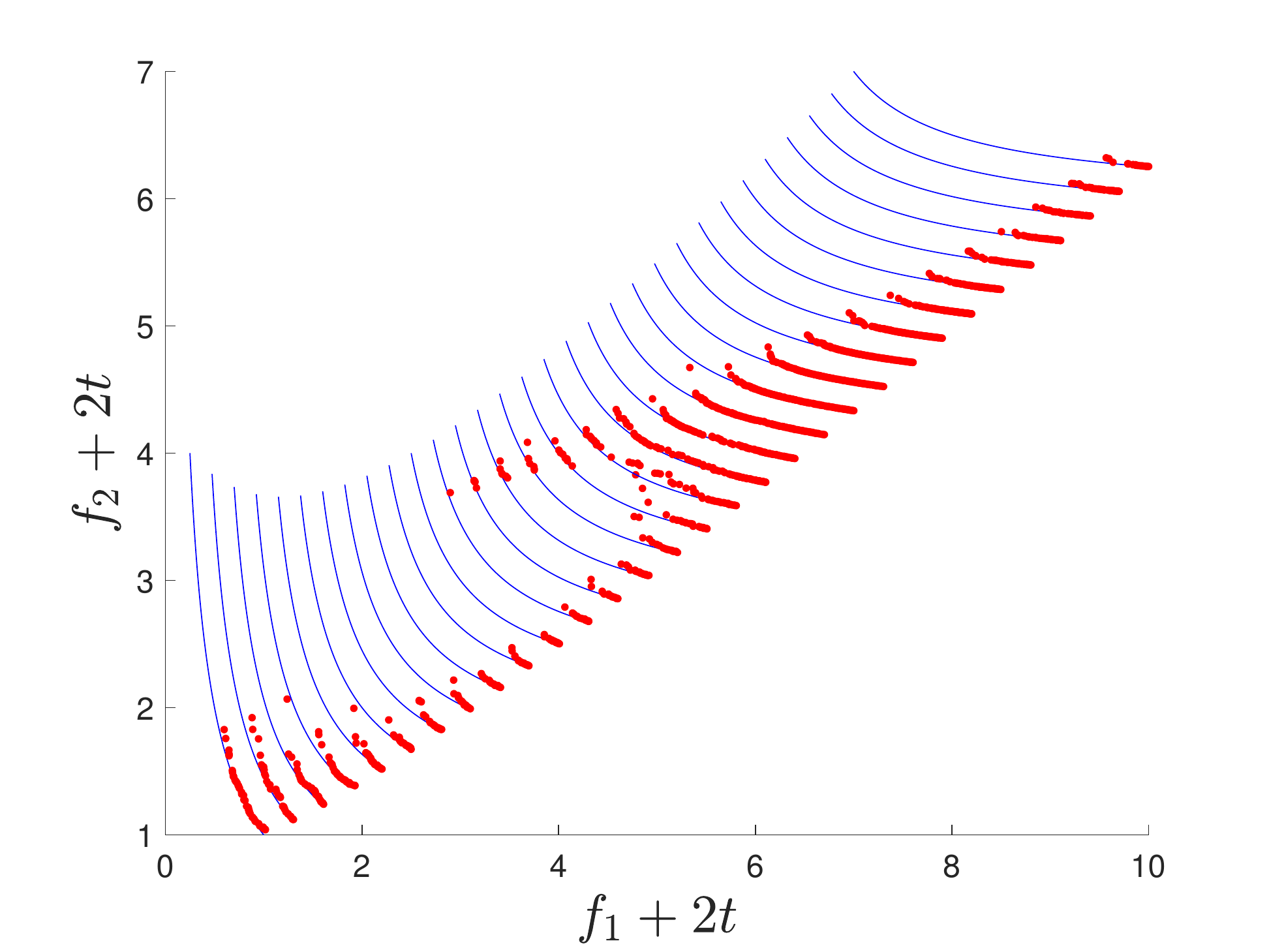} \\
		(a) SGEA & (b) MOEA/D & (c) DNSGA-II \\
		\includegraphics[width=5.5cm, height=3cm]{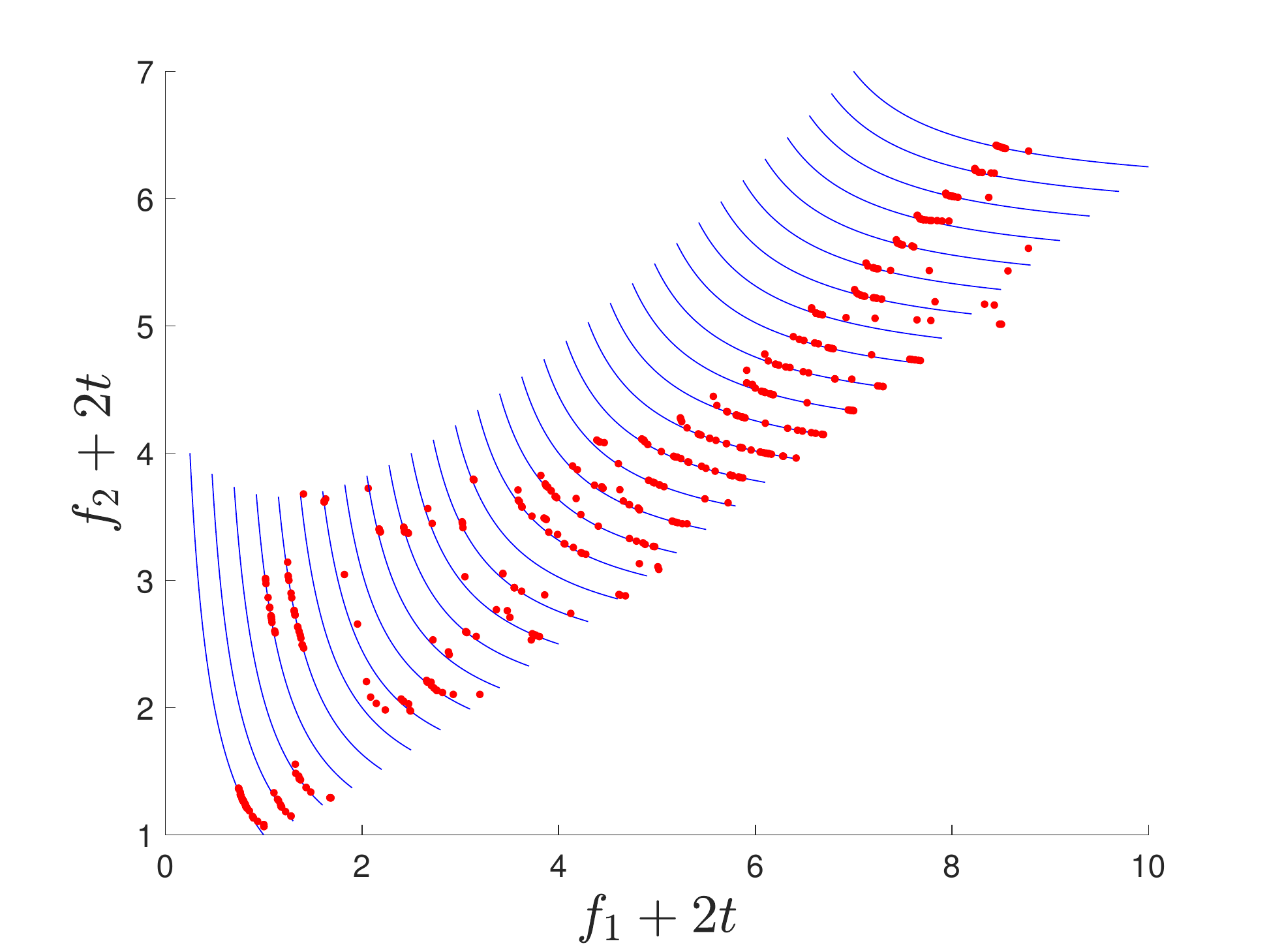}  & 
		\includegraphics[width=5.5cm, height=3cm]{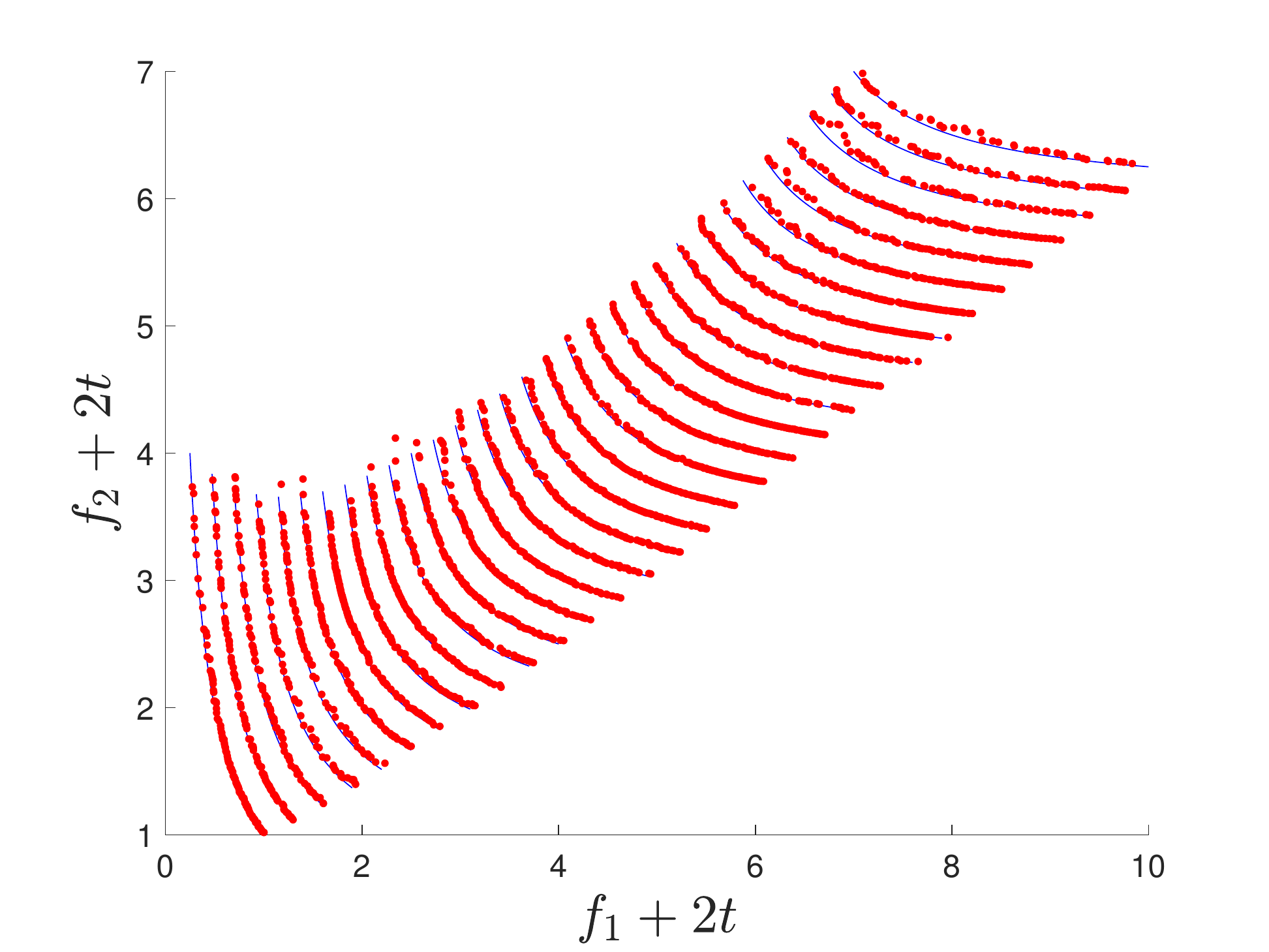} &
		\includegraphics[width=5.5cm, height=3cm]{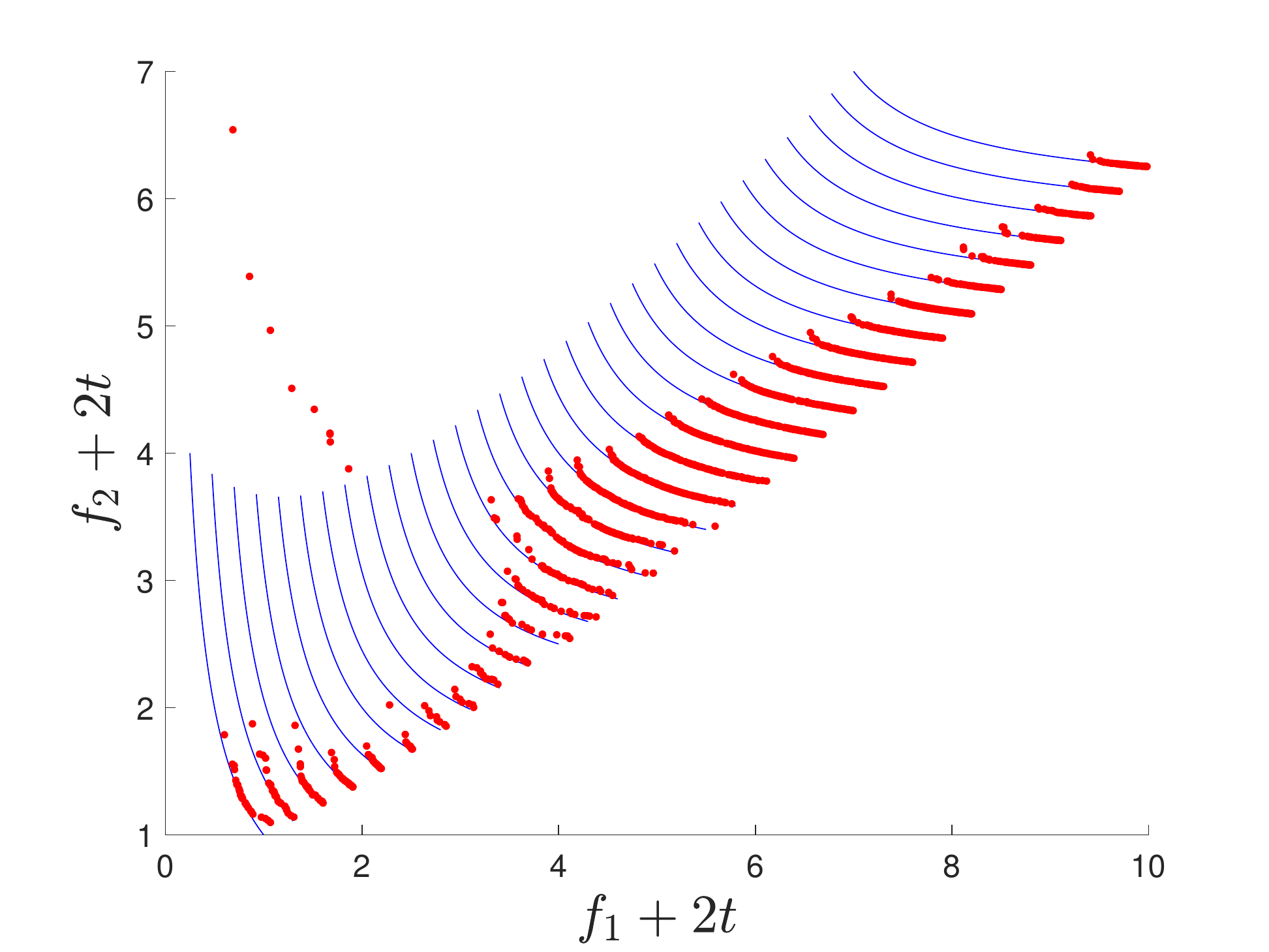} \\
		(d) dCOEA & (e) PPS+RM2 & (f) PPS+NS \\
	\end{tabular}
	\caption{PF approximations of six algorithms for SDP2.}
	\label{fig:sdp2_pa}
	\vspace{-4mm}
\end{figure*}

The ranking table shows that SGEA, dCOEA, and PPS+RM2 are the top three best algorithms for SDP. It suggests that SGEA is most robust although it wins less than dCOEA over 15 SDP problems. dCOEA is in the first place for 8 problems but it obtains poor ranks for the other problms, making it the second in the average ranking. 

dCOEA has the best performance on SDP1 and SDP3, showing it has good ability to track unpredictable PS moves (SDP1) and handle time-varying multimodality (SDP3). For SDP2, the time-dependent stretch of the PF range and variable dependencies present a big challenge to all the algorithm except PPS+RM2, as evidenced in Fig.~\ref{fig:sdp2_pa}. The outperformance of PPS+RM2 can be probably attributed to the regularity model (RM2), which has an advantage of cracking variable linkages. The suspicion is also supported by the performance difference between PPS+RM2 and PPS+NS on other SDP cases, e.g., SDP4-5. This means RM2 sometimes may play a more important role in the PPS+RM2 algorithm than PPS for handling dynamics. 

In addition to variable-linkage, SDP4 and SDP5 also have dynamics on the number of local knee regions and the size of the PF manifold, respectively. dCOEA seems more sensitive to these dynamics and thus not an ideal solver for these problems compared to PPS+RM2 or SGEA.
\\

\begin{figure*}[thb]
	\begin{tabular}{ccc}
		\includegraphics[width=5.5cm]{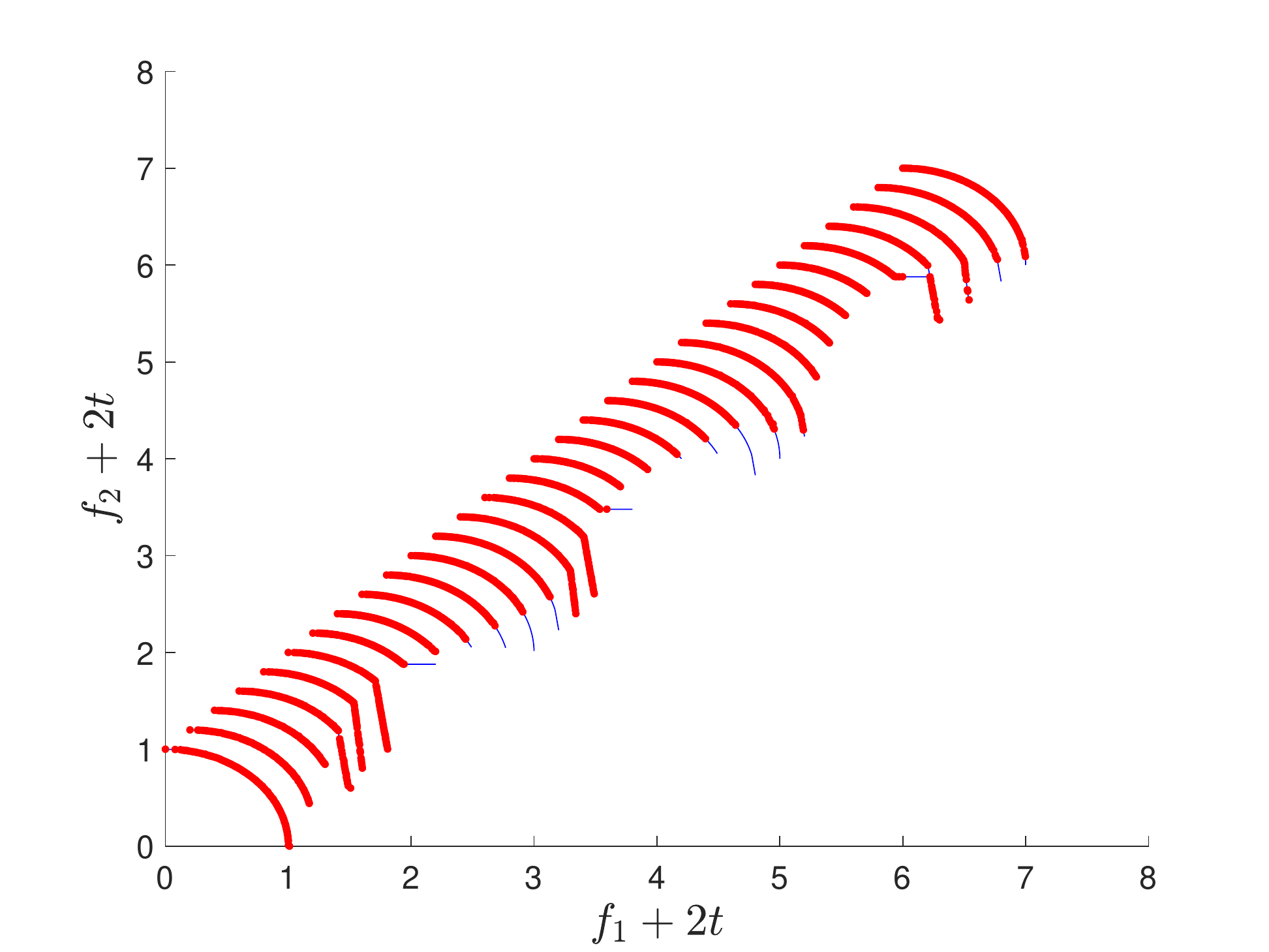}  & 
		\includegraphics[width=5.5cm]{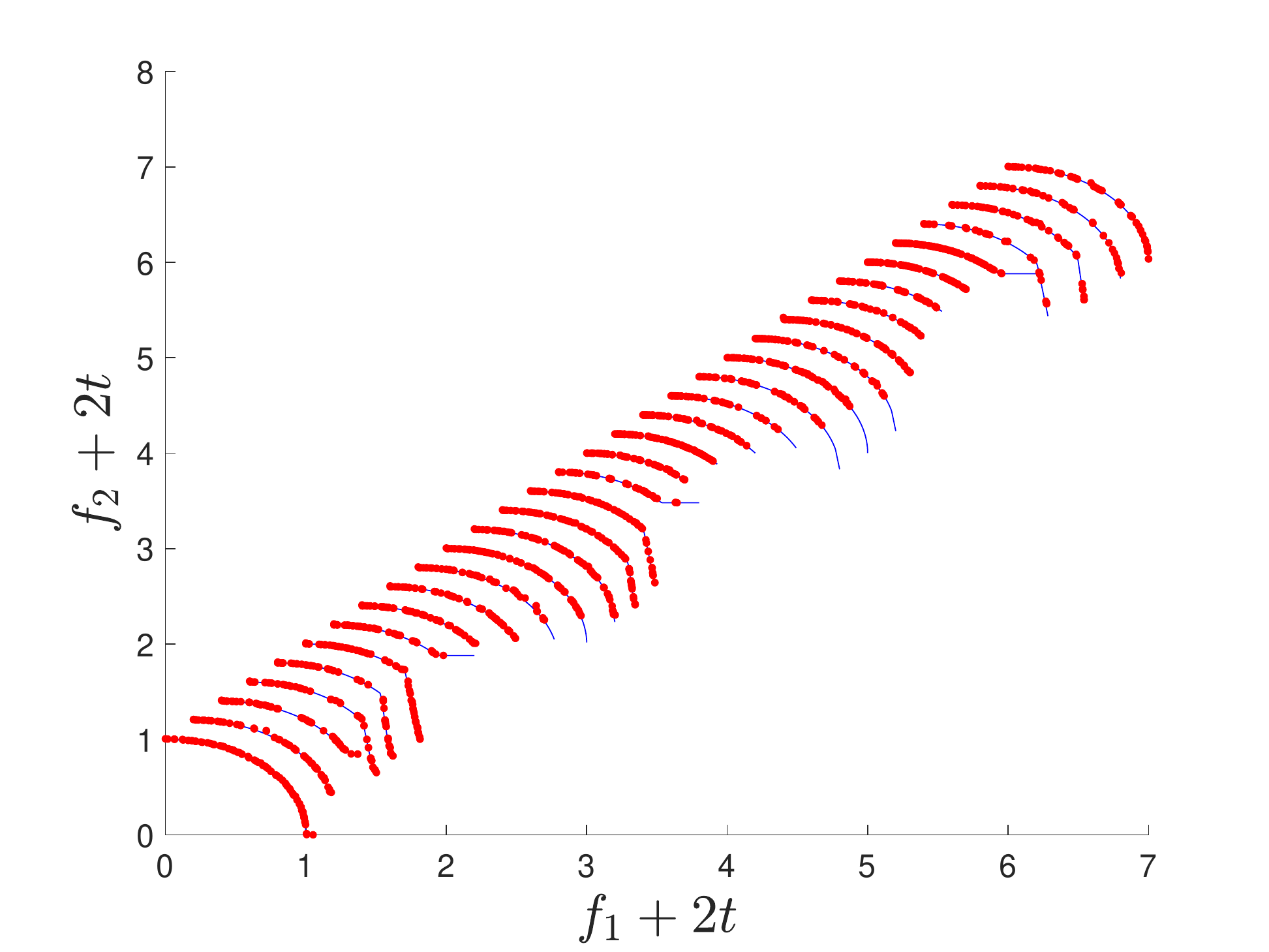} &
		\includegraphics[width=5.5cm]{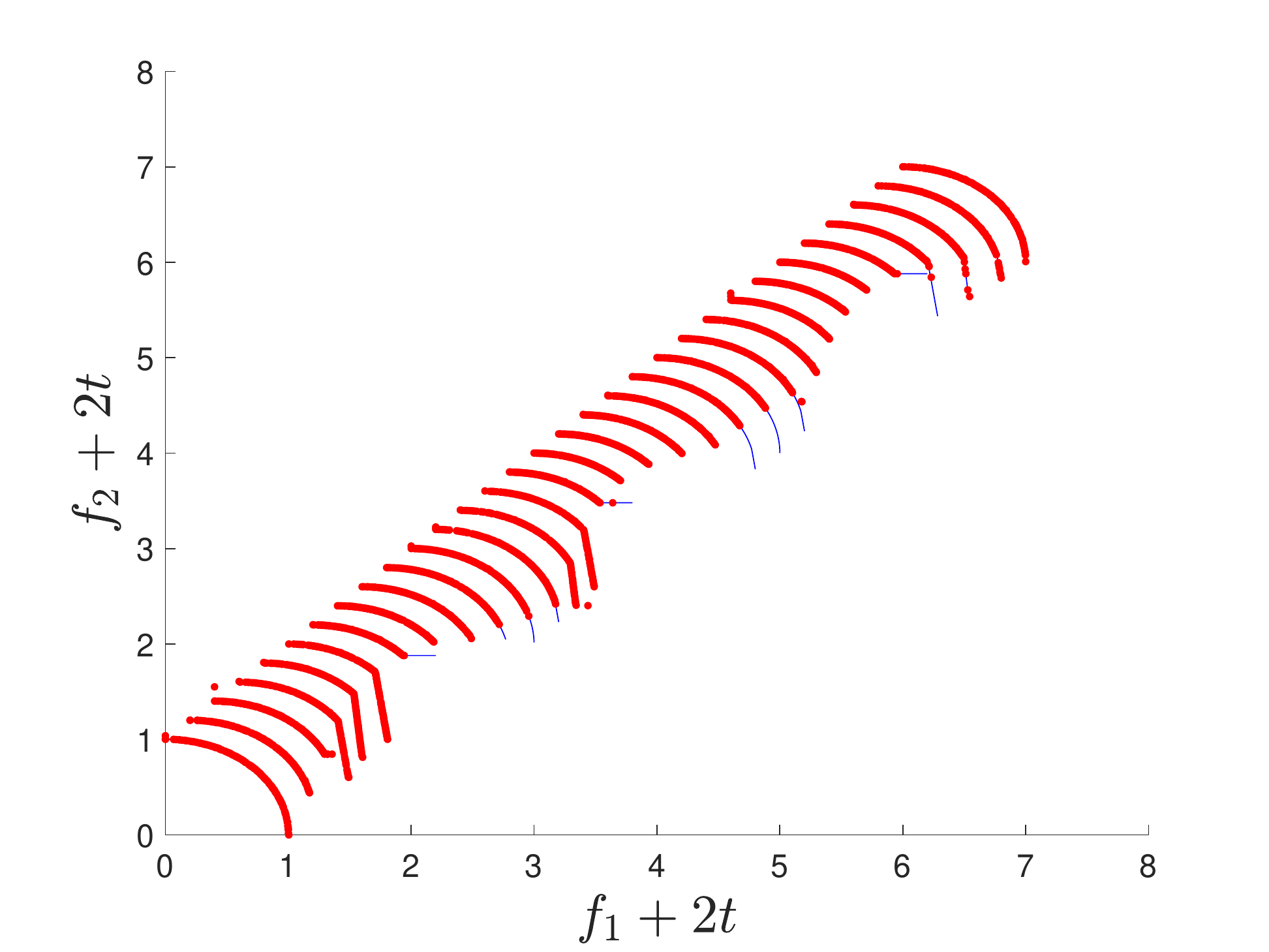} \\
		(a) SGEA & (b) MOEA/D & (c) DNSGA-II \\
		\includegraphics[width=5.5cm]{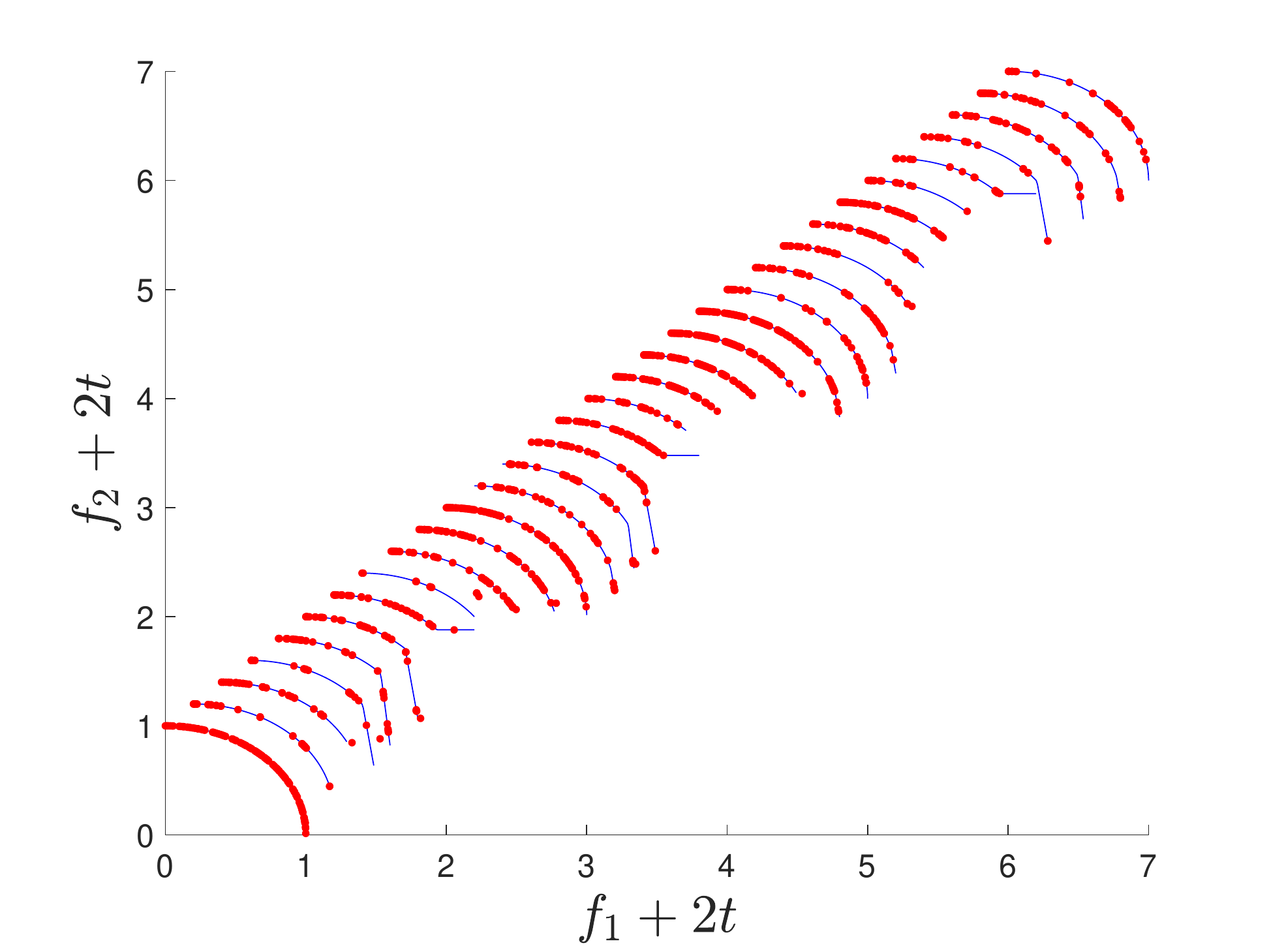}  & 
		\includegraphics[width=5.5cm]{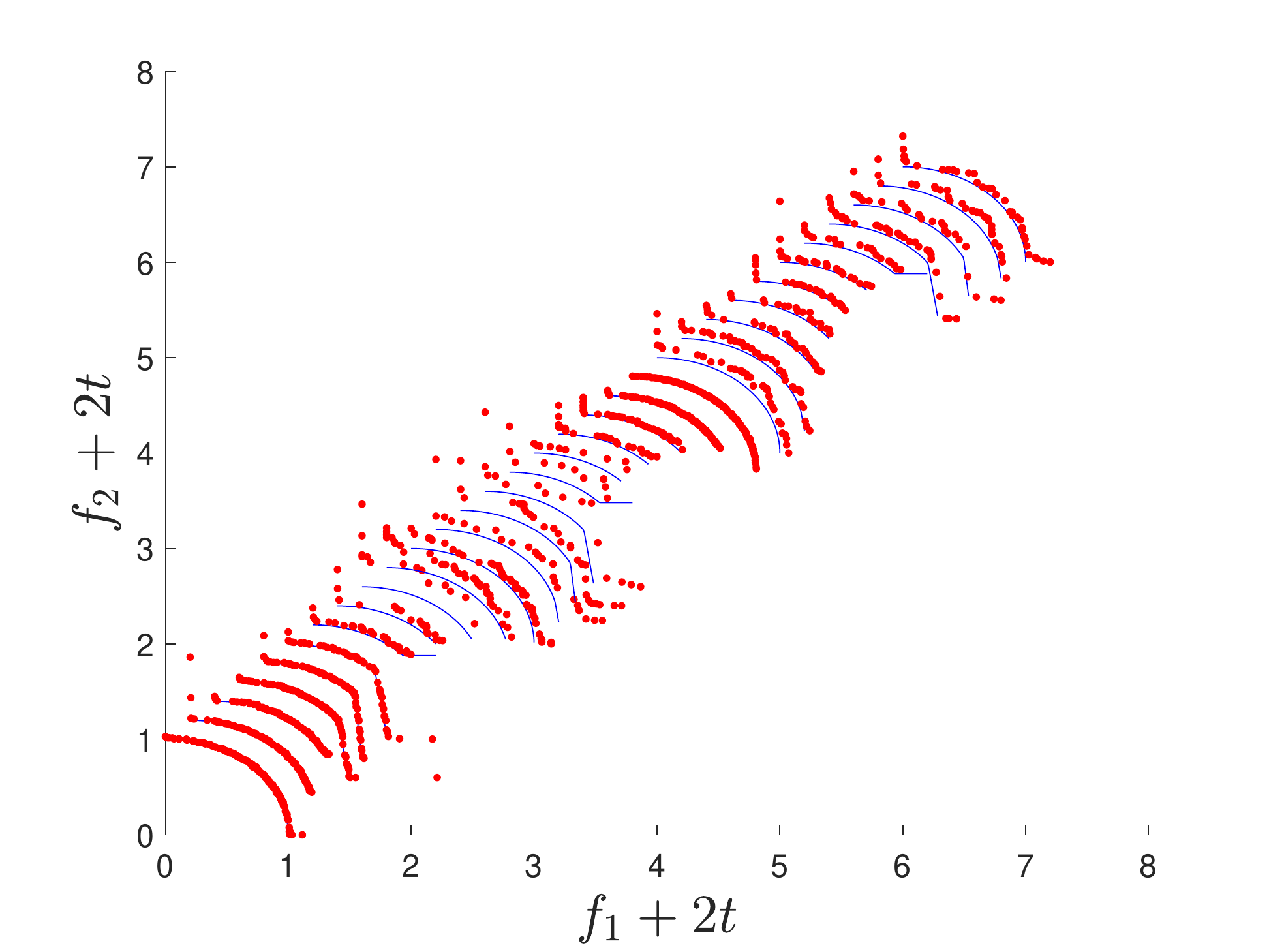} &
		\includegraphics[width=5.5cm]{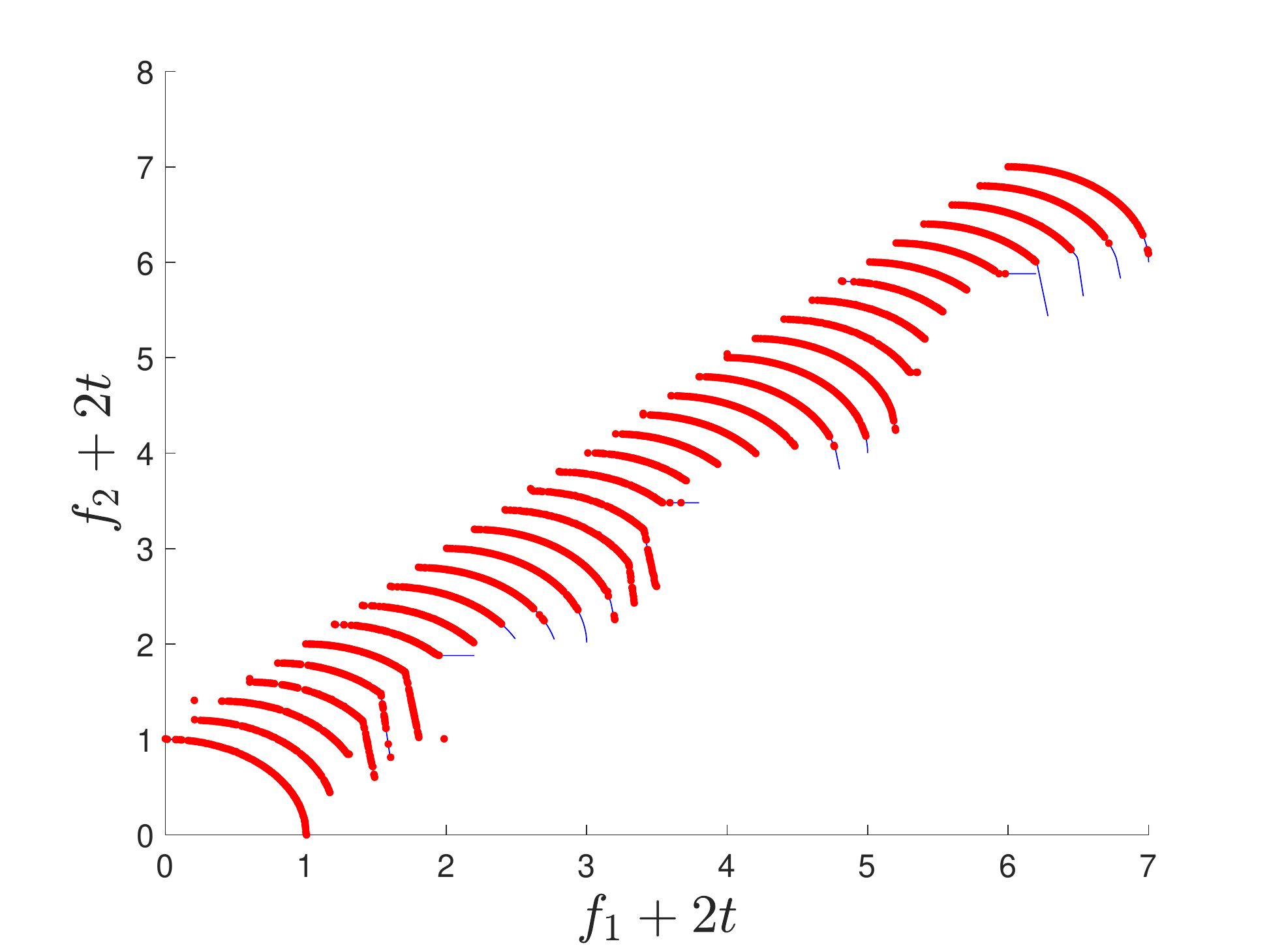} \\
		(d) dCOEA & (e) PPS+RM2 & (f) PPS+NS \\
	\end{tabular}
	\caption{PF approximations of six algorithms for SDP6.}
	\label{fig:sdp6_pfs}
\end{figure*}

SDP6-7 take into account the detectability of environmental changes, in addition to multimodality. DNSGA-II and dCOEA are top solvers for these two problems, respectively. It can be also noticed that PPS+RM2 is least suited to SDP7. We use MDT values, presented in table \ref{tab:detect}, to help us analyse the key challenge in SDP6-7. It is clear that all the six algorithms require multiple reevaluations for change detection. Despite slow change detection performance, SGEA has good MIGD values for both of the problems, implying that SGEA benefits greatly from its change reaction mechanism. PPS+RM2 also has relatively large MDT values, which can be an explanation for its poor performance on SDP6. One the other hand, PPS+RM2 detects changes promptly for SDP7, as evidenced by very small MDT values. However, its large MIGD values imply that PPS+RM2 struggles to track the moving global optima in a multimodal landscape. It is worth mentioning that SDP6-7 are specialised in testing algorithms' detectability of changes, a characteristic that existing test suites do not have. Such speciality is demonstrated by the visualization of PF approximations of SDP6-7 in \cref{fig:sdp6_pfs} and \cref{fig:sdp6_pfs}, from which it is clear to see that a failure to detect changes leads to the loss of PF tracking.

\begin{table*}[htbp]
	\centering
	\footnotesize
	\addtolength{\tabcolsep}{-4pt}
	\renewcommand{\arraystretch}{-0.2}
	\caption{Mean and standard deviation values of MDT obtained by six algorithms for SDP6-7 (Best values are highlighted in boldface)}\vspace{-2mm}
	\begin{tabular}{ccllllll}
		\toprule
		Prob. & M     & \multicolumn{1}{c}{SGEA} & \multicolumn{1}{c}{MOEA/D} & \multicolumn{1}{c}{DNSGA-II} & \multicolumn{1}{c}{dCOEA} & \multicolumn{1}{c}{PPS+RM2} & \multicolumn{1}{c}{PPS+NS} \\
		\midrule
		\multirow{3}[2]{*}{SDP6} & 2     & 9.8350E-2(4.6869E-2) & 6.6128E-2(2.6956E-2) & 7.9680E-2(2.5718E-2) & 9.6768E-2(2.8321E-2) & 8.9680E-2(1.7084E-2) & \textbf{5.7929E-2(2.0722E-2)} \\
		& 3     & 8.9175E-2(1.8792E-2) & 5.7239E-2(1.2390E-2) & 5.3316E-2(1.7277E-2) & \textbf{5.2054E-2(1.1462E-2)} & 5.4579E-2(1.8596E-2) & 5.4697E-2(1.4627E-2) \\
		& 5     & 8.5354E-2(2.9489E-2) & 3.8939E-2(1.3097E-2) & \textbf{3.1279E-2(1.1426E-2)} & 3.7071E-2(1.6682E-2) & 7.5471E-2(1.7673E-2) & 3.3872E-2(8.5727E-3) \\[0.6mm]
		\hline \\[0.6mm]
		\multirow{3}[2]{*}{SDP7} & 2     & 3.0690E-2(2.0408E-2) & 2.2896E-2(2.1634E-2) & 1.2795E-2(8.5305E-3) & 2.2222E-3(4.1459E-3) & \textbf{1.8350E-3(4.9067E-3)} & 1.5825E-2(1.7273E-2) \\
		& 3     & 3.0976E-2(2.1158E-2) & 3.0741E-2(2.0845E-2) & 3.9815E-2(2.5565E-2) & 7.6936E-3(9.9103E-3) & \textbf{2.9125E-3(2.2587E-4)} & 5.6869E-2(2.5704E-2) \\
		& 5     & 3.5354E-2(2.2739E-2) & 2.6364E-2(2.0363E-2) & \textbf{8.9226E-4(2.2100E-3)} & 4.7643E-3(5.7931E-3) & 2.9293E-3(7.3574E-3) & 1.0943E-3(9.1315E-4) \\
		\hline
	\end{tabular}%
	\label{tab:detect}%
	\vspace{-4mm}
\end{table*}%
\begin{figure*}[thb]
	\begin{tabular}{ccc}
		\includegraphics[width=5.5cm]{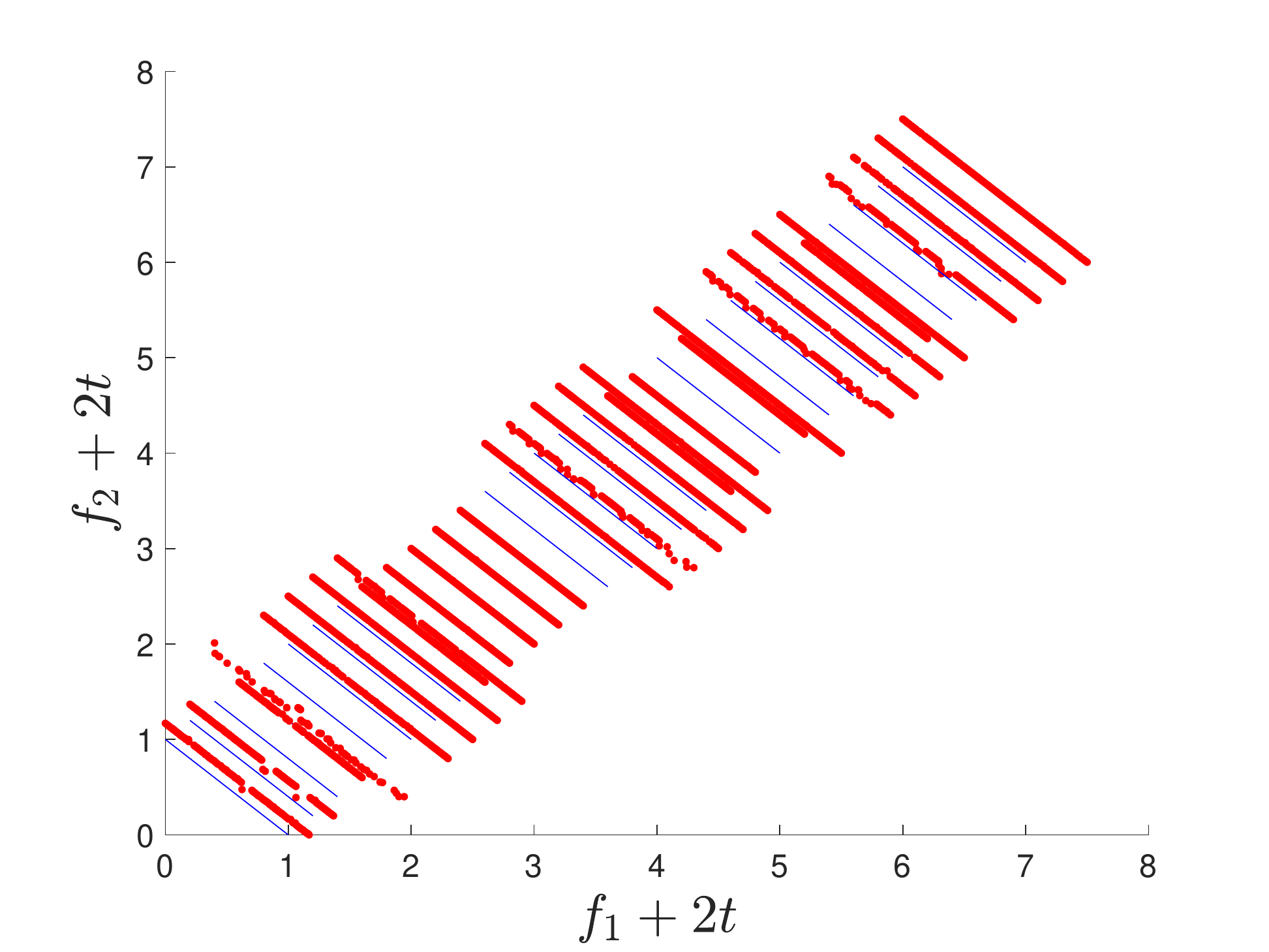}  & 
		\includegraphics[width=5.5cm]{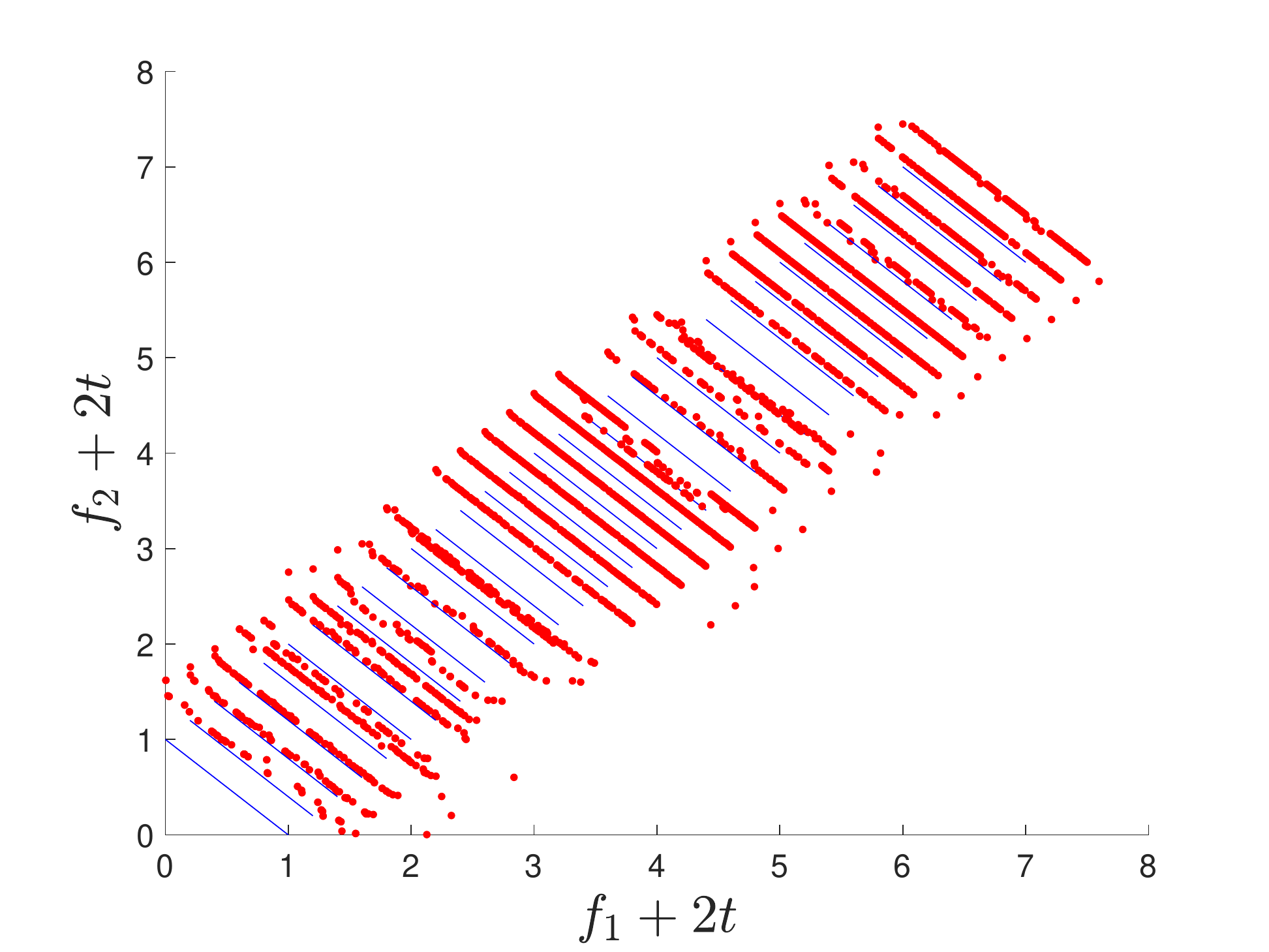} &
		\includegraphics[width=5.5cm]{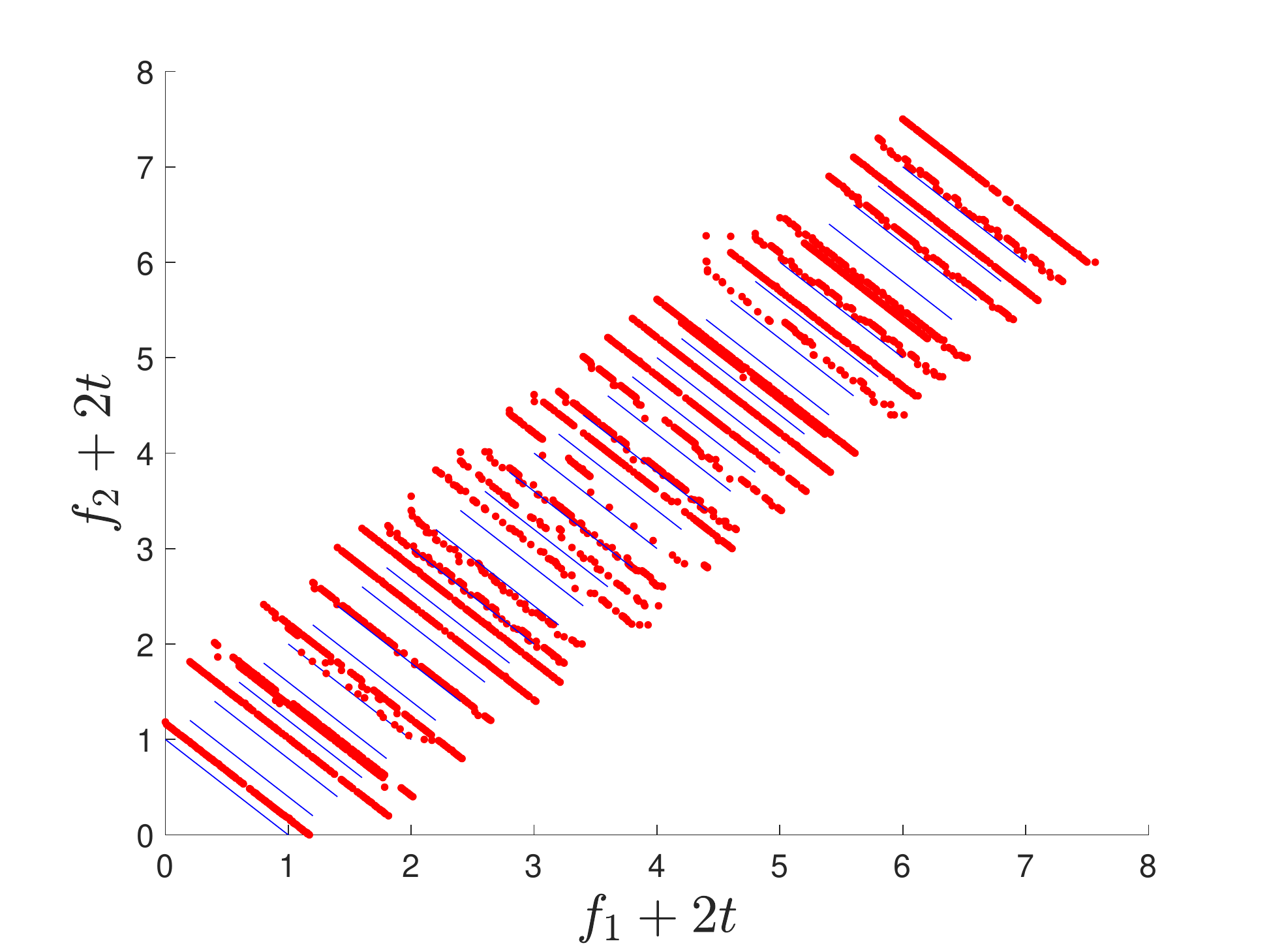} \\
		(a) SGEA & (b) MOEA/D & (c) DNSGA-II \\
		\includegraphics[width=5.5cm]{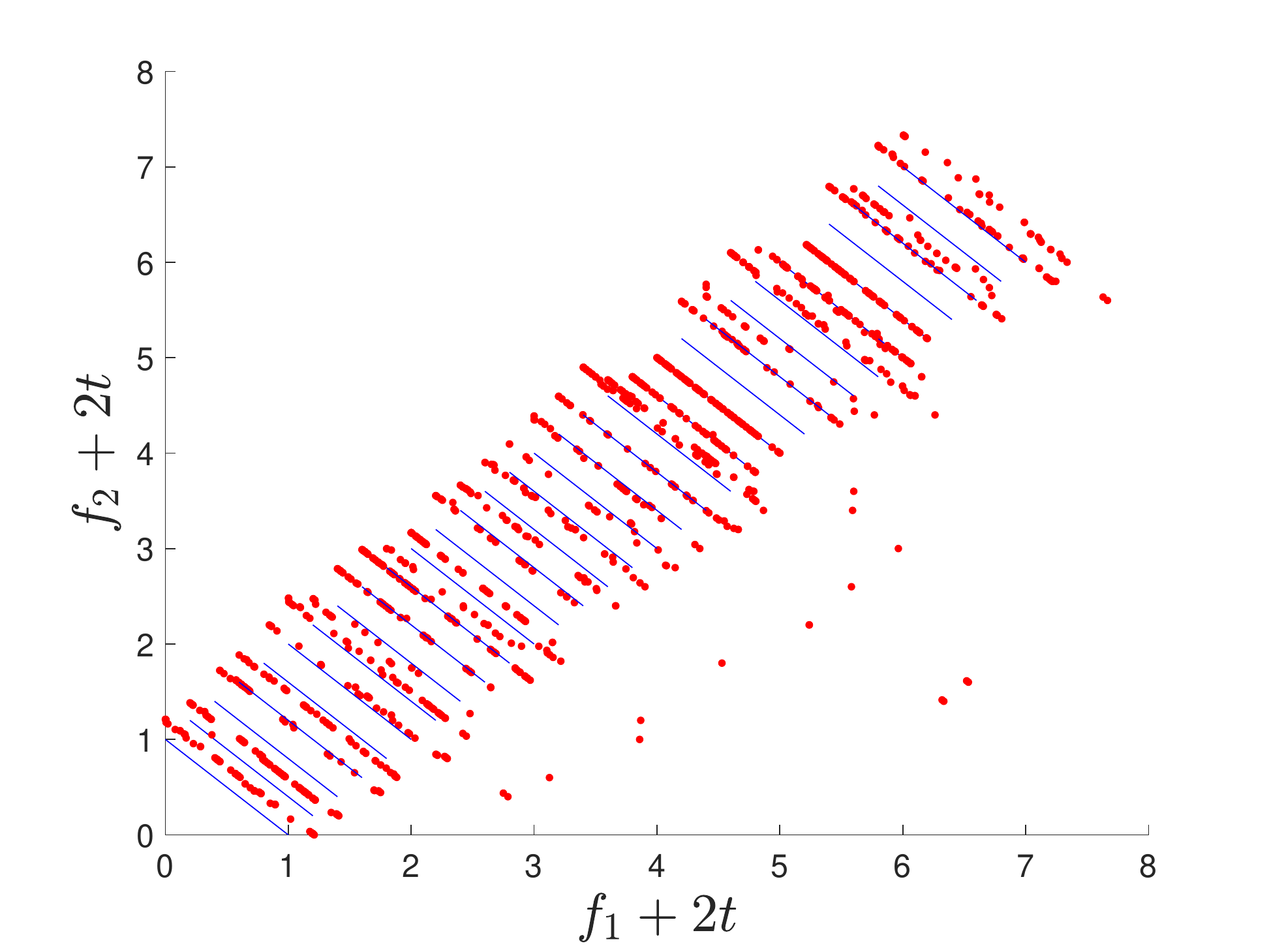}  & 
		\includegraphics[width=5.5cm]{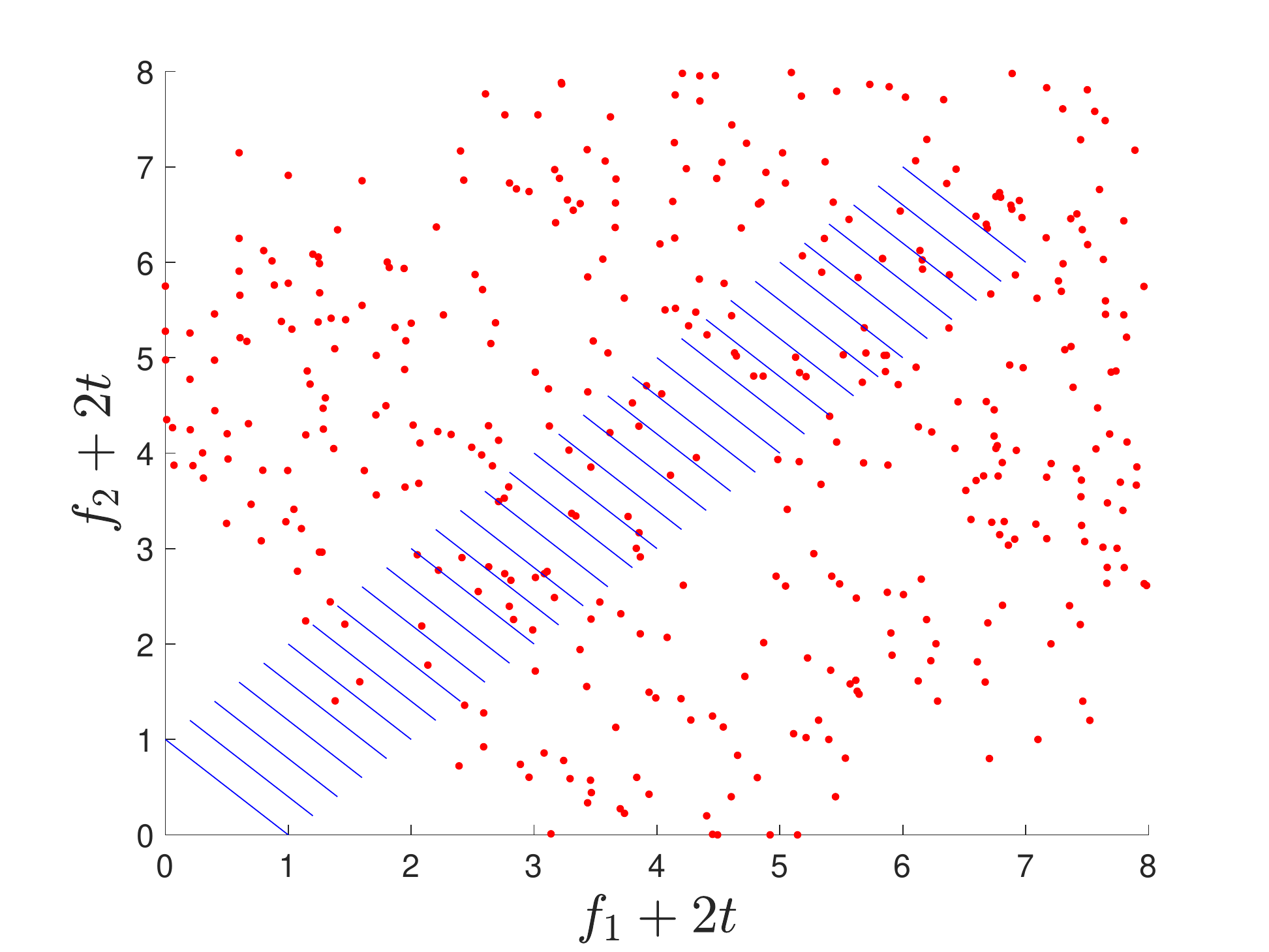} &
		\includegraphics[width=5.5cm]{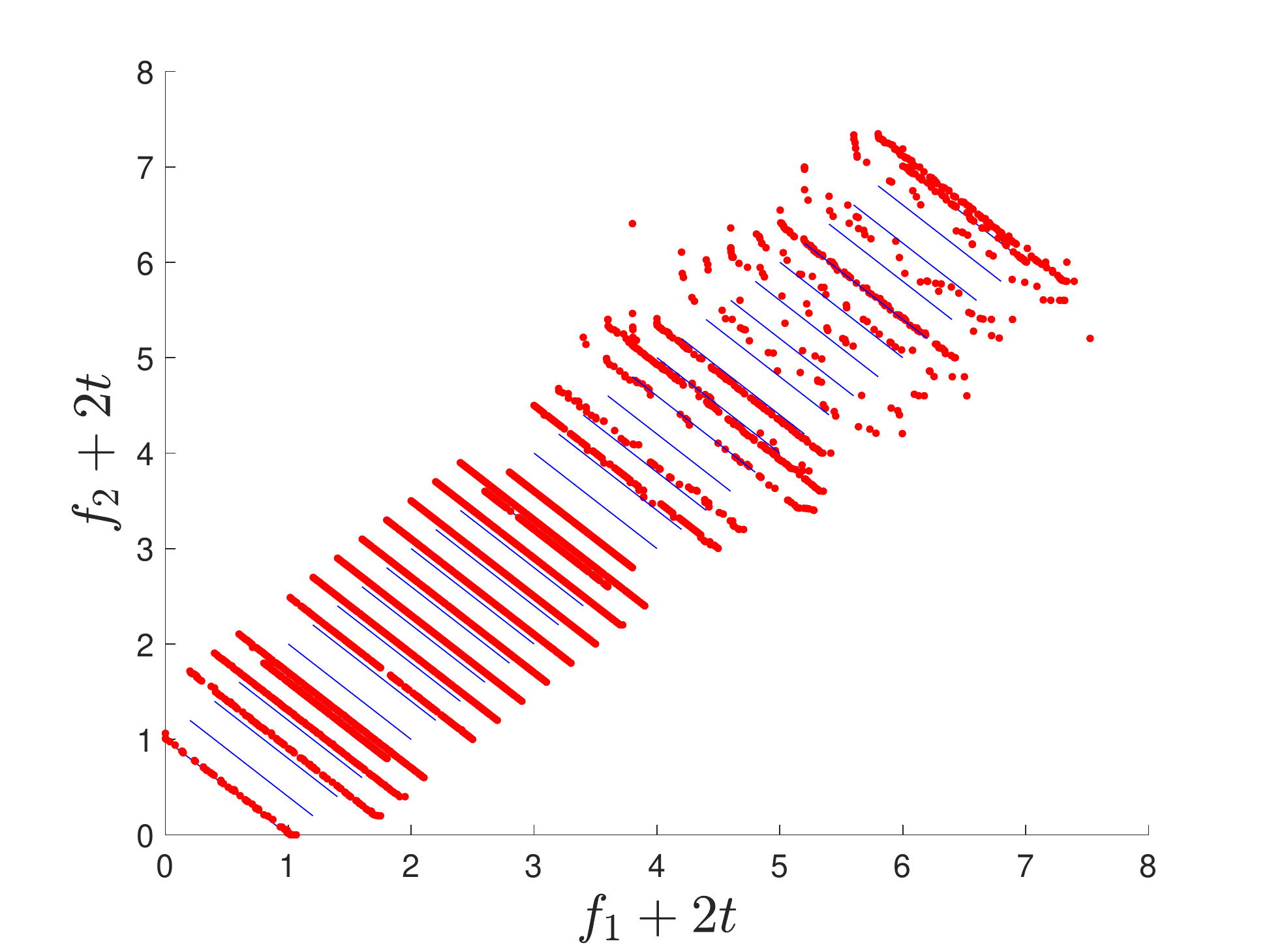} \\
		(d) dCOEA & (e) PPS+RM2 & (f) PPS+NS \\
	\end{tabular}
	\caption{PF approximations of six algorithms for SDP7.}
	\label{fig:sdp7_pfs}
\end{figure*}
The time-varying and non-simply connected PF of SDP8 influences MOEA/D most. This is due to the waste of weight vectors in MOEA/D that pass through PF holes. Thus, SDP8 is a good adversarial example for decomposition-based DMOP solvers. 

\begin{figure*}[thb]
	\begin{tabular}{cccc}
		\includegraphics[width=4.1cm]{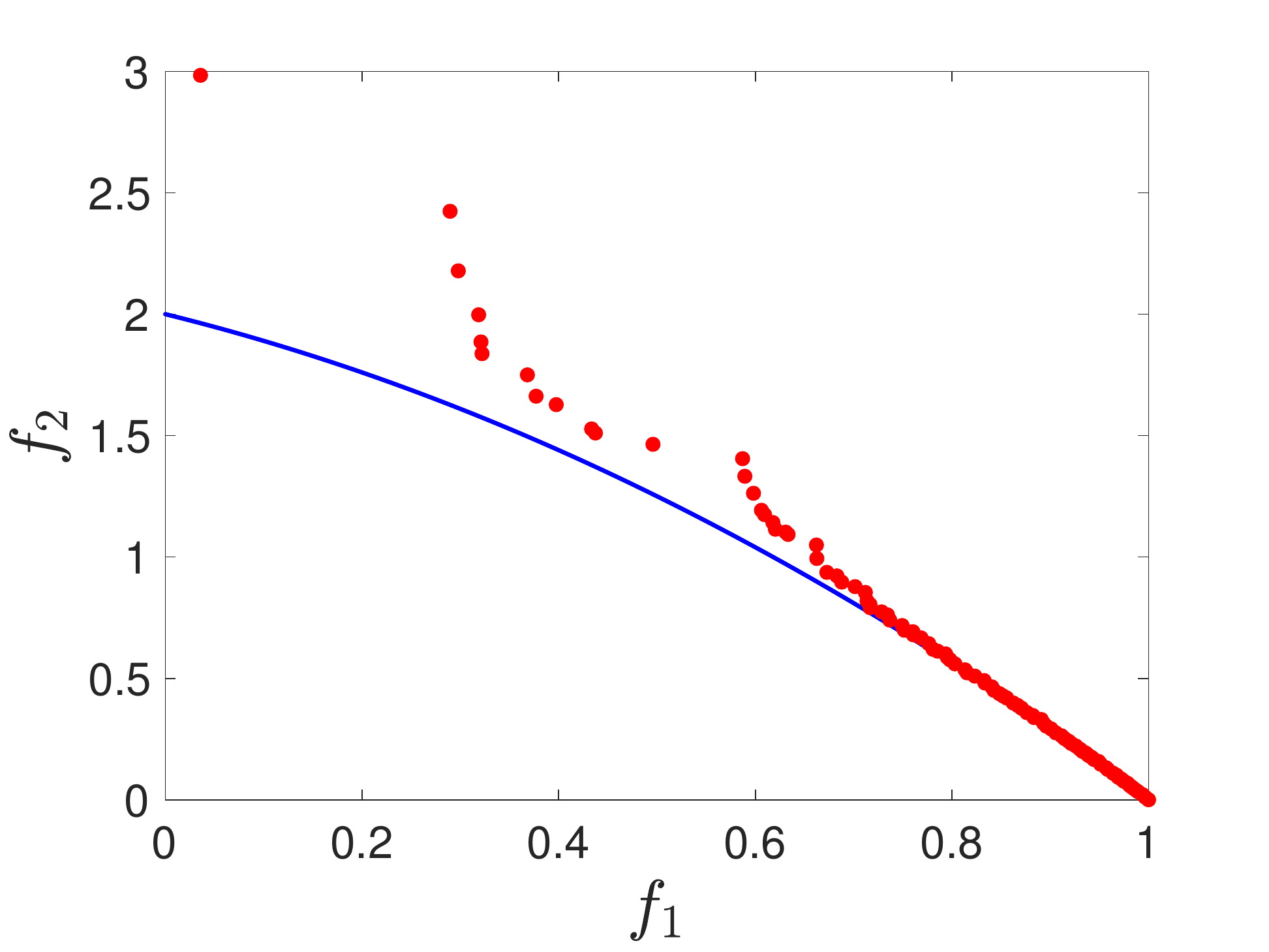}  & 
		\includegraphics[width=4.1cm]{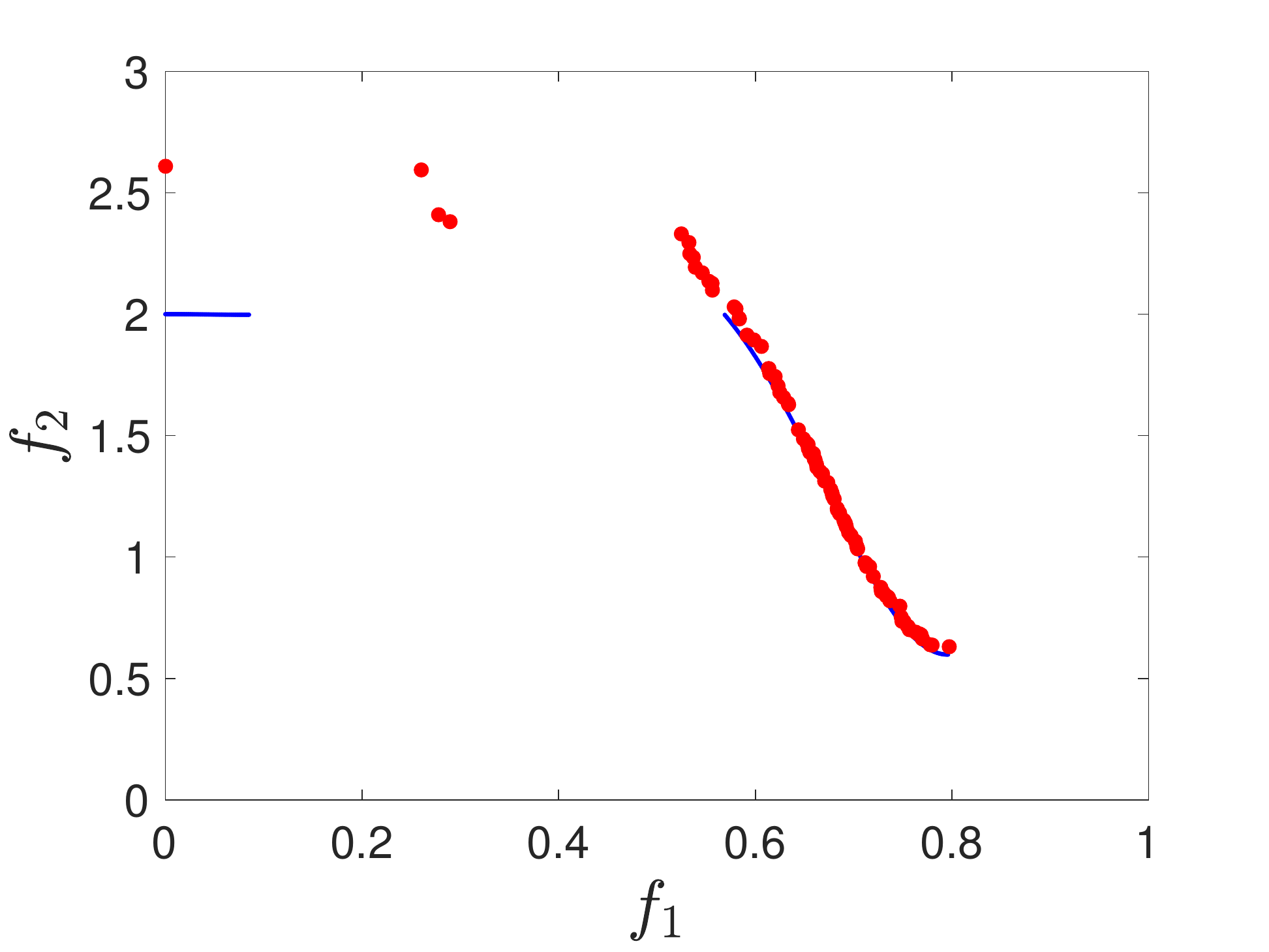} &
		\includegraphics[width=4.1cm]{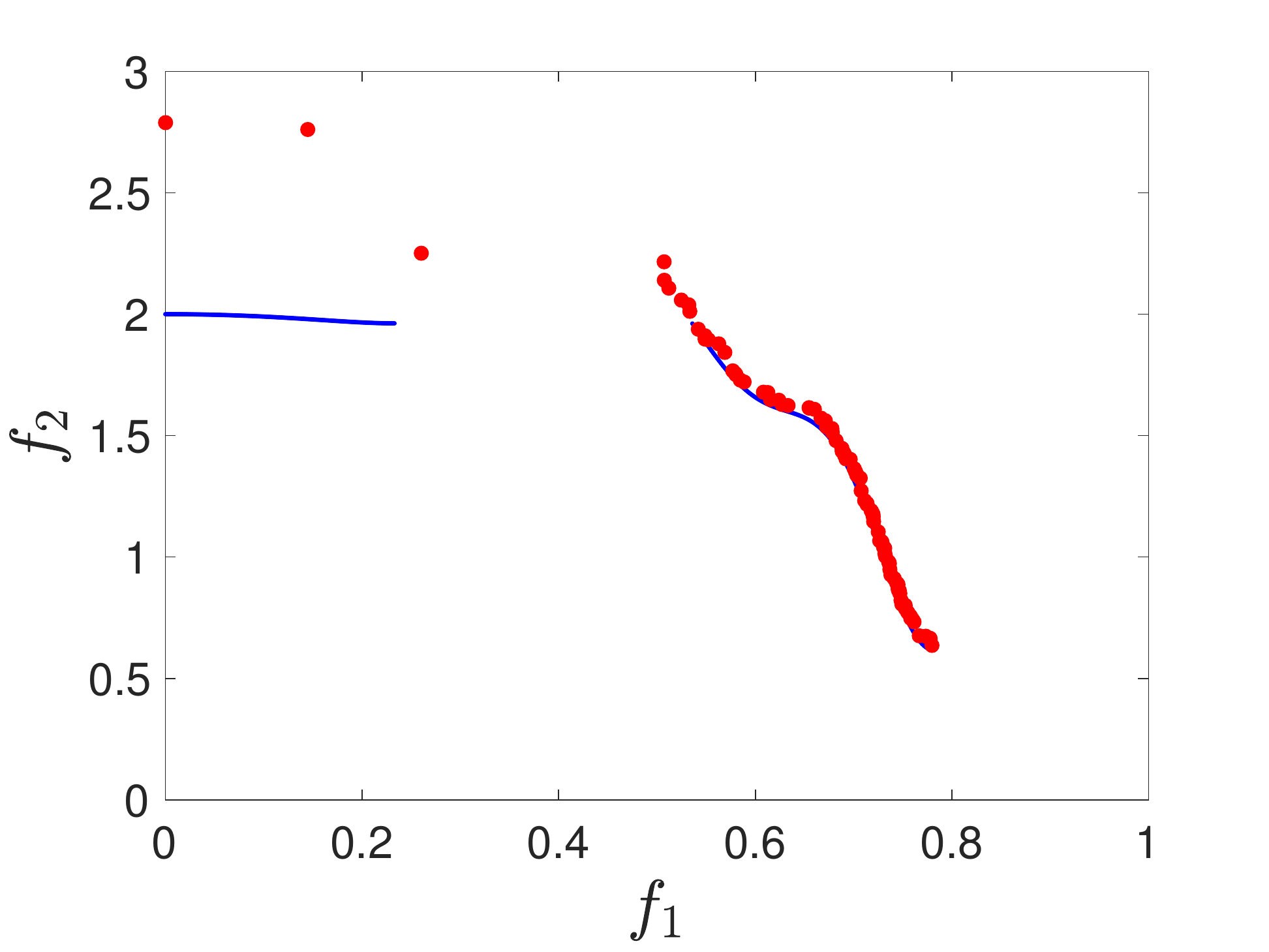} &
		\includegraphics[width=4.1cm]{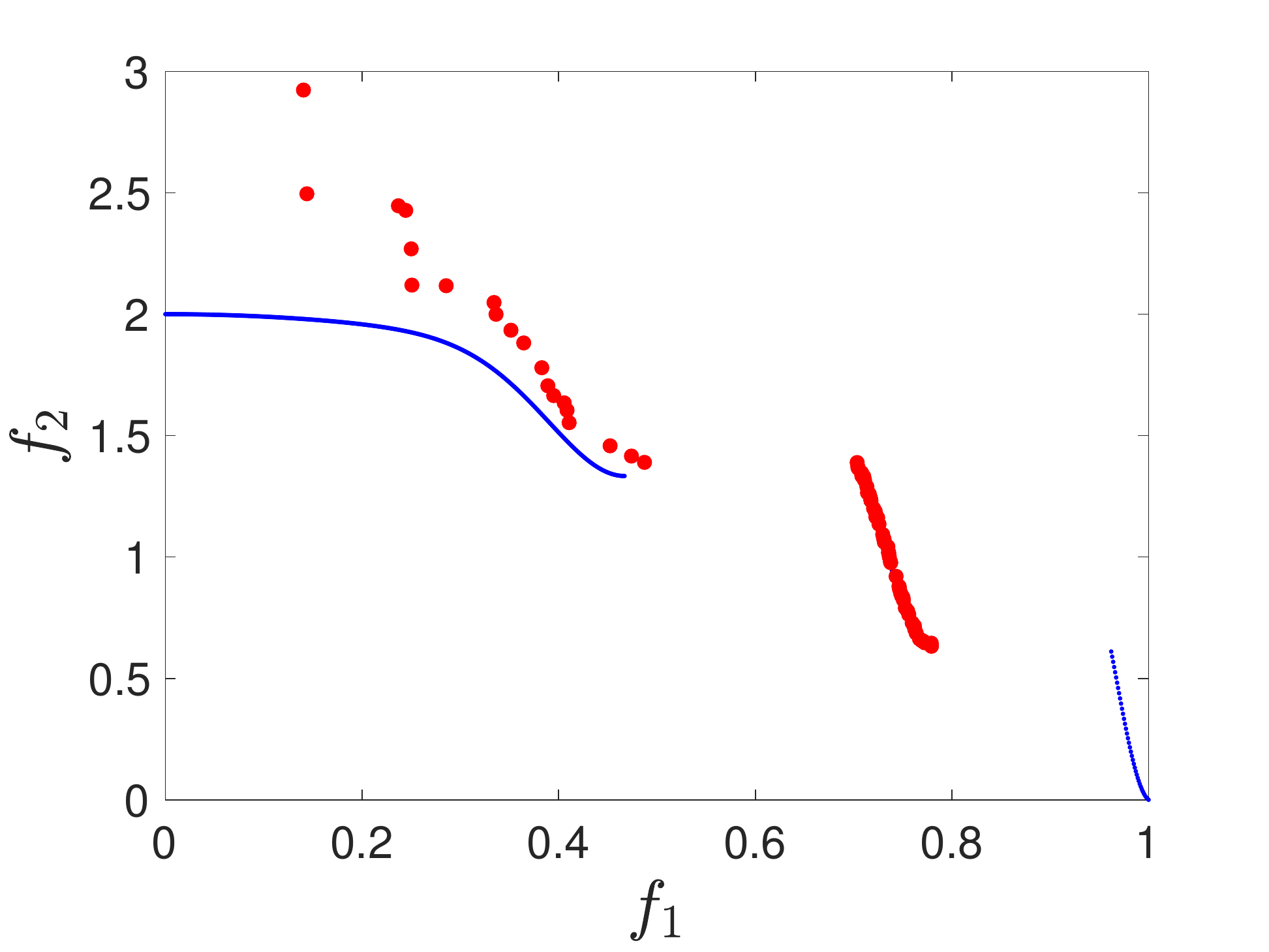} 	\\
		(a) $t=0$ & (b) $t=0.1$ & (c) $t=0.2$ & (d) $t=0.3$
	\end{tabular}
	\caption{PF approximations of SGEA for SDP10 from $t=0$ to $t=0.3$.}
	\label{fig:sdp10_SGEA}
	\vspace{-4mm}
\end{figure*}

Disconnected DMOPs are challenging. This applies especially to dCOEA for SDP9-10. 
For disconnected DMOPs, algorithms may have difficulty in finding the whole disconnected PF segments. Taking SDP10 as an example, we plot the PF approximations obtained by SGEA in Fig.~\ref{fig:sdp10_SGEA}, for the first four time steps. As seen, SGEA struggles to approximate the leftmost PF segment. There is a piece of PF segment for $f_1 \in [0.8,1]$ at $t=0$, but it disappears at $t$=0.1 and 0.2. Then, it is present at $t=0.3$ and SGEA fails to identify its appearance. The dynamics of this problem cause SGEA to lose diversity when certain PF regions disappear suddenly and appear after several environmental changes. Other optimisers experience similar difficulties.

\begin{table*}[htbp]
	\centering
	\caption{Comparison between some SDP instances and other popular DMOPs on the MHVD metric}
	\begin{tabular}{lllllll}
		\toprule
		Prob. & M     & \multicolumn{1}{c}{SGEA} & \multicolumn{1}{c}{MOEA/D} & \multicolumn{1}{c}{DNSGA-II} & \multicolumn{1}{c}{dCOEA} & \multicolumn{1}{c}{PPS+RM2} \\
		\midrule
		SDP1  & 2     & 5.2670E-1(8.4645E-2) & 3.7076E+0(2.8571E-1) & 2.3853E+0(2.6906E-1) & 4.5425E-1(8.1736E-2) & 1.6651E+0(1.0725E-1) \\
		SDP3  & 2     & 4.6648E-1(5.3196E-2) & 1.2608E+0(4.8253E-2) & 1.4479E+0(4.0328E-2) & 6.8189E-1(6.2141E-2) & 1.4802E+0(9.5330E-2) \\
		SDP8  & 2     & 4.9242E-1(1.5994E-2) & 5.2028E-1(1.6206E-2) & 5.0538E-1(1.2588E-2) & 4.6989E-1(1.2951E-2) & 4.5508E-1(3.2778E-3) \\
		FDA1  & 2     & 3.8112E-2(1.4430E-2) & 2.8825E-1(2.9076E-2) & 1.3610E-1(1.7463E-2) & 8.5252E-2(2.0248E-2) & 2.9712E-1(1.6596E-2) \\
		FDA2  & 2     & 1.6745E-2(1.4126E-2) & 6.2906E-2(1.8881E-2) & 2.0598E-2(1.4744E-2) & 1.2486E-1(4.6708E-2) & 2.6663E-1(1.4716E-2) \\
		dMOP1  & 2     & 1.9048E-2(1.4569E-2) & 2.5712E-2(1.5458E-2) & 2.2844E-2(2.0365E-2) & 1.1236E-1(2.0863E-2) & 9.2754E-2(1.3906E-1) \\
		dMOP3  & 2     & 3.1866E-1(2.9555E-2) & 4.5433E-1(2.8194E-2) & 4.7415E-1(2.8497E-2) & 6.5770E-2(1.3652E-2) & 2.7970E-1(2.7200E-2) \\\hline
		SDP1  & 3     & 1.0081E+1(1.1146E+0) & 1.8618E+1(1.1888E+0) & 1.5581E+1(1.3468E+0) & 2.1401E+0(2.9449E-1) & 9.3530E+0(2.8352E-1) \\
		SDP3  & 3     & 8.7498E-1(1.0329E-1) & 2.0218E+0(1.4425E-1) & 2.6696E+0(3.3988E-2) & 7.4566E-1(9.7200E-2) & 2.5846E+0(1.3062E-1) \\
		SDP8  & 3     & 1.0964E+0(8.8786E-3) & 1.1930E+0(1.3547E-2) & 1.1764E+0(4.3077E-3) & 1.1010E+0(1.1515E-2) & 1.2464E+0(5.8024E-3) \\
		FDA4  & 3     & 2.7403E-1(2.4337E-2) & 1.2458E+0(1.3708E-1) & 1.5893E+0(6.6551E-2) & 2.7081E-1(3.5707E-2) & 4.3455E-1(7.2947E-2) \\
		\bottomrule
	\end{tabular}%
	\label{tab:compres}%
	\vspace{-4mm}
\end{table*}%

\begin{figure*}[thb]
	\setlength{\tabcolsep}{-5pt}
	\begin{tabular}{ccc}		 
		\includegraphics[width=6.5cm,height=3cm]{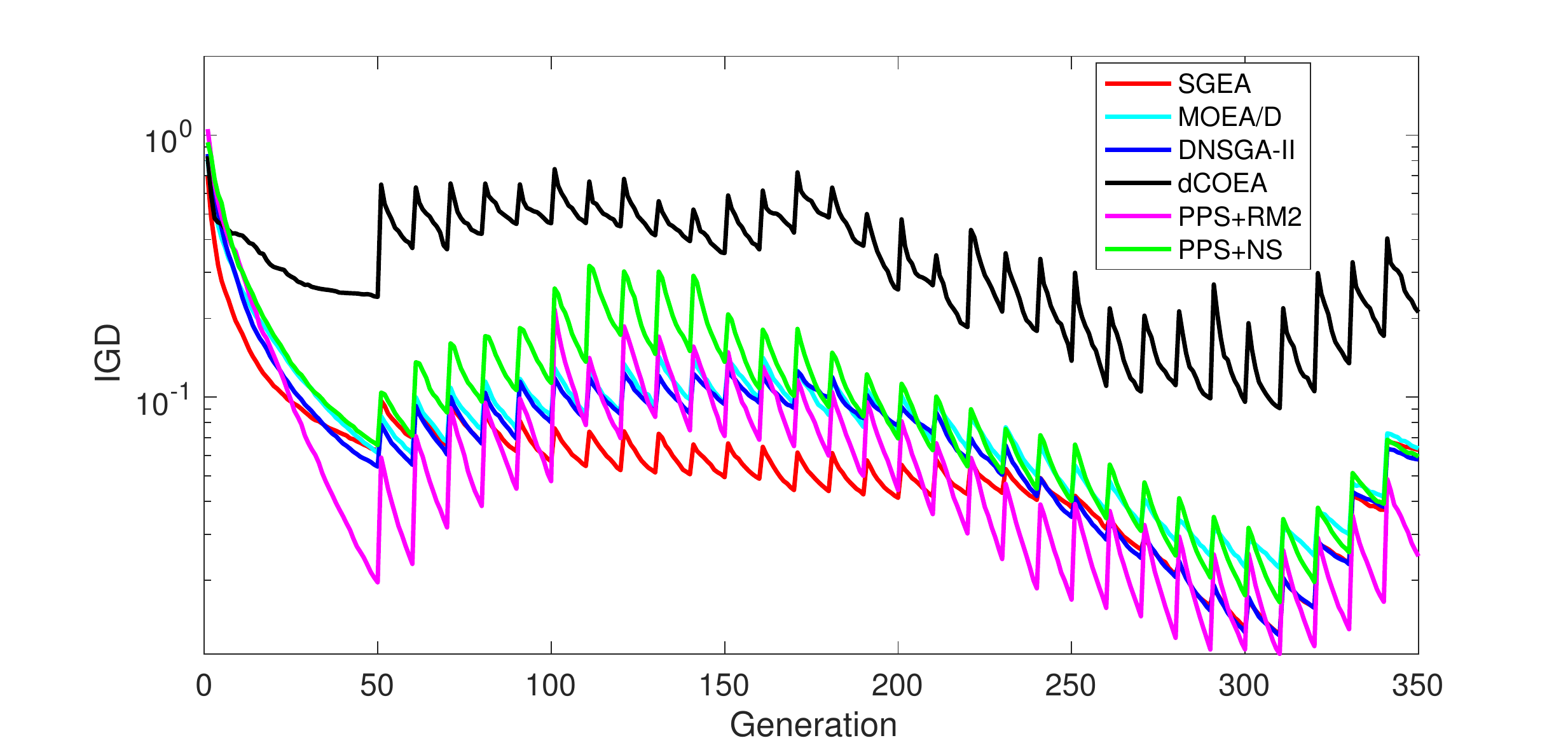} &
		\includegraphics[width=6.5cm,height=3cm]{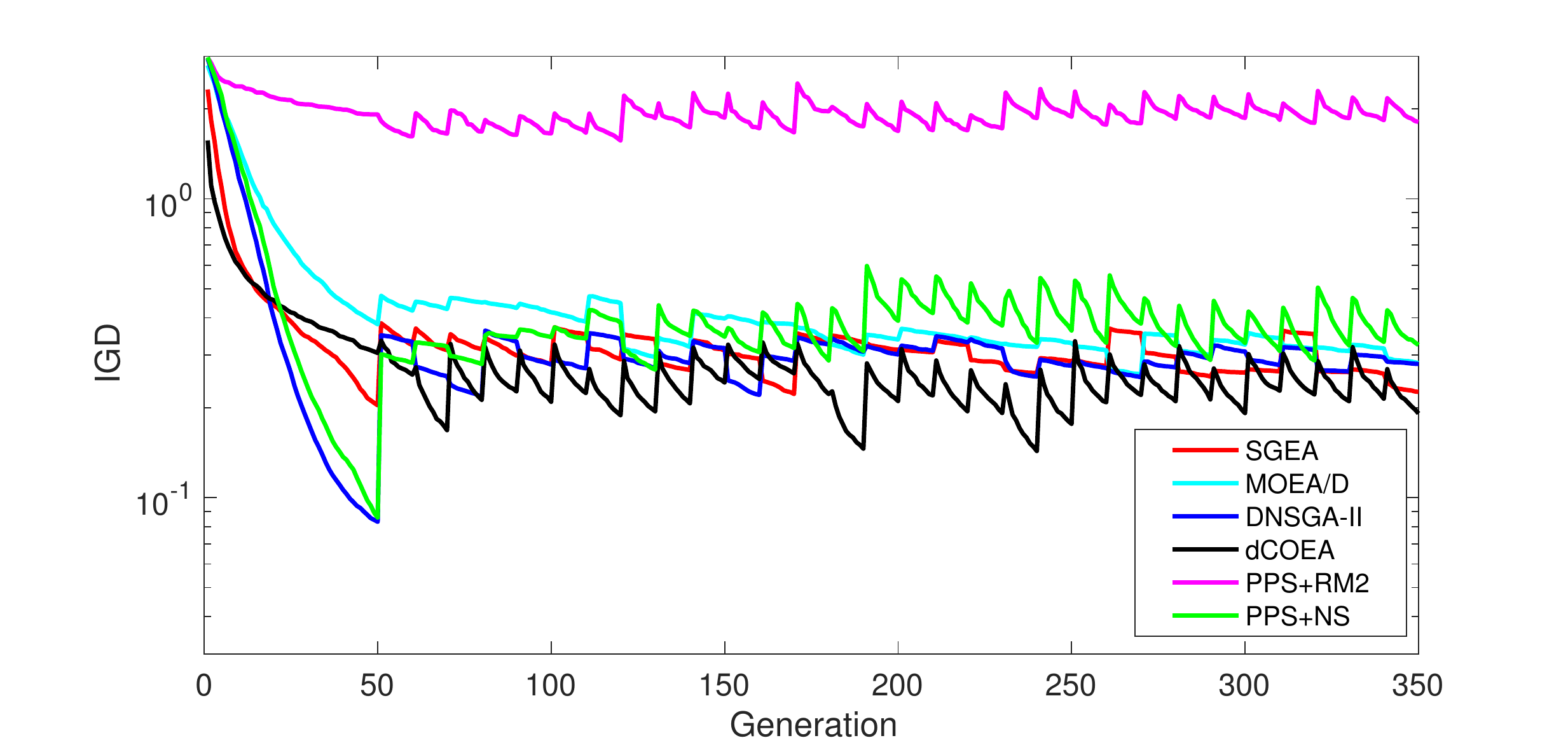} &
		\includegraphics[width=6.5cm,height=3cm]{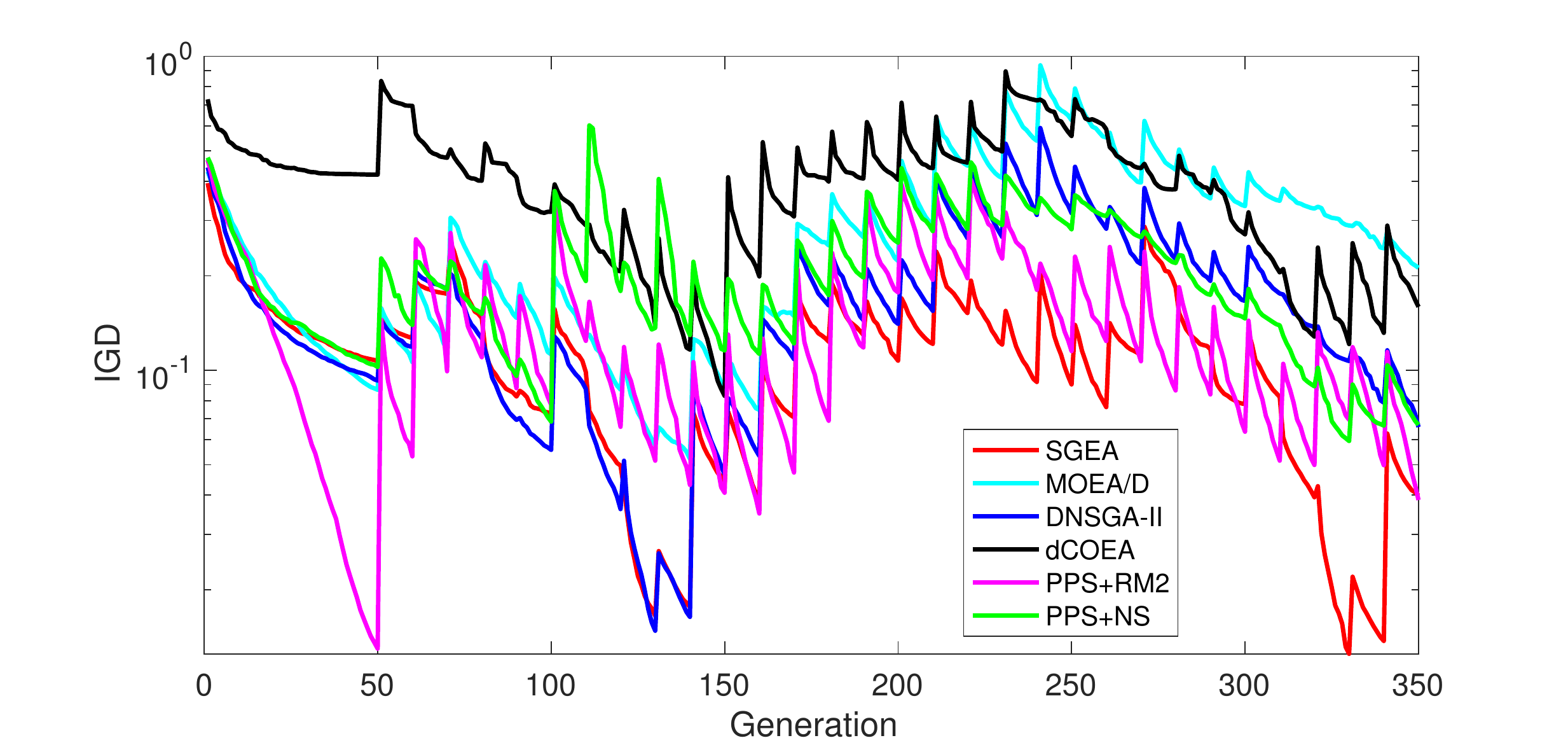}\\
		(a) SDP4 & (b) SDP7 & (c) SDP10 \\[-1mm]
	\end{tabular}
	\caption{Evolution curves of averaged IGD for selected problems.}
	\label{fig:igd}
	\vspace{-4mm}
\end{figure*}

\begin{figure*}[thb]
	\setlength{\tabcolsep}{-5pt}
	\begin{tabular}{ccc}
		\includegraphics[width=6.5cm, height=3cm]{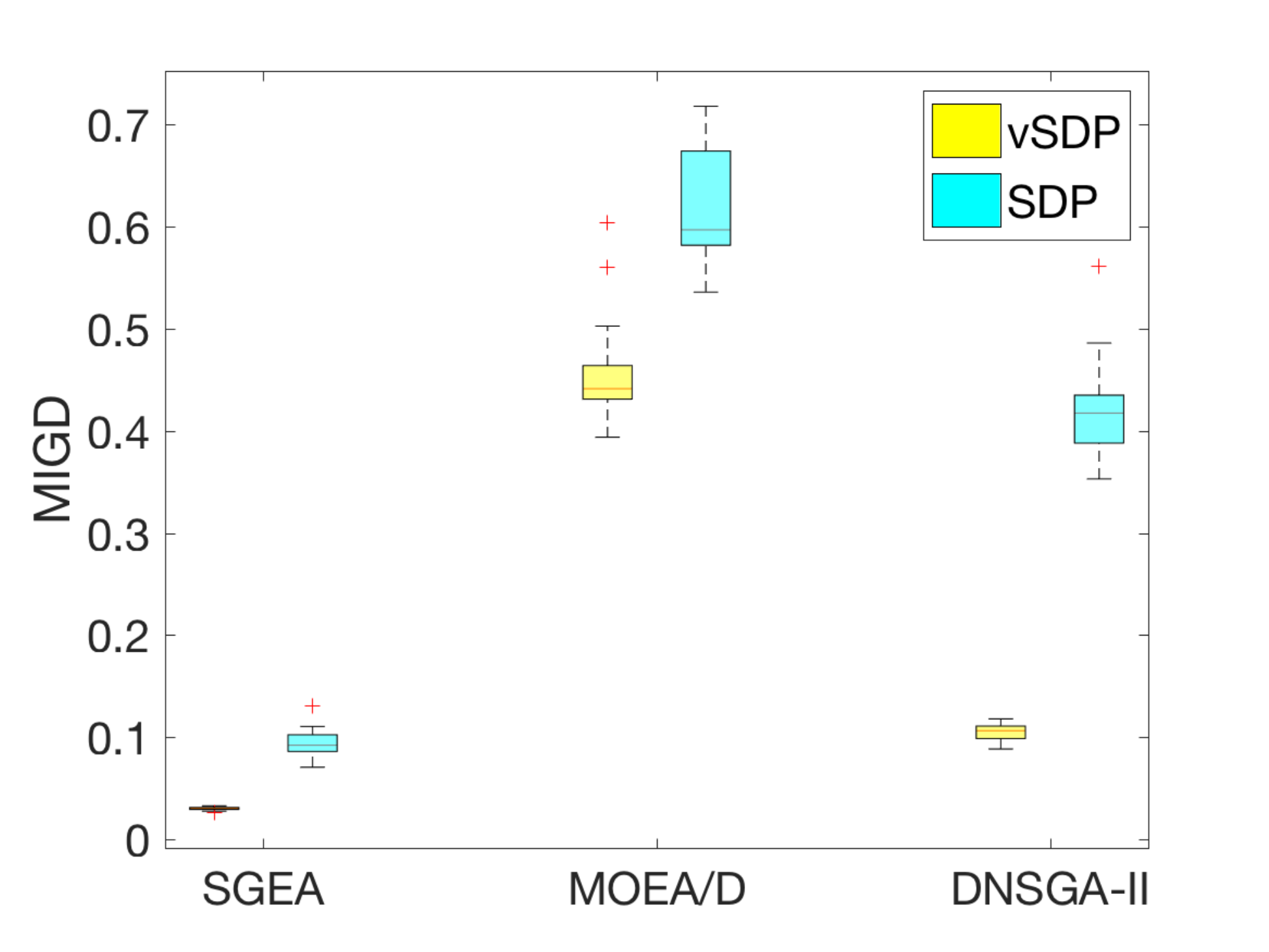}  & 
		\includegraphics[width=6.5cm,height=3cm]{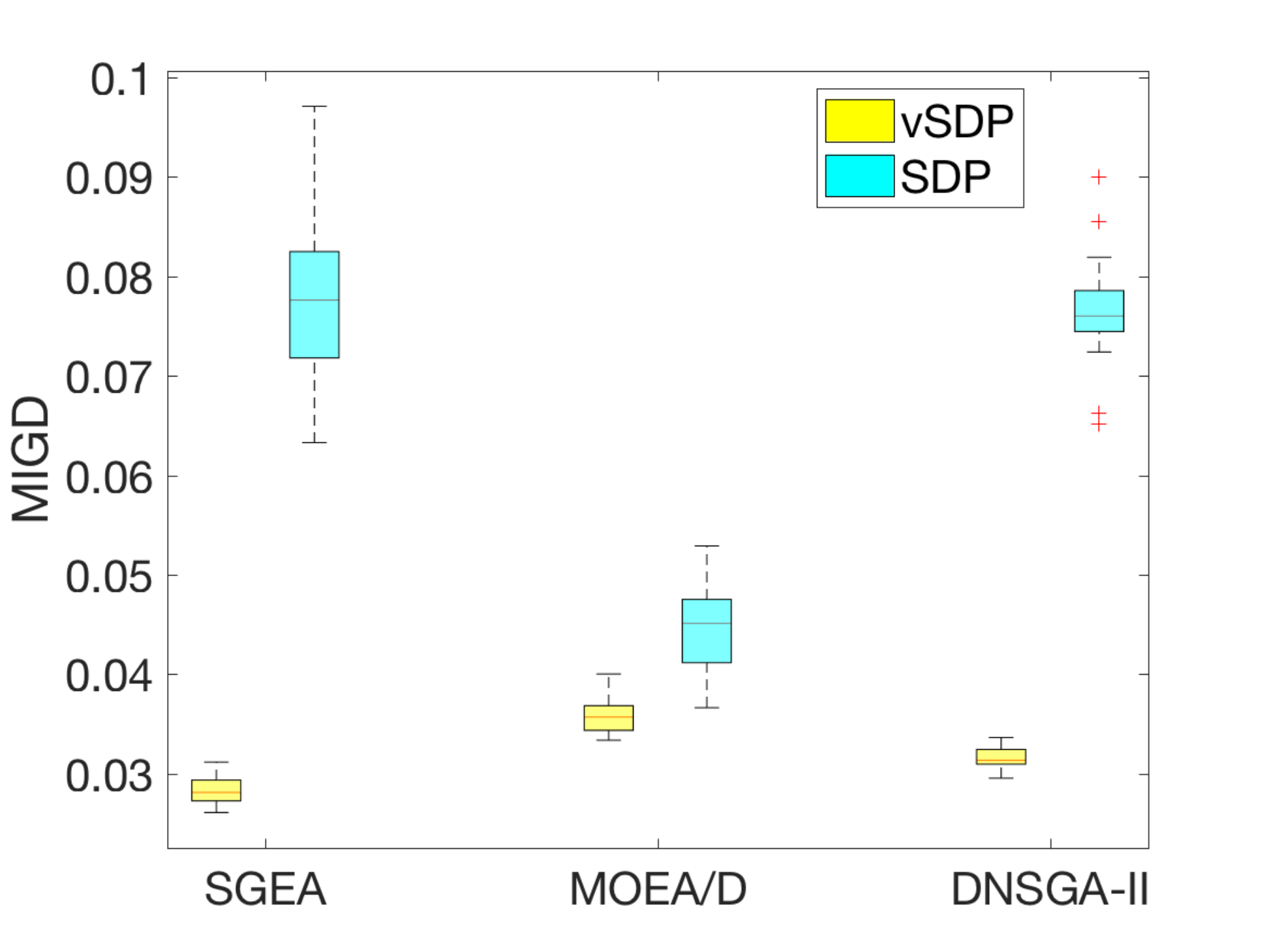} &
		\includegraphics[width=6.5cm,height=3cm]{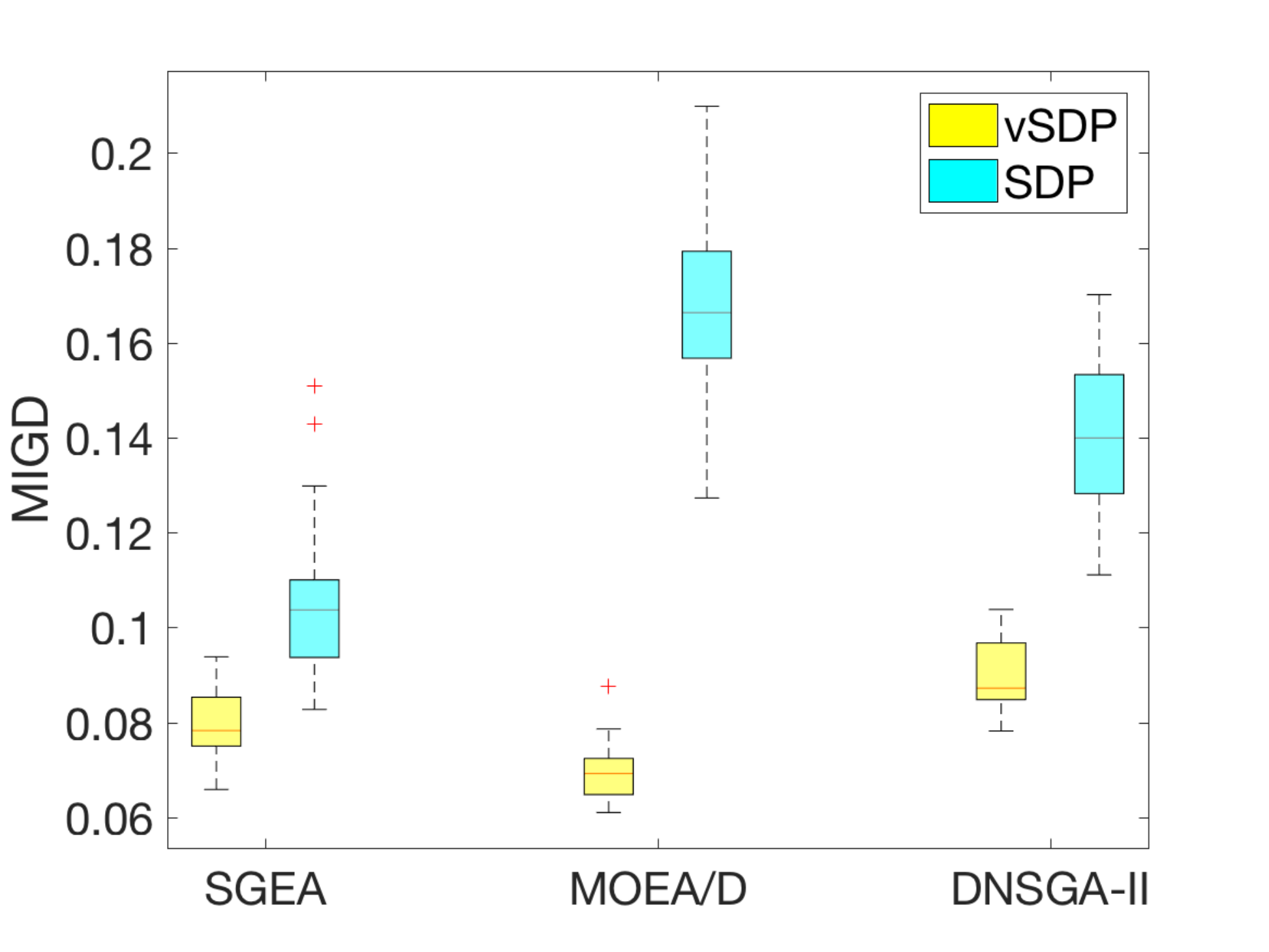} \\
		(a) vSDP1 vs SDP1 & (b) vSDP5 vs SDP5 & (c) vSDP9 vs SDP9 \\
	\end{tabular}
	\caption{Comparison between problems with and without SDP features.}
	\label{fig:vsdp}
	\vspace{-4mm}
\end{figure*}

SDP11 has a time-varying PF segment that is difficult to approximate. This dynamic affects considerably the performance of MOEA/D and DNSGA-II. dCOEA is more robust than the other algorithms for this problem. The variable-based multipopulation in dCOEA renders a good performance of tracking the movement of the search-unfavoured PF segment. The multipopulation strategy in dCOEA is further shown to be a big advantage for solving SDP12 with a changing number of variables. PPS-based algorithms also have good performance for these two problems.

SDP13-15 have very simple PSs and satisfy the condition of PPS that PSs should be similar in two consecutive environments. However, the two PPS-based algorithms show poor performance on these problems, demonstrating that dynamics in the number of objectives and degeneration affect greatly PPS-based algorithms. In contrast, dCOEA and SGEA are more suited to these kind of dynamics. 

Unsurprisingly, MOEA/D performs poorly on the SDP problems. This means that its fast convergence is no longer an advantage in handling dynamic scenarios like SDP. While fast convergence is desirable, it is more important to have a sound reaction response strategy when handling DMOPs.

We display the averaged IGD evolution curves of the algorithms for some selected 2-objective SDP problems in Fig. \ref{fig:igd} to illustrate the difficulty of the underlying dynamics. The IGD curves and PF approximations for other SDP instances can be found in the supplementary material. Clearly, the plots demonstrate that the SDP problems are able to identify the strengths and weaknesses of the algorithms.

\vspace{-4mm}
\subsection{Advantages of SDP Features} 
We empirically compare DMOPs with and without SDP features in order to understand how advantageous SDP is over existing test problems for testing algorithms. To do so, we created another SDP variant, named vSDP, by deactivating important SDP features and adding existing popular features to the original SDP1-14 (changes detailed in the supplementary material). vSDP1-14 was compared against SDP1-14 based on their MIGD values obtained by three algorithms. 

Fig.~\ref{fig:vsdp} shows the boxplots of MIGD values for the comparison between some selected SDP variants (please refer to the supplementary material for more comparisons). As seen, the SDP features are more challenging than existing features and they help to discriminate algorithms. For example, the three algorithms perform similarly on existing features in vSDP5 and vSDP9.
In contrast, they obtain significantly different results on the SDP features for the corresponding problems. In addition, some SDP features can reveal the sensitivity of algorithms. It is shown that DNSGA-II is highly sensitive to unpredictability in SDP1 whereas SGEA and MOEA/D not. MOEA/D is more robust than SGEA and DNSG-II to PF/expansion/shrinkage captured by SDP5. Thus, SDP has a great advantage in discriminating algorithms and understanding the sensitivity of algorithms when used.

\subsection{Comparison with Other Test Problems}
Here, we would like to make a comparison between SDP problems and some widely-used test problems. Three SDP problems are chosen to compare with three FDA \cite{Fari_04:1} problems and two dMOP \cite{GT09} problems, for which the number of variables was set to the same as SDP. The comparison includes both 2-objective and 3-objective scenarios (the MHVD metric is reported in \cref{tab:compres}). It  shows that the SDP problems are more challenging than the FDA and dMOP ones. The MHVD values for SDP problems are over an order of magnitude larger than most FDA or dMOP problems. Some algorithms that work well on existing problems experience performance deterioration when tested on SDP problems.

\begin{figure*}[t]
	\begin{tabular}{ccc}
		\includegraphics[width=5.7cm, height=3cm]{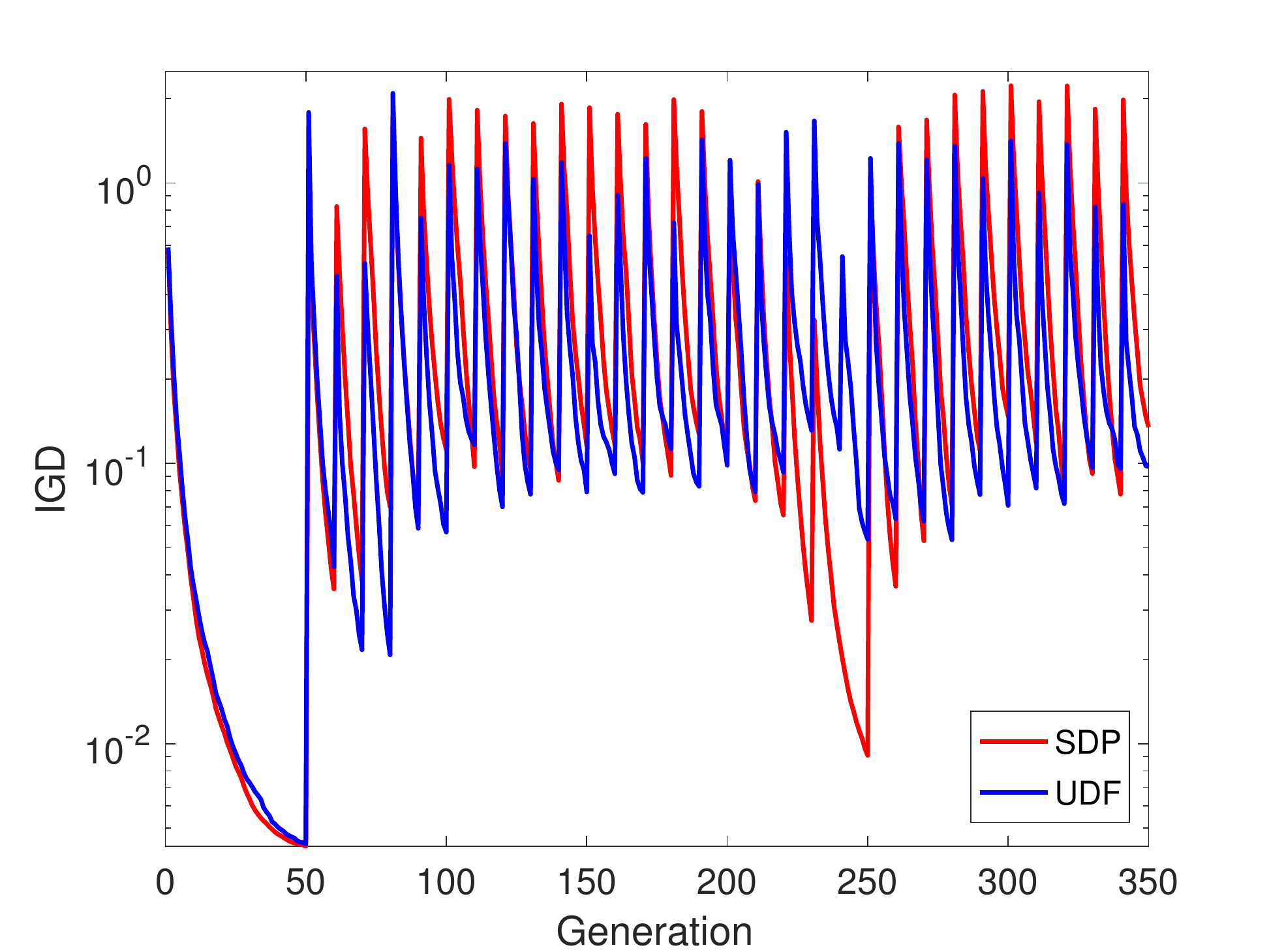}  & 
		\includegraphics[width=5.7cm, height=3cm]{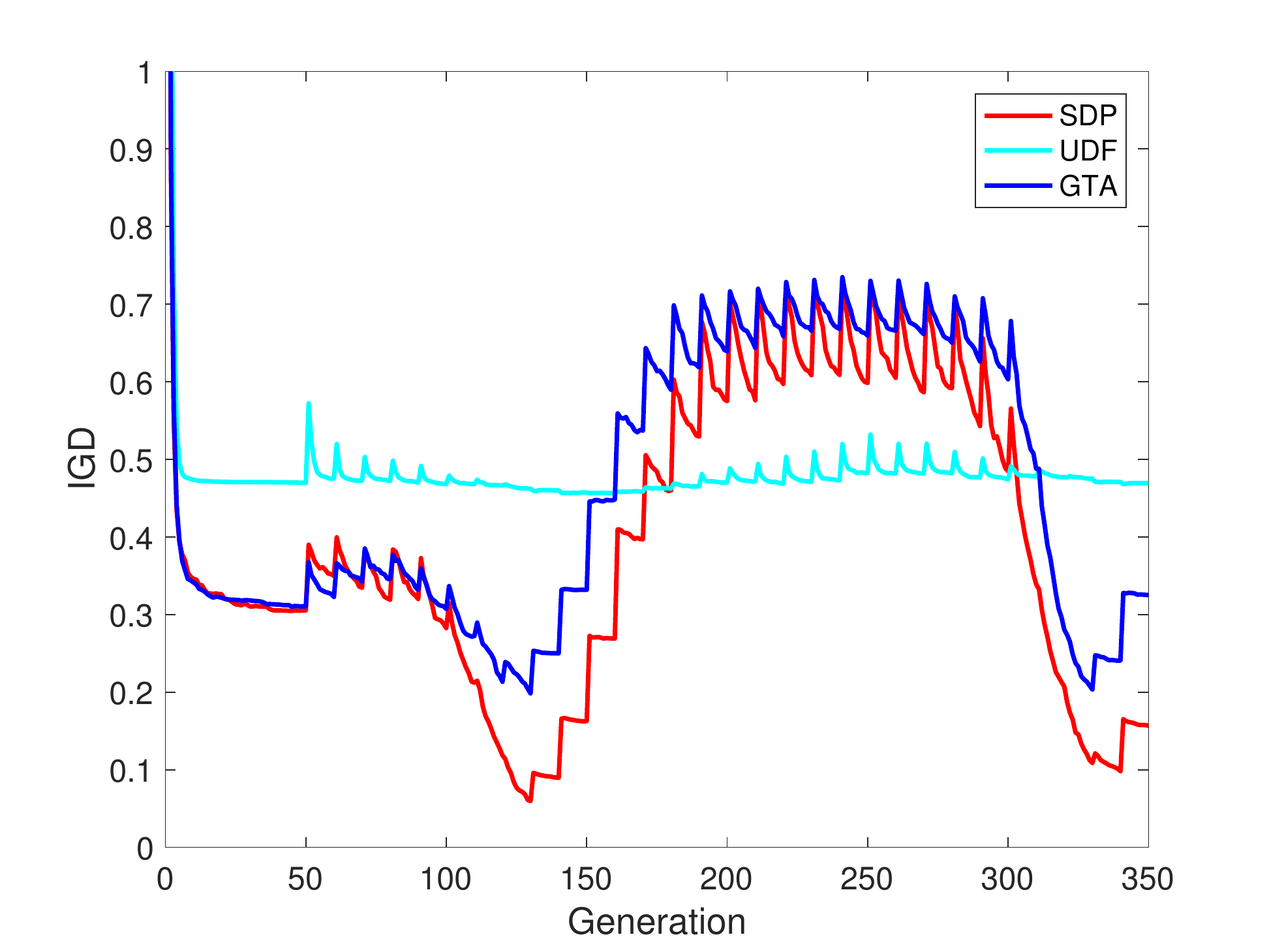} &
		\includegraphics[width=5.7cm, height=3cm]{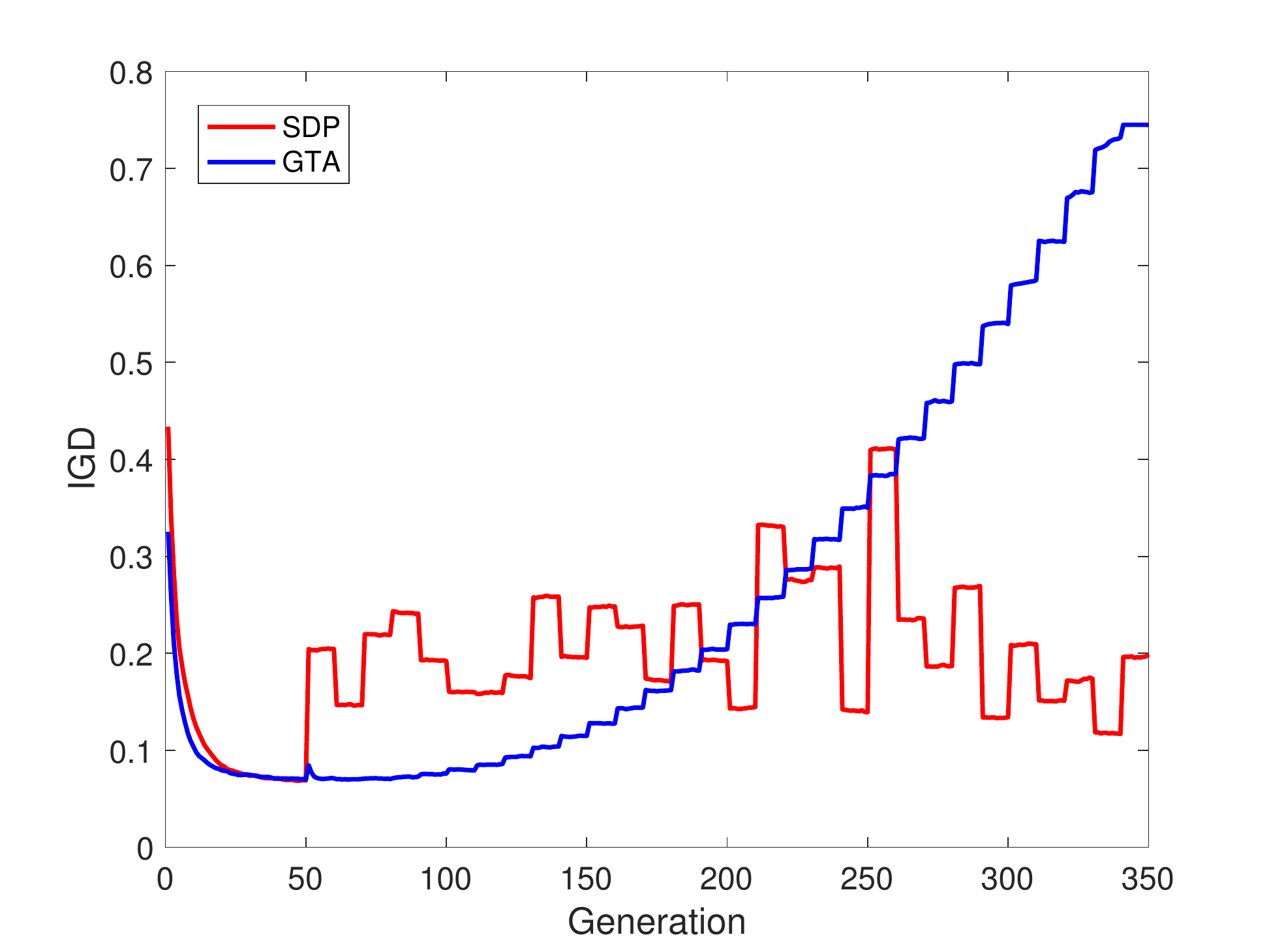} \\
		(a) unpredictability & (b) variable-linkage & (c) degeneration \\
	\end{tabular}
	\caption{Average IGD evolution curves obtained by SGEA for different predictability, variable-linkage, and degeneration schemes.}
	\label{fig:rnd_link_deg}
	\vspace{-2mm}
\end{figure*}

We noticed that some existing test suites, i.e., UDF \cite{BDSC14} and GTA \cite{Gee2017},  also recognise the importance of several SDP features, including variable linkage, degeneration, and unpredictability. Thus, we compared SDP against the test suites with these features (settings for comparison are detailed in the supplementary material), and the comparison is illustrated in Fig.~\ref{fig:rnd_link_deg}. It is observed from the figure that the unpredictability of SDP presents more intense changes to the algorithm SGEA than that of UDF  for most of time steps (when spikes appear). The difference is due to that variables in SDP have different magnitudes of random variations when an environmental change occurs, which is not the case with UDF. In Fig.~\ref{fig:rnd_link_deg}(b), variable linkages in SDP and GTA induce similar impacts on SGEA. In contrast, variable linkages in UDF seem to overwhelmingly outweigh dynamics and therefore make it difficult to analyse the effect of the underlying dynamics. In Fig.~\ref{fig:rnd_link_deg}(c),  SDP-based degeneration shows more diverse dynamics variations compared to GTA. The IGD values for GTA-based degeneration keeps increasing after the first environmental change. A further analysis on problem formalism reveals that degeneration in GTA occurs very rarely, only when the PF surface becomes a curve. It is the changing size of the PF in GTA that causes the deterioration of SGEA over time. This GTA feature is different from degeneration in SDP and thus is not effective to be used for studying dynamic degeneration. 

We also studied other important characteristics, i.e., spread of solutions, time-changing concavity-convexity, and disconnected PF components. Comparison of SDP and existing benchmark functions (mainly FDA \cite{Fari_04:1}, dMOP \cite{GT09},  and HE \cite{HE14}).

\begin{figure*}[!t]
	\begin{tabular}{cc}
		\includegraphics[width=0.5\linewidth]{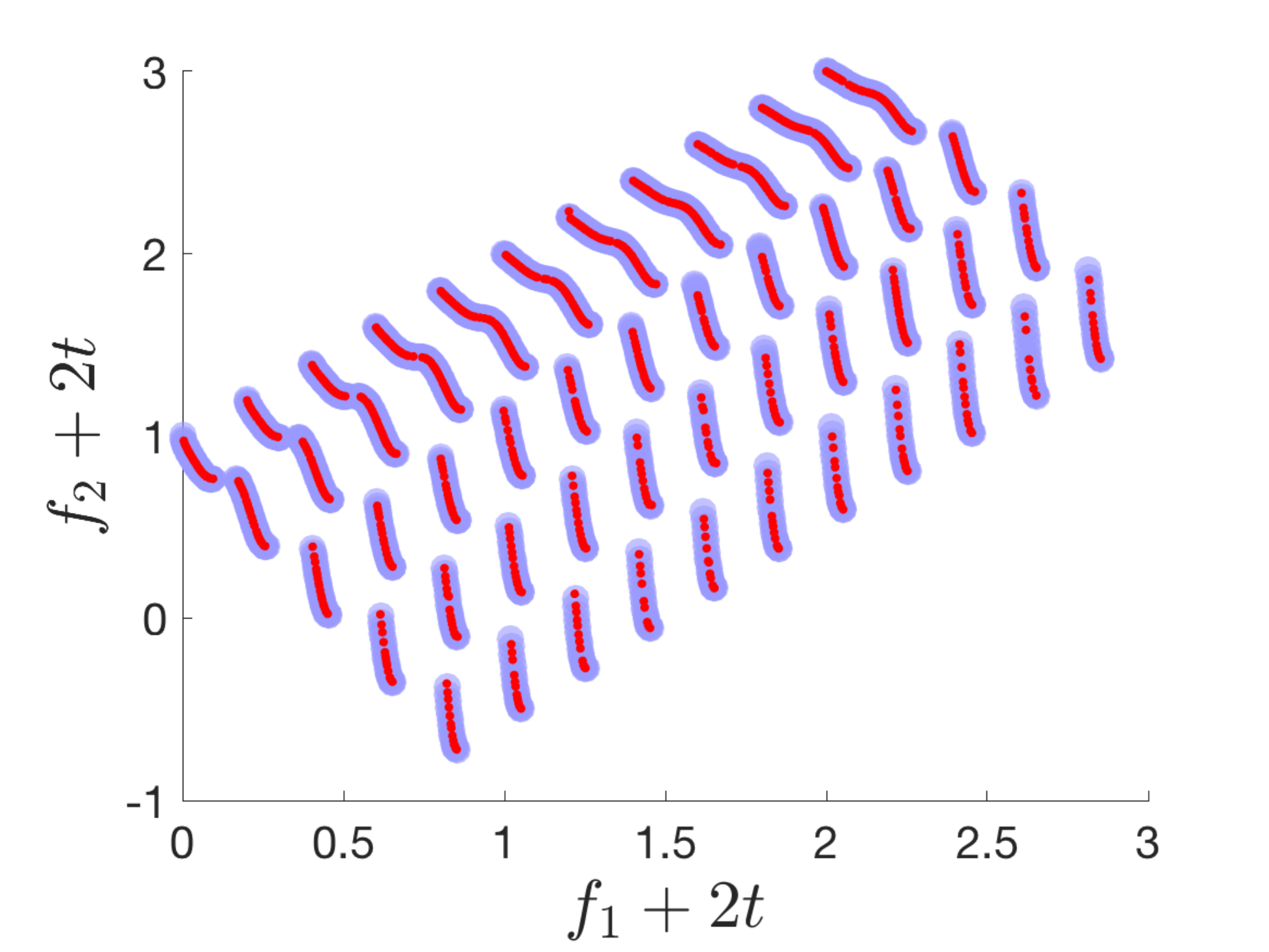}  & 
		\includegraphics[width=0.5\linewidth]{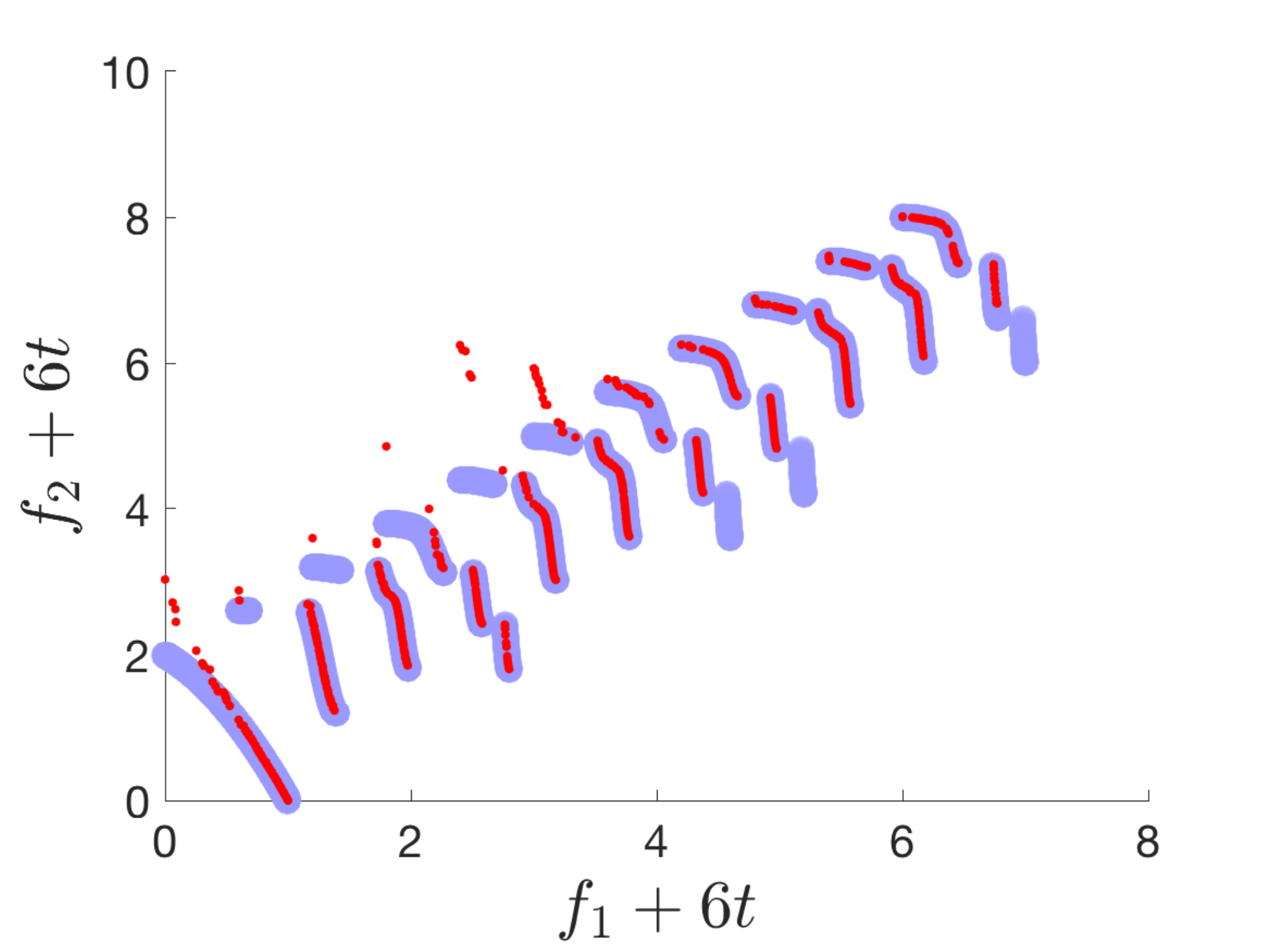} \\
		(a) HE2 & (b) SDP10 \\
	\end{tabular}
	\caption{A comparison of time-varying disconnectivity for HE2 and SDP10 in 2-objective case. The PF is in light blue and its approximation in red.}
	\label{fig:discon}
	\vspace{-2mm}
\end{figure*}
\begin{figure*}[!t]
	\begin{tabular}{cc}
		\includegraphics[width=0.45\linewidth]{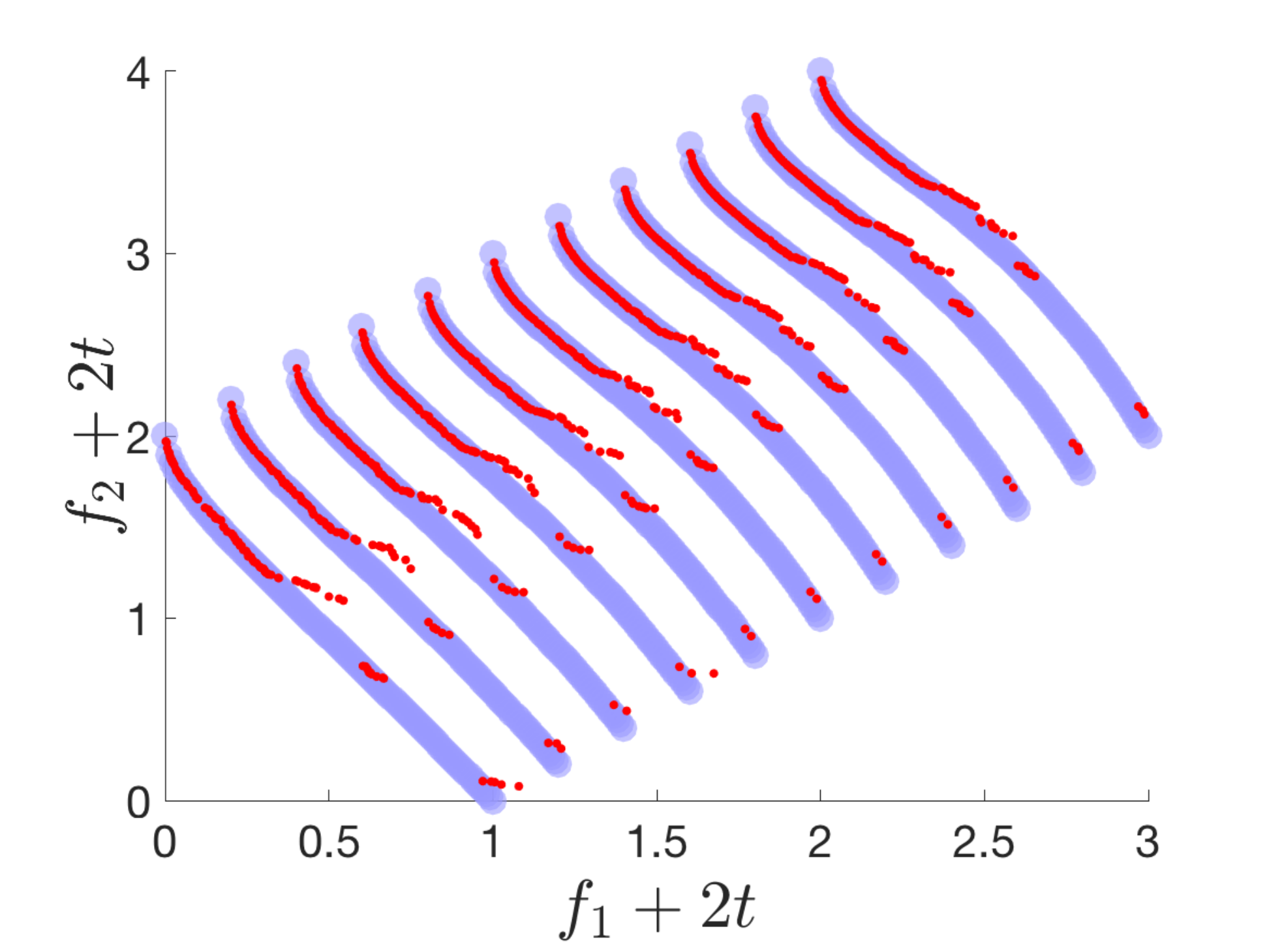}  & 
		\includegraphics[width=0.45\linewidth]{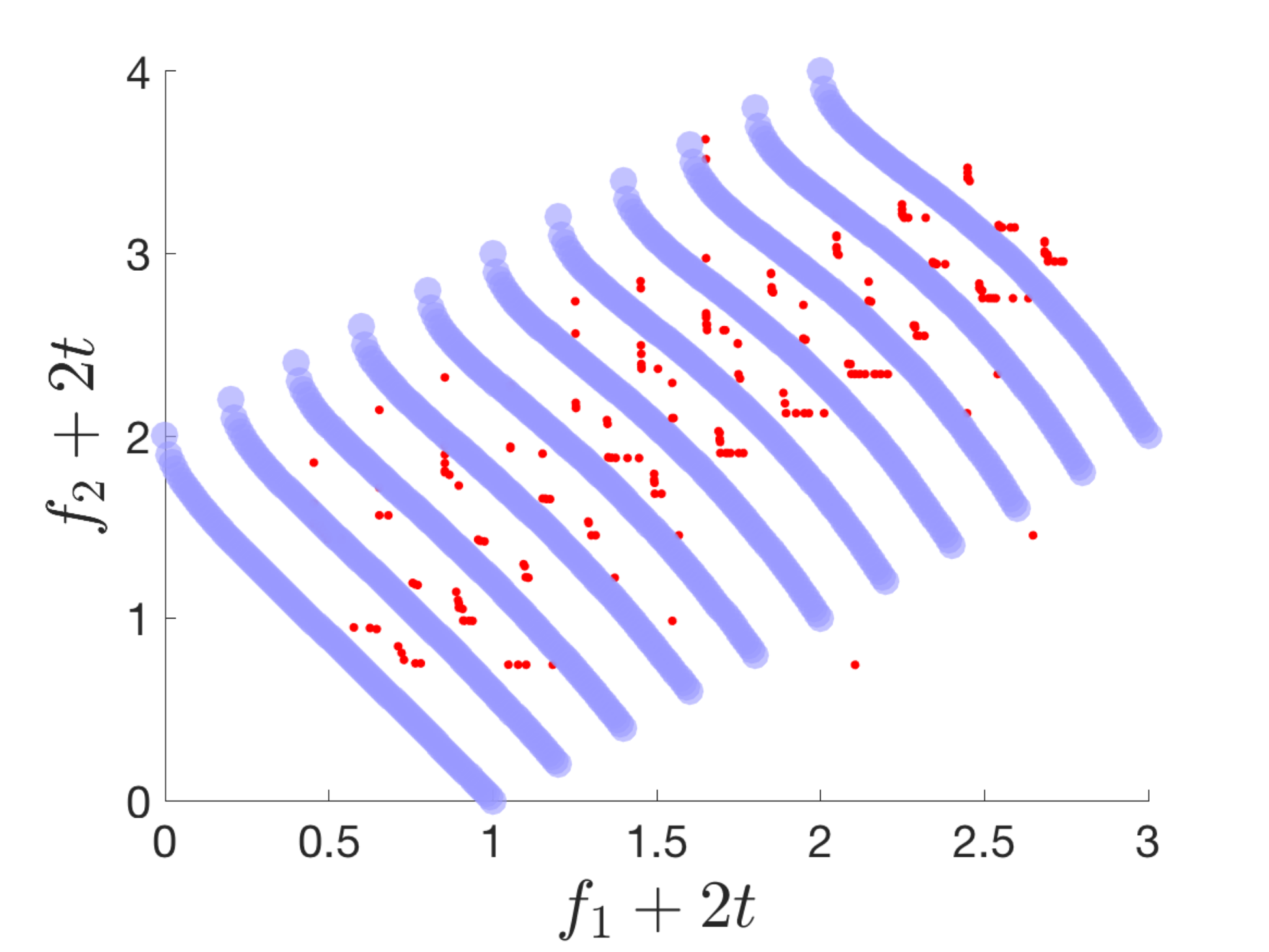} \\
		(a) HE7 & (b) HE9 \\
		\includegraphics[width=0.45\linewidth]{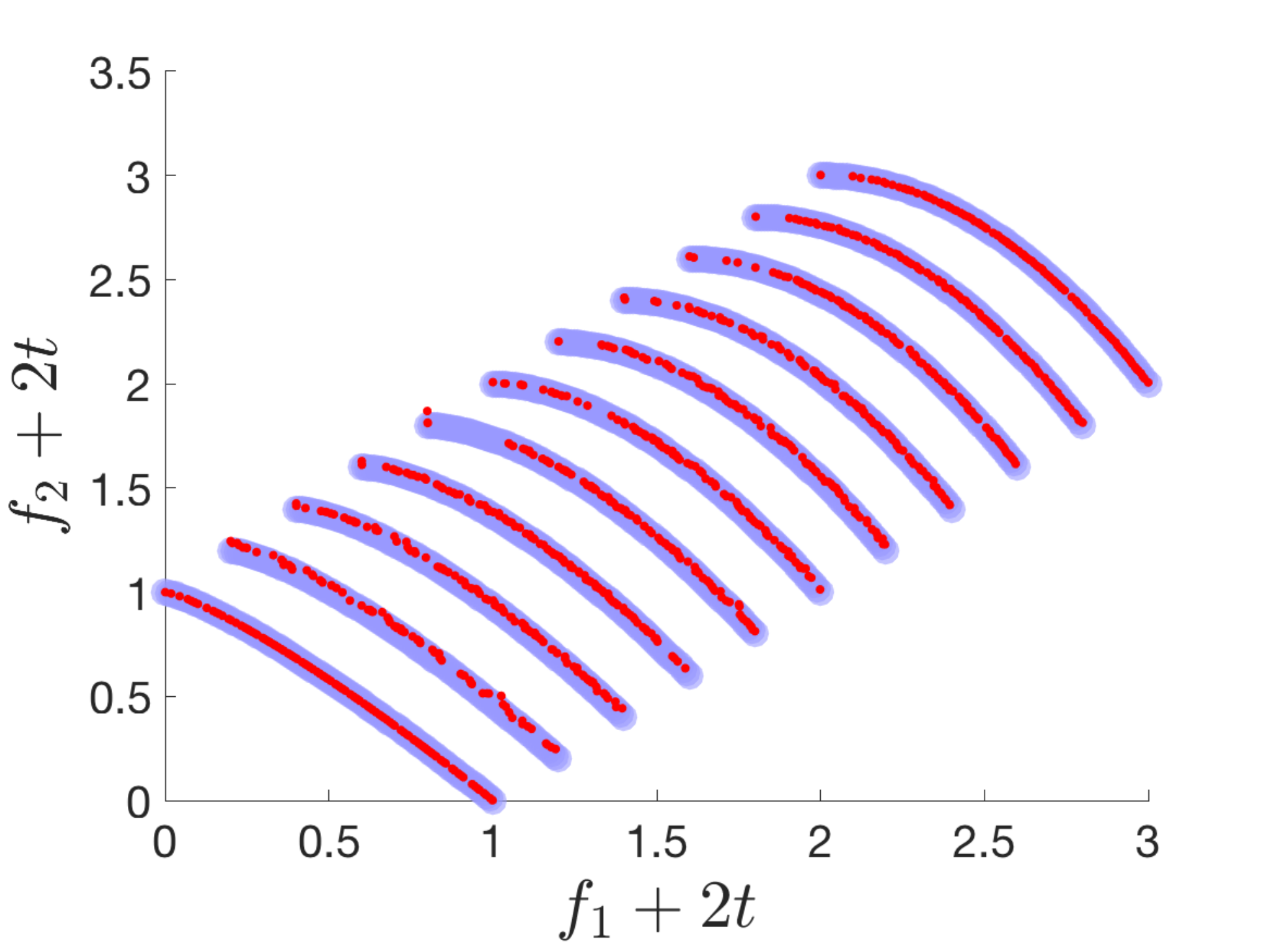}&
		\includegraphics[width=0.45\linewidth]{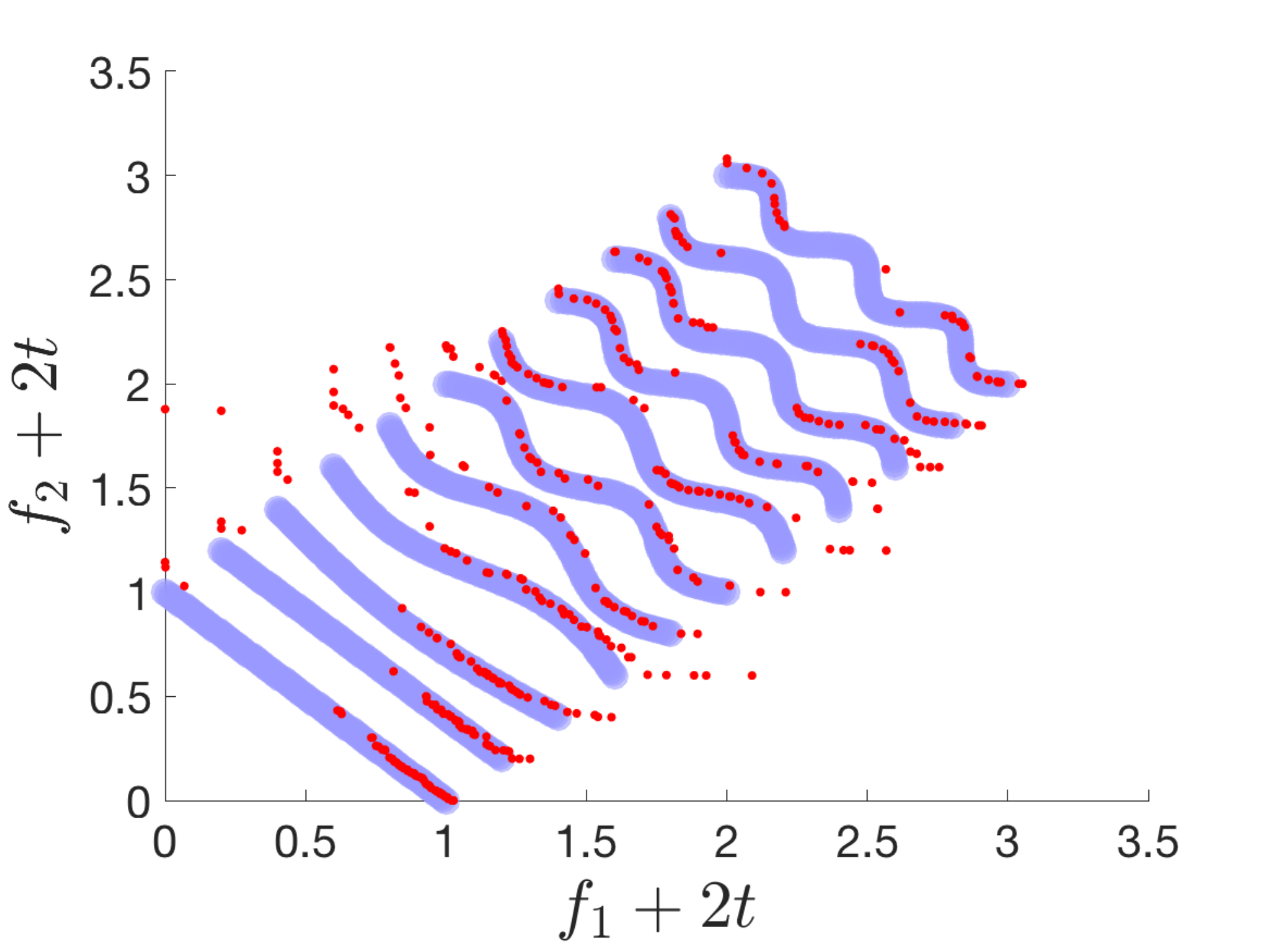} \\
		(c) dMOP2 & (d) SDP4 \\
	\end{tabular}
	\caption{A comparison of time-varying convexity-concavity for HE7, HE9, dMOP2 and SDP4 in 2-objective case. The PF is in light blue and its approximation in red.}
	\label{fig:convexity}
	\vspace{-2mm}
\end{figure*}

Disconnectivity is an important feature in DMOPs. Here we compared a disconnected SDP problem (i.e., SDP10), with an existing problem (i.e., HE2 \cite{HE14}). PF approximations obtained by SGEA are shown in Fig.~\ref{fig:discon}. We can see from the figure that SGEA is capable of locating all disconnected PF components of HE2 over time. In contrast, SGEA does not work well for SDP10: it could not always identify all disconnected PF components, and the changing number of disconnected PF components challenges greatly the diversity maintenance of SGEA. Thus, the unconventional disconnectivity property in SDP10 facilitates a new way to study algorithms systematically.

\begin{figure*}[t]
	\begin{tabular}{cc}
		\includegraphics[width=0.45\linewidth]{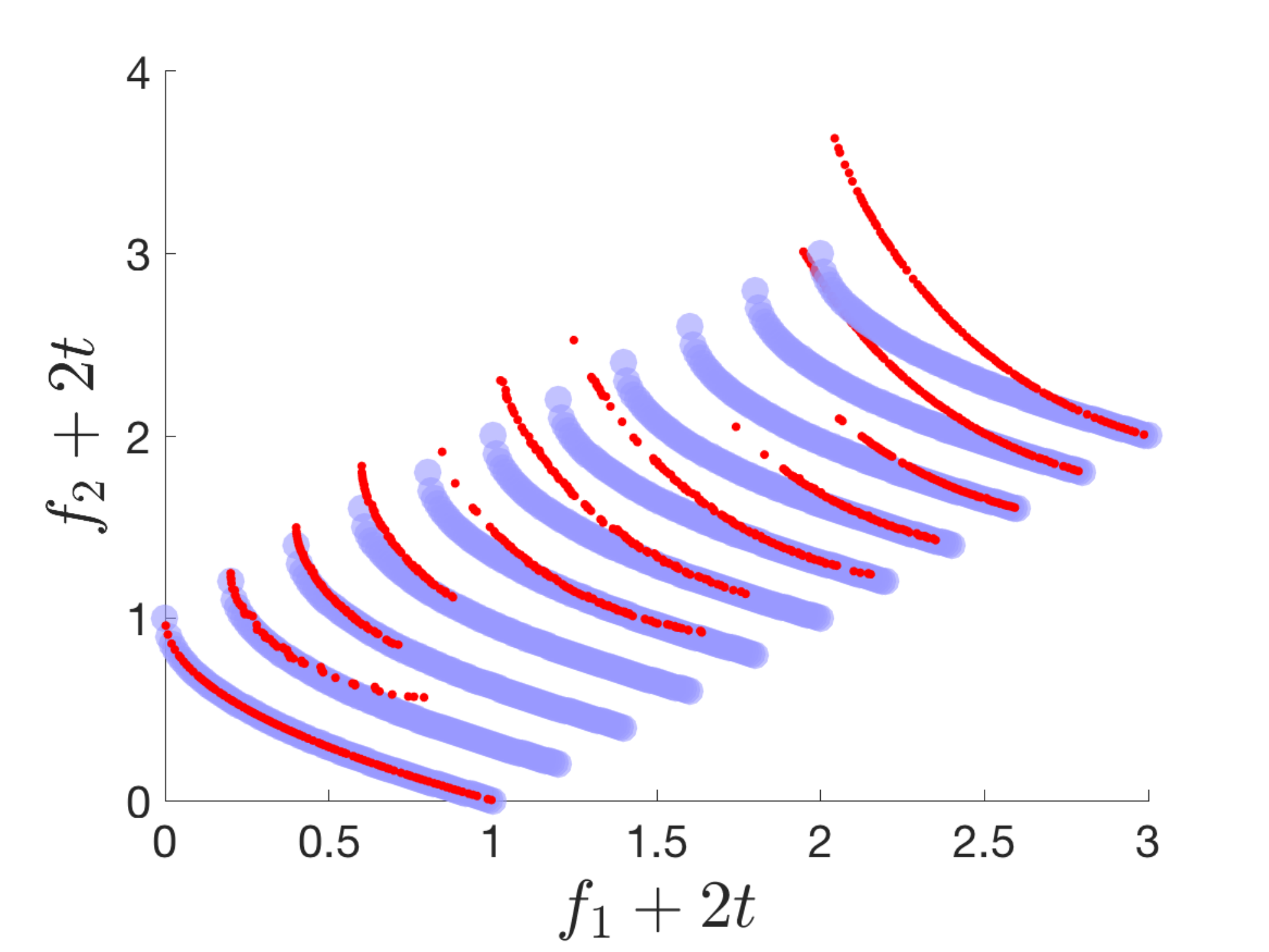}  & 
		\includegraphics[width=0.45\linewidth]{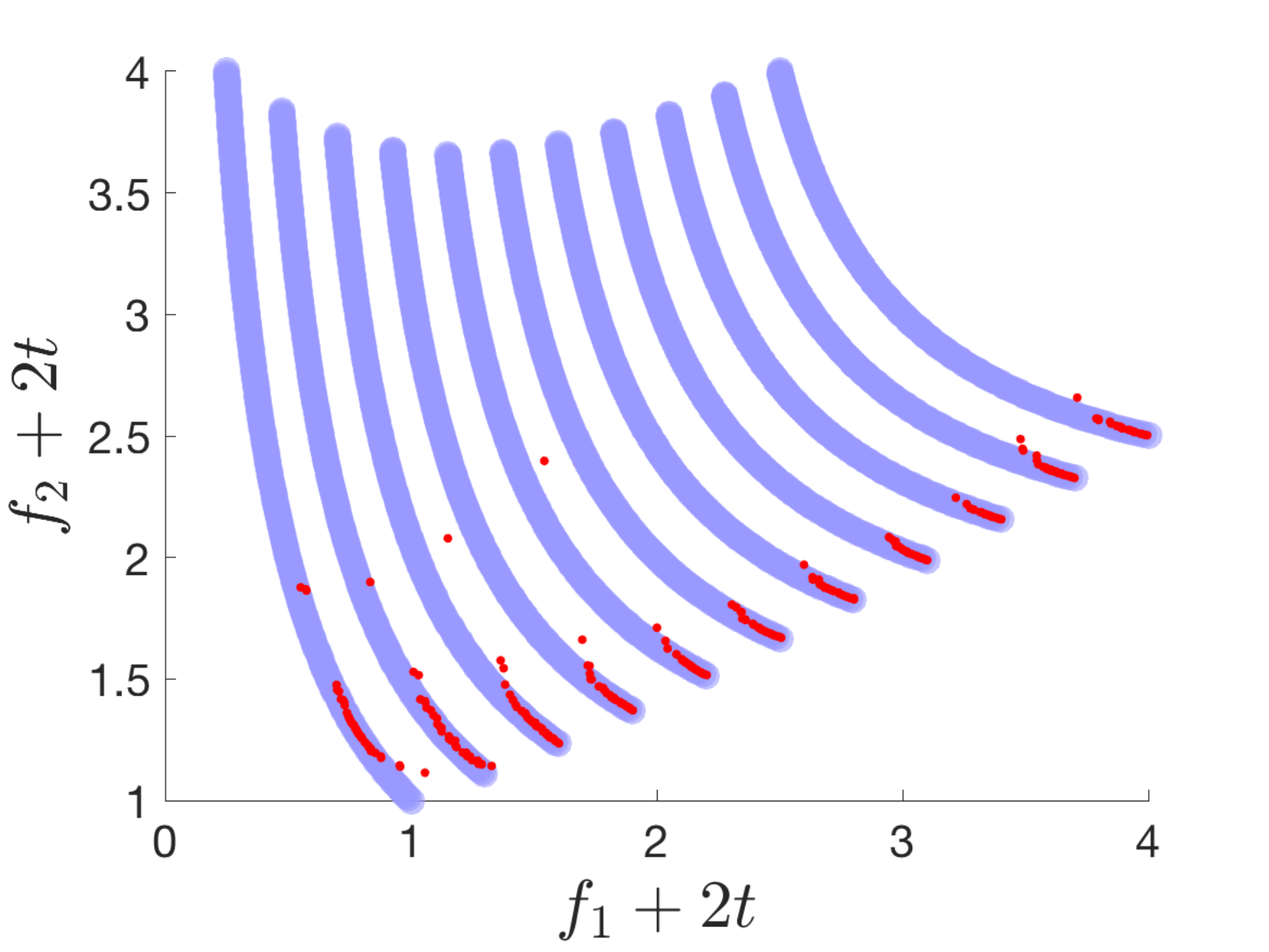} \\
		(a) dMOP3 & (b) SDP2 \\
	\end{tabular}
	\caption{A comparison of spread of solutions for dMOP3 and SDP2 in 2-objective case. The PF is in light blue and its approximation in red.}
	\label{fig:spread1}
	\vspace{-2mm}
\end{figure*}
\begin{figure*}[!t]
	\begin{tabular}{c}
		\includegraphics[width=0.9\linewidth, height=7cm, trim=3cm 0cm 3cm 0 cm]{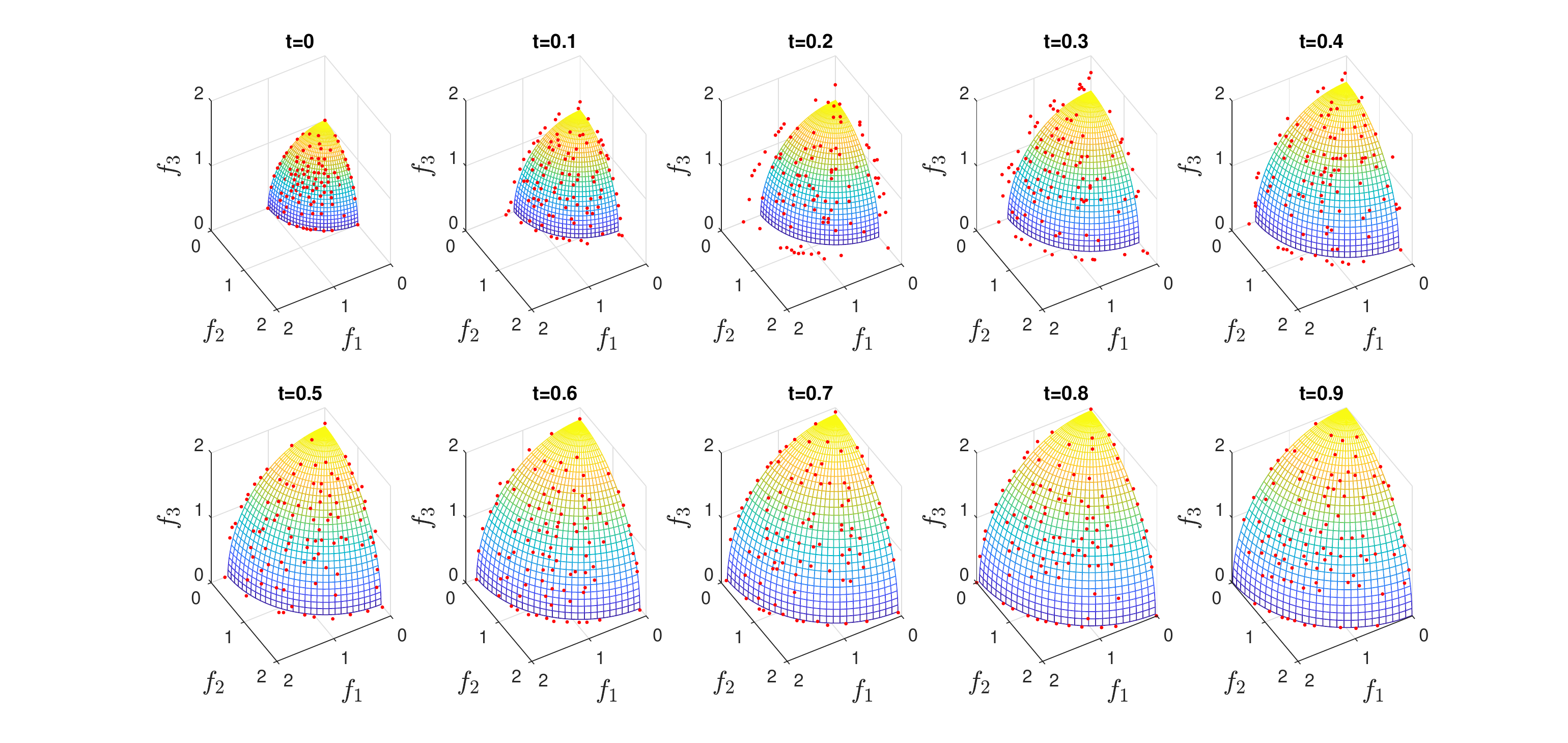}\\
		(a) FDA5\\
		\includegraphics[width=0.9\linewidth, height=7cm, trim=3cm 0cm 3cm 0 cm]{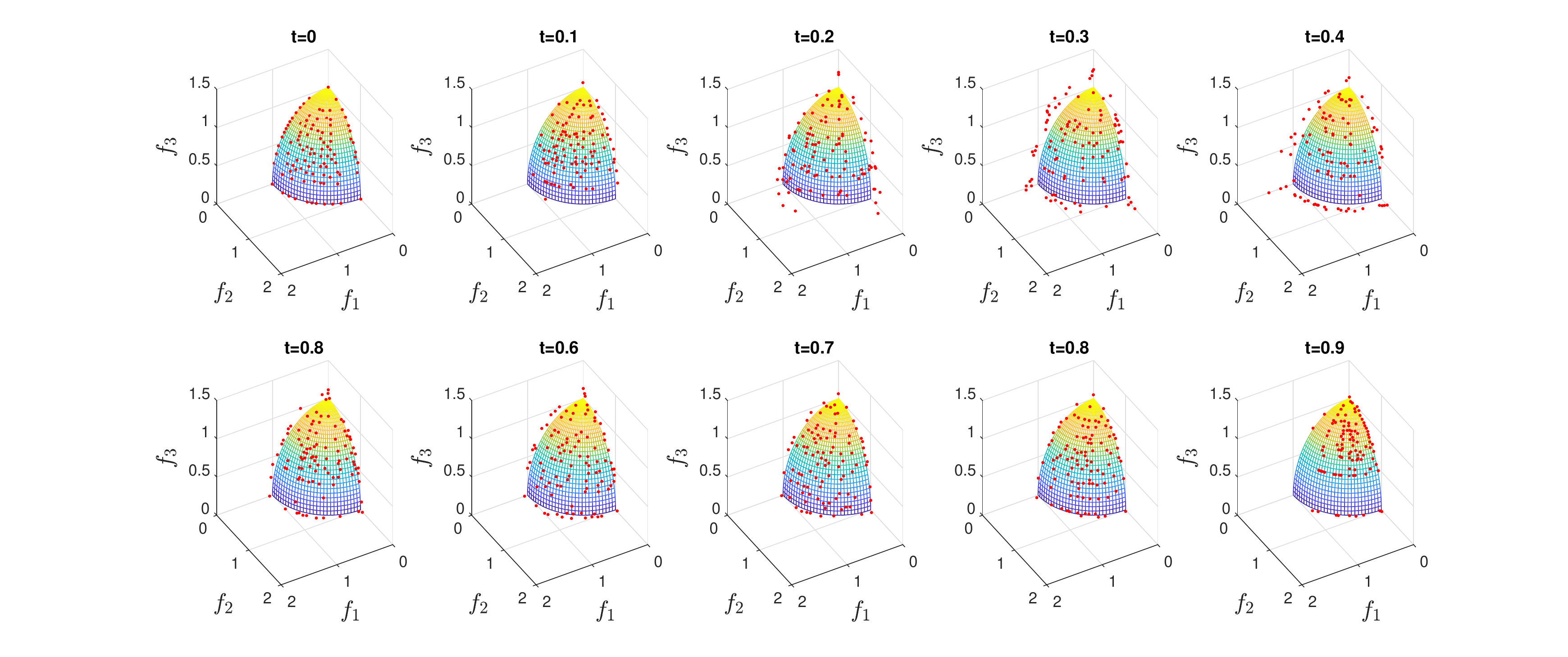} \\
		(b) FDA$_{iso}$\\
		\includegraphics[width=0.9\linewidth, height=7cm, trim=3cm 0cm 3cm 0 cm]{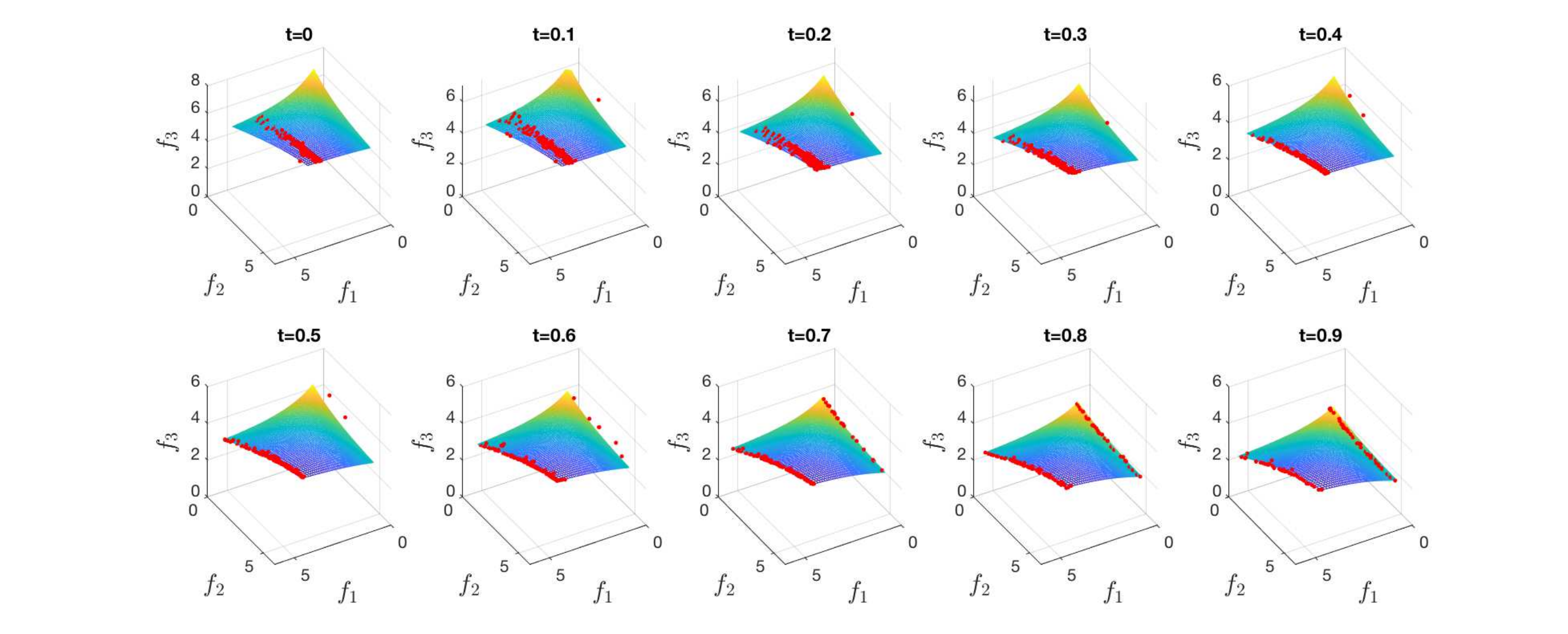} \\
		(c) SDP2 \\
	\end{tabular}
	\caption{A comparison of spread of solutions for FDA5, FDA$_{iso}$ and SDP2 in 3-objective case. The PF is in light blue and its approximation in red.}
	\label{fig:spread2}
\end{figure*}

Time-varying convexity-concavity of problems is not new in benchmarking dynamic environments, and there are already some well-defined benchmark functions whose PF can switch from convexity to concavity over time, and vice versa. We compared one of our test problems ,i.e., SDP4, with three existing problems having this property. The comparison of PF approximations of SGEA is presented in Fig.~\ref{fig:convexity}. It can be seen that SDP4 has larger changes in PF geometry than the existing ones. There is no difficulty for SGEA to track convexity-concavity changes in dMOP2 \cite{GT09}. SGEA identified correctly part of the PF for HE7 but failed completely for HE9. One main reason for the poor performance of SGEA is that the two HE problems involves additionally strong variable linkages, which explains the poor coverage of solutions on the PF. More importantly, strong variable linkages make it difficult to reach a clear conclusion whether the failure of algorithms is mainly due to environmental changes. In contrast, the lack of solutions on the concave part of the PF for several time steps shows that SGEA solves convex PFs better than concave ones. In this sense,the time-varying convexity-concavity we introduced in SDP test suite is more suitable to deeply analyse algorithms' performance.

Some existing benchmark functions have the feature of SDP2 that the spread of solutions varies over time. Here, we would like to compare SDP2 with these existing ones to see how unique SDP2 is. Specifically, SDP2 was compared with dMOP3 \cite{GT09} in 2-objective case, and with FDA5 \cite{Fari_04:1} and FDA5$_{iso}$ \cite{HE14} in 3-objective case. The PF approximations obtained by SGEA for the first several time points ($t=0, 0.1, \dots$) are plotted in Figs.~\ref{fig:spread1} and \ref{fig:spread2}. In 2-objective case, the loss of spread of solutions on dMOP3 due to changes is likely to be recovered in latter environments, although the convergence of solutions gets aggravated. In contrast, the spread of solutions on SDP2 becomes smaller and smaller as time goes by. In the 3-objective case, both FDA5 and FDA$_{iso}$ are able to show how widespread solutions are on the PF, and the latter seems more competent to do so. The 3-objective SDP2, however, illustrates variations in solution spread more clearly than the FDA5 variants, demonstrating the usefulness of SDP2 to examine algorithms' ability to spread solutions widely.

\subsection{SDP as Adversarial Problems}
SDP has another use for adversarial examples to optimisation solvers. It is shown in previous sections that the existing solvers encounter vast difficulties in a number of real-life dynamic features captured by SDP. We highlight these features that are poorly understood in the literature: time-dependent multimodality (SDP2-3), time-dependent knees (SDP3-4), deceptive changes (SDP6-7), dynamics in disconnectivity (SDP8-10), dynamics in objective and/or variable scalability (SDP12-13), and dynamic degeneracy (SDP14-15). These SDP adversarial examples facilitate a tool not only to better understand the robustness of solvers and make the right decision on selection of solvers for problems in question, but also to develop advanced techniques against them.

\section{Conclusion}
Dynamic multiojective optimisation has received growing attention over the last 10 years, yet the corresponding test environments facilitating algorithm analysis have not been well understood. While this paper helps to provide a better understanding of DMOPs, additional issues are identified after a wide review of existing test problems. These issues limit the assessment of EAs in dynamic environments.

Having realised the limitations of existing DMOPs, we highlighted a diverse set of dynamics and features that a good test suite should have. We also developed a SDP test suite to include desirable dynamics that have been rarely considered in the DMO literature but occur frequently in real life. Our experiments showed that the SDP test suite presents different degrees of difficulty to the algorithms tested and is able to identify the strengths and weaknesses of these algorithms. Furthermore, it was demonstrated that the SDP problems can test EAs in ways that commonly used DMOPs can not. SDP can be also used as adversarial examples for solvers.

It is expected that more advanced algorithms will be available for DMO, and a wider range of DMOPs as well. The SDP test suite could be adjusted or modified to create appropriate scenarios for other types or test environments, e.g. constrained and/or noisy DMOPs.

\bibliography{IEEEabrv}

\end{document}